\documentclass[10pt,journal,compsoc]{IEEEtran}

\ifCLASSOPTIONcompsoc
  \usepackage[nocompress]{cite}
\else
  \usepackage{cite}
\fi

\ifCLASSINFOpdf
  \usepackage[pdftex]{graphicx}
\else
  \usepackage[dvips]{graphicx}
\fi

\usepackage{amsmath}

\usepackage{comment}

\usepackage[breaklinks=true,bookmarks=false]{hyperref}
\usepackage{amssymb}
\usepackage{booktabs}
\usepackage{epsfig}
\usepackage{multirow}
\usepackage{placeins}
\usepackage{times}
\usepackage{tikz}
\usepackage{pgfplots}
\usepackage{xspace}
\usepackage{review}
\usepackage{comment}
\pgfplotsset{compat=newest}
\usepackage{xcolor}

\usepackage{pifont}
\usepackage[export]{adjustbox}
\usepackage{verbatimbox}

\usepackage[breaklinks=true,bookmarks=false]{hyperref}

\usepackage{cellspace}
\def\eg{\textit{e.g.}\xspace}
\def\ie{\textit{i.e.}\xspace}

\newcommand{\bx}{\boldsymbol{x}}
\newcommand{\by}{\boldsymbol{y}}
\newcommand{\bz}{\boldsymbol{z}}
\newcommand{\bp}{\boldsymbol{p}}

\newcommand*{\black}{\textcolor{black}}

\begin{document}
\title{Self-Supervised Learning Across Domains}

\author{Silvia Bucci, Antonio~D'Innocente, Yujun Liao, Fabio M. Carlucci, Barbara Caputo and~Tatiana Tommasi
\IEEEcompsocitemizethanks{
\IEEEcompsocthanksitem S. Bucci, B. Caputo, T. Tommasi are with Politecnico di Torino, Italy, Italian Institute of Technology. E-mail \{name.surname\}@polito.it
\IEEEcompsocthanksitem A. D'Innocente is with University of Rome Sapienza, Italy, Italian Institute of Technology. E-mail: dinnocente@diag.uniroma1.it
\IEEEcompsocthanksitem Y. Liao is with Politecnico di Torino, Italy. E-mail  s274673@studenti.polito.it
\IEEEcompsocthanksitem F.M. Carlucci is with Huawei AI Theory, London. Work done while at University of Rome Sapienza, Italy. E-mail: fabio.maria.carlucci@huawei.com 
}
}

\markboth{Journal of \LaTeX\ Class Files,~Vol.~..., No.~..., Month~...}%
{Shell \MakeLowercase{\textit{et al.}}: Bare Demo of IEEEtran.cls for Computer Society Journals}

\IEEEtitleabstractindextext{%
\begin{abstract}
Human adaptability relies crucially on learning and merging knowledge from both supervised and unsupervised tasks: the parents point out few important concepts, but then the children fill in the gaps on their own. This is particularly effective, because supervised learning can never be exhaustive and thus learning autonomously allows to discover invariances and regularities that help to generalize.
In this paper we propose to apply a similar approach to the problem of object recognition across domains: our model learns the semantic labels in a supervised fashion, and broadens its understanding of the data by learning from self-supervised signals on the same images.  {This secondary task helps the network to focus on object shapes, learning concepts like spatial orientation and part correlation, while acting as a regularizer for the classification task over multiple visual domains}. 
Extensive experiments confirm our intuition and show that our multi-task method combining supervised and self-supervised knowledge shows competitive results with respect to more complex domain generalization and adaptation solutions. It also proves its potential in the novel and challenging predictive and partial domain adaptation scenarios. 
\end{abstract}

}

\maketitle

\IEEEdisplaynontitleabstractindextext

\IEEEpeerreviewmaketitle

\IEEEraisesectionheading{\section{Introduction}\label{sec:introduction}}
\label{intro}
\IEEEPARstart{M}any definitions of \emph{intelligence} have been formulated by psychologists and learning researches along the years. Despite the differences, they all indicate the \emph{ability to adapt and achieve goals under a wide range of conditions} as a key component \cite{wikipedia}. Artificial intelligence inherits these definitions, with the most recent research demonstrating the importance of knowledge transfer and domain generalization \cite{csurka_book}. Indeed, in many practical applications the underlying distributions of training (\ie source) and test (\ie target) data are inevitably different, asking for robust and adaptable solutions. When dealing with visual domains, most of the current strategies are based on \emph{supervised learning}. These processes search for semantic spaces able to capture basic data knowledge regardless of the specific appearance of input images: some decouple image style from the shared object content \cite{Bousmalis:DSN:NIPS16}, others generate new samples \cite{Volpi_2018_NIPS}, or impose adversarial conditions to reduce feature discrepancy \cite{Li_2018_CVPR,Li_2018_ECCV}. 
With the analogous aim of getting general purpose feature embeddings, an alternative research direction is pursued by \emph{self-supervised learning} that captures visual invariances and regularities solving tasks that do not need data annotation, like image orientation recognition \cite{gidaris2018unsupervised} or image coloring \cite{zhang2016colorful}. 
\begin{figure}[!t]
    \centering
\includegraphics[width=0.48\textwidth]{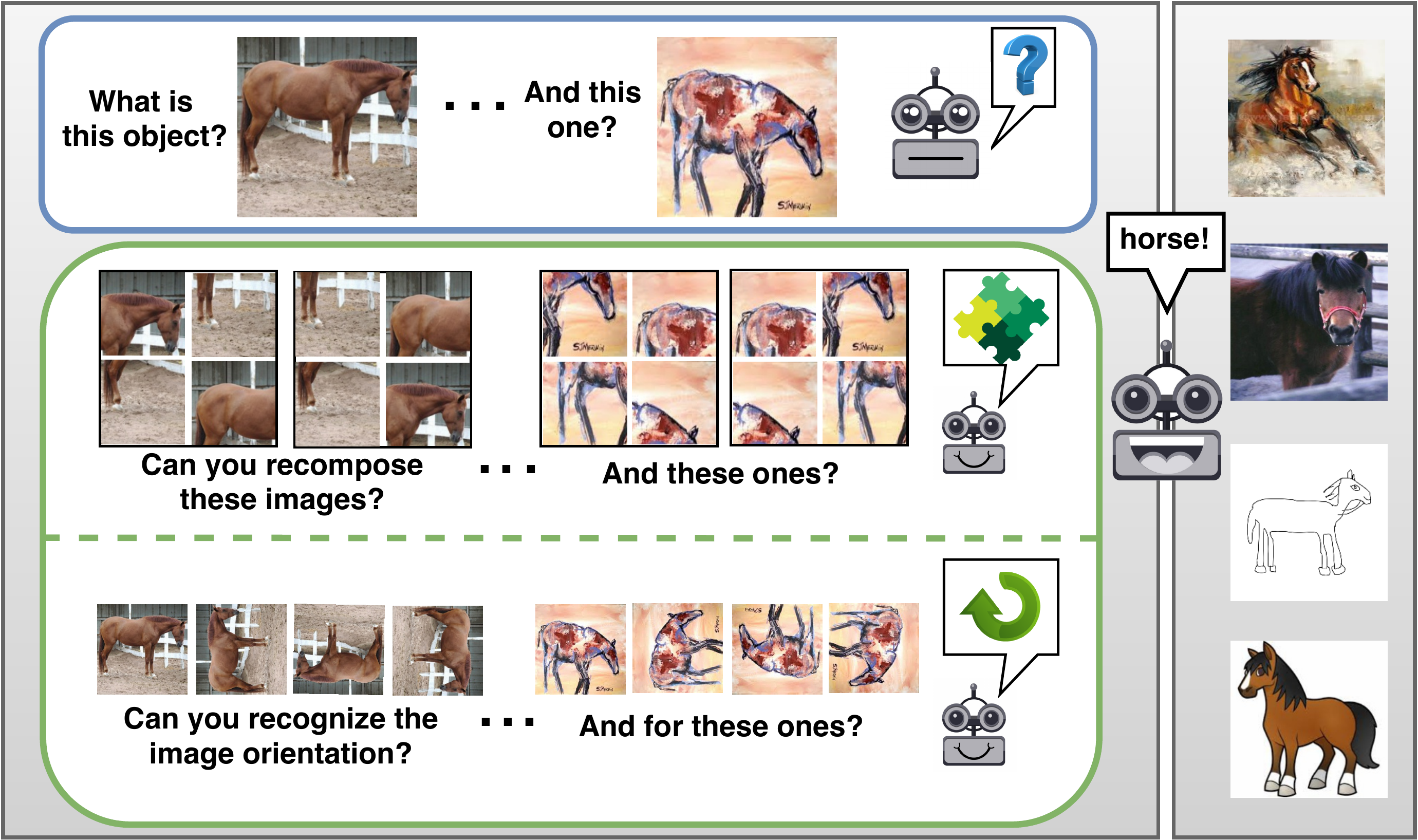}   \vspace{-2mm}
    \caption{Recognizing objects across visual domains is a challenging task that requires high generalization abilities.
   Self-supervisory image signals allow to capture natural invariances and regularities that can help to bridge across large style gaps. With our multi-task approach we learn jointly to classify objects and solve jigsaw puzzles or recognize image orientation, showing that this supports generalization to new domains.}
    \label{fig:copertina}\vspace{-3mm}
\end{figure}
\begin{figure*}[!t]
    \centering
\includegraphics[width=0.8\textwidth]{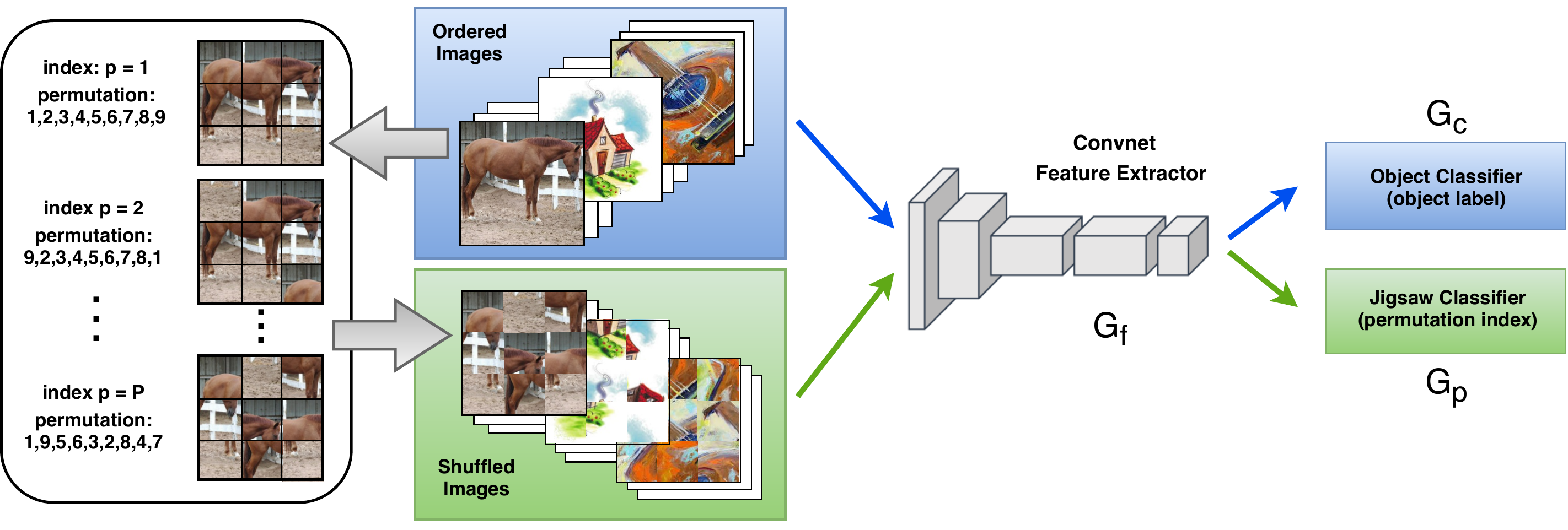}
    \caption{Illustration of the proposed multi-task approach when using jigsaw puzzle as self-supervised task. 
    We start from images of multiple domains and use a $3\times 3$ grid to decompose them in 9 patches
    which are then randomly shuffled and recomposed into images of the same dimension of the original ones.
    Through the maximal Hamming distance algorithm in \cite{NorooziF16} we define a set of $P$ patch permutations
    and assign an index to each of them. Both the original ordered and the shuffled images are fed to 
    a convolutional network that is optimized to satisfy two objectives: object classification on the 
    ordered images and jigsaw classification (\ie permutation index recognition) on the shuffled images. An analogous scheme holds when using rotation recognition as self-supervision. The names assigned to each network part refer to the notation adopted in Sec. \ref{method}.}
    \label{fig:jigen}\vspace{-3mm}
\end{figure*}
Unlabeled data are largely available and by their very nature are less prone to bias (no labeling bias issue 
\cite{TorralbaEfros_bias}), thus they seem the perfect candidate to provide visual information independent from specific domain styles. However their potential has not been fully 
exploited: the existing self-supervised approaches often come with tailored architectures
that need dedicated fine-tuning strategies to re-engineer the acquired knowledge \cite{Noroozi_2018_CVPR}. Moreover, they are mainly applied on real-world photos without considering cross-domains scenarios with images of paintings or sketches.

This clear separation between learning intrinsic regularities from images (self-supervised knowledge) and robust classification across domains (supervised knowledge) 
is in contrast with the visual learning strategies of biological systems, and in particular of the human visual system. 
Indeed, numerous studies highlight that infants and toddlers learn both to categorize objects and about regularities at 
the same time \cite{children_learning}. For instance, popular toys for infants teach to recognize different categories by fitting them 
into shape sorters; jigsaw puzzles of animals or vehicles to encourage learning of object parts' spatial relations 
are equally widespread among 12-18 months old. 
This joint learning is certainly a key ingredient in the ability of humans to reach sophisticated visual generalization abilities at an early age \cite{PLOS}. 

Inspired by this, our original paper \cite{jigsawCVPR19} was the first to introduce a multi-task approach that learns simultaneously how to recognize objects by exploiting supervised data, and how to generalize to new domains by leveraging intrinsic self-supervised information about spatial co-location of image parts (Fig. \ref{fig:copertina} and \ref{fig:jigen}). Specifically, we proposed to recover an original image from its shuffled parts, re-purposing the popular game of \emph{solving jigsaw puzzles}. 
Differently from previous approaches that deal with feature extraction from separate image patches \cite{NorooziF16,Noroozi_2018_CVPR}, we moved the patch re-assembly at the 
image level and we formalized the jigsaw task as a classification problem over recomposed images with the same dimension of the original one. In this way object recognition and patch reordering can share the same network backbone and we 
can seamlessly leverage over any convolutional learning structure as well as several pretrained models without the need of specific architectural changes.

Here we  extend our previous work providing a wider overview on self-supervised learning across domains. \textbf{(1)} We consider rotation recognition and jigsaw puzzle as self-supervised tasks showing their effect both as pretext and in the multi-task model together with supervised learning for domain generalization; \textbf{(2)} we delve into the details of the multi-task method with an extensive ablation analysis and visualizing successful as well as failure cases; \textbf{(3)} we consider both single source and multi-source domain adaptation experiments with a thorough analysis against the most recent state-of-the art methods; \textbf{(4)} we discuss the effect of our multi-task model in the challenging predictive and partial domain adaptation scenarios also extending \cite{tackling_iciap19}.

\section{Related Work}
\label{related}
\textbf{Self-Supervised Learning.}
Self-Supervised Learning is a paradigm developed to learn visual features from large-scale unlabeled data \cite{SSLsurvey}. 
Its first step is a \emph{pretext} task that exploits inherent data attributes
to automatically generate surrogate labels: part of the existing knowledge about the images is manually removed (\eg the color, the orientation, the patch order) and the task consists in recovering it.
It has been shown that the first layers of a network trained in this way capture useful semantic knowledge \cite{asano20a-critical}. The second step of the learning process consists in transferring the self-supervised learned model of those initial layers to a supervised \emph{downstream} task (\eg classification, detection), while the ending part of the network is newly trained.

The possible pretext tasks can be organized in three main groups. 
One group relies only on original visual cues and involves either the whole image with
geometric transformations (\eg translation, scaling, rotation \cite{gidaris2018unsupervised,NIPS2014_geometric}), 
clustering  \cite{caron2018deep}, inpainting \cite{pathakCVPR16context} and colorization \cite{zhang2016colorful},
or considers image patches focusing on their equivariance (learning to count \cite{learningtocount})
and relative position (solving jigsaw puzzles \cite{NorooziF16,Noroozi_2018_CVPR}). 
A second group uses external sensory information either real or synthetic: this solution
is often applied for multi-cue (visual-to-audio \cite{audiovisual}, RGB-to-depth \cite{ren-cvpr2018}) 
and robotic data \cite{grasp2vec, visiontouch}.
Finally, the third group relies on video and on the regularities introduced by the temporal dimension 
\cite{videosiccv15,SSLvideo}.
The most recent self-supervised learning research focuses on 
proposing novel pretext tasks or combining several of them together, to then compare their initialization performance for a downstream task with 
respect to using supervised models as in standard transfer learning  
\cite{gidaris2020learning,jenni2020steering,multitaskSSL,ren-cvpr2018}. 

Our work investigates a new research direction: we combine supervised and self-supervised knowledge in a multi-task framework, studying its effect on domain generalization and adaptation.

\textbf{Domain Generalization and Adaptation.}
Several algorithms have been developed to cope with domain shift, mainly in two different settings: \emph{Domain Generalization} (DG) and \emph{Domain Adaptation} (DA).
In DG the target is unknown at training time: the learning process can usually leverage multiple labeled sources to define a model robust to any new, previously unseen domain \cite{shallowDG}.
In DA the learning process has access to the labeled source data and to the unlabeled target data, so the aim is to generalize to the given specific target set \cite{csurka_book}. 
In multi-source DA the source domain label may be unknown \cite{mancini2018boosting, hoffman_eccv12,carlucci2017auto}, while for most of the DG methods it remains a crucial information to leverage on. 

There are three main families of solutions for both DG and DA.
\emph{Feature-level} strategies focus on learning domain invariant data representations mainly by minimizing 
different domain shift measures \cite{Long:2015,LongZ0J17,dcoral,hdivergence}. The domain shift can also be 
reduced by training a domain classifier and inverting the optimization to guide the features towards maximal domain confusion \cite{Ganin:DANN:JMLR16,Hoffman:Adda:CVPR17}. This adversarial approach has several variants, some of which also exploit class-specific domain recognition modules \cite{saito2017maximum,Li_2018_ECCV}. Metric learning \cite{doretto2017}  and deep autoencoders \cite{DGautoencoders,Li_2018_CVPR,Bousmalis:DSN:NIPS16} have also been used to search for domain-shared embedding spaces. In DG, these approaches leverage on the availability of multiple sources and on the access to the domain label for each sample. 
\emph{Model-level} strategies either change how the data are loaded with ad-hoc episodes \cite{episodic_hospedales}, or modify conventional learning algorithms to search for more robust minima of the objective function \cite{MLDG_AAA18}. Besides these main approaches, other solutions consists in introducing domain alignment layers \cite{carlucci2017auto}, aggregation layers \cite{Antonio_GCPR18,episodic_hospedales}, or using low-rank network parameter decomposition \cite{hospedalesPACS,Ding2017DeepDG} with the goal of identifying and neglecting domain-specific signatures.
Finally, \emph{data-level} techniques exploit variants of the Generative Adversarial Networks (GANs, \cite{Goodfellow:GAN:NIPS2014}) to synthesize new images. 
Indeed, producing source-like target images or/and target-like source images \cite{russo17sbadagan,cycada} help to reduce the domain gap.

Some recent works have started investigating intermediate settings between DA and DG. In \emph{Predictive Domain Adaptation} (PrDA)
a labeled source and several auxiliary unlabeled domains are available at training time together with meta-data that describe their relation \cite{adagraph,multivariatereg}.
The target data are not available, but their meta-data are provided and used to compose an adapted model directly from the sources.

In both DA and DG, the main assumption is that source and target share the 
same label set, with few works studying exceptions to this basic condition 
\cite{PADA_eccv18, cocktail_CVPR18,Saito_2018_ECCV}. 
In particular, in \emph{Partial Domain Adaptation} (PDA) the target to covers only a subset of the source class set. In this case it is important to adjust the adaptation process so that the samples with not-shared labels would not influence the learned model. The more commonly used techniques consist in adding a \emph{re-weight source sample strategy} to a standard DA approach \cite{PADA_eccv18,SAN,IWAN}. Alternative solutions leverage on two separate deep classifiers and their prediction inconsistency on the target \cite{TWIN_PDA} or on feature norm matching \cite{featurenorm_PDA}.

As indicated by this brief overview, previous literature did not investigate self-supervision for DA or DG. In this work we present a thorough study of self-supervised learning across domains.

\section{Method}
\label{method}
We introduce here the technical notation for our multi-task approach across domains and specify the objectives in each of the considered settings. 
Let us assume to observe data $\{(\bx_i^s,\by_i^s)\}_{i=1}^{n^s}$ from one or more source distributions. 
Here $\bx_i^s$ represent the $i$-th image while 
$\by_i^s$ is the corresponding one-hot vector label of dimension $|\mathcal{Y}^s|$.
Starting from these images we can always apply different procedures to generate self-supervised variants. One simple choice is that of applying rotation to produce 4 copies of each sample with \{$0^\circ$, $90^\circ$, $180^\circ$, $270^\circ$\} orientation. The related self-supervised task consists in choosing the correct image rotation. 
A more structured alternative is that of decomposing the original images according to a $3\times 3$ grid: this produces $9$ squared patches from every sample, which are then moved from their original locations and re-positioned to form a set of $9!$ shuffled images. {This task is reminiscent of the jigsaw puzzle game}, where the tiles have to be rearranged to get back the original image.
For both the described cases,  $\{(\bz_k^s,\bp_k^s)\}_{k=1}^{K^s}$ refer to the newly obtained images. The dimension of the one-hot vector label $\bp$ is $4$ when applying rotation, while for patch shuffling we choose a subset $P$ of the $9!$ possible permutations selected by following the Hamming distance based algorithm in \cite{NorooziF16}. The total number of images changes depending on the self-supervised task:  $K^s=4\times n^s$ for rotation and  $K^s=P\times n^s$ for patch shuffling.
Regardless of the specific chosen self-supervised objective we can combine it with supervised learning through a 
standard hard-parameter sharing 
multi-task model realized with a multi-branch ending network  \cite{Caruana:1997}. One output branch will be dedicated to the supervised task exploiting the labels of the source data, while the other will solve the self-supervised problem: rotation or jigsaw puzzle permutation recognition (see Figure \ref{fig:jigen}). 
{The auxiliary self-supervised objective contributes in extracting relevant semantic features from the data, with a final beneficial effect on the object recognition performance. 
Since the self-supervised objective is label agnostic it can run both over supervised and unsupervised domains, supporting generalization and adaptation.
}

\subsection{Domain Generalization}
For our network we indicate the convolutional feature extraction backbone with $G_f$, parametrized by $\theta_f$. The parameters of the object classifier $G_c$ and of the self-supervised task $G_p$ are respectively $\theta_c$ and $\theta_p$. Overall we train the network to obtain the optimal model through \vspace{-3mm}
\begin{align} \small
    \arg \min_{\theta_f, \theta_c, \theta_p} ~~  & 
    \frac{1}{n^s} \sum_{i=1}^{n^s}   \mathcal{L}_c(G_c(G_f(\bx^s_i)),\by_i^s) +  \nonumber \\
    & \alpha^s \frac{1}{K^s}\sum_{k=1}^{K^s}\mathcal{L}_p(G_p(G_f(\bz^s_k)),\bp_k^s)
    \label{equationDG}\vspace{-5mm}
\end{align}
where $\mathcal{L}_c$ and $\mathcal{L}_p$ are cross entropy losses for both the object and self-supervised classifiers. We underline that the self-supervised loss is also calculated on the original images. Indeed, the $0^\circ$ orientation as well as the correct patch sorting correspond
also to one of the possible self-supervised image transformation variants. Differently, the supervised classification loss is not influenced by the shuffled or rotated images, as this would make object recognition tougher. At test time we use the object classifier $G_c$ to predict on the new target images.

\subsection{Domain Adaptation}
By its nature self-supervised learning does not need manual annotation and it can exploit the unlabeled target data $\{{\bx}_j^t\}_{j=1}^{n^t}$ when available in the DA setting.
The target samples are transformed (rotated, shuffled) so that each newly produced instance $\{{\bz}_k^t\}_{k=1}^{K^t}$ gets its own self-supervised label ${\bp}_k^t$.

{An alternative and widely used way 
to involve the target data in the learning process consists in applying the source supervised knowledge on them to evaluate the pseudo-labels $\hat{{\by}}^t=G_c(G_f({\bx}^t))$, and minimize the prediction uncertainty measured by the entropy  $H=-\sum_{l=1}^{|\mathcal{Y}^s|}\hat{\by}^t_{l} \log \hat{\by}^t_{l}$ ~\cite{mancini2018boosting,featurenorm_PDA}. This is a semi-supervised technique which guides the class decision boundary to pass through 
low-density target areas, but its success across domains depends on moderate levels of domain shift to avoid wrong pseudo-labels.
}
{Given their orthogonal and possibly complementary nature, in our DA analysis we combine the entropy term with the supervised and self-supervised loss.}
The overall learning objective is formalized as \vspace{-1mm}
\begin{align} \small
    \arg \min_{\theta_f, \theta_c, \theta_p} ~~  
    \frac{1}{n^s} \sum_{i=1}^{n^s}   & \mathcal{L}_c(G_c(G_f(\bx^s_i)),\by_i^s) + \nonumber \\
    \alpha^s \frac{1}{K^s}\sum_{k=1}^{K^s} & \mathcal{L}_p(G_p(G_f(\bz^s_k)),\bp_k^s) +  \nonumber \\
    \eta~\frac{1}{n^t} \sum_{j=1}^{n^t}  H(G_c(G_f(\bx^t_j))) + 
   \alpha^t \frac{1}{K^t} \sum_{k=1}^{K^t} & \mathcal{L}_p(G_p(G_f(\bz^t_k)),\bp_k^t) .
    \label{equationDA}
\end{align}

\subsection{Partial Domain Adaptation}
\label{subsec:PDA}
In PDA the label space of the target domain is contained in that of the source domain $\mathcal{Y}^t \subseteq \mathcal{Y}^s$. This further shift in the label space makes the problem even more challenging: 
if the matching between the whole source and target data is forced, any adaptive method may incur in a degenerate case producing worse performance than its plain non-adaptive version due to negative transfer \cite{Rosenstein05totransfer}. 

The two $\mathcal{L}_p$ terms in (\ref{equationDA}) help domain shift reduction, however their co-presence may be redundant: the features are already chosen to minimize the 
source classification loss and the self-supervised task on the target back-propagates inducing a cross-domain adjustment on the learned features. Thus, for PDA we can drop the source self-supervised term, which corresponds to setting $\alpha^s=0$. This choice has a double positive effect: on one side it reduces the number of hyper-parameters in the learning process, leaving space for the introduction of other complementary learning conditions, on the other we let the self-supervised module focus only on the target without involving the extra classes of the source. 

\begin{figure}[!t]
  \centering
\includegraphics[width=0.5\textwidth]{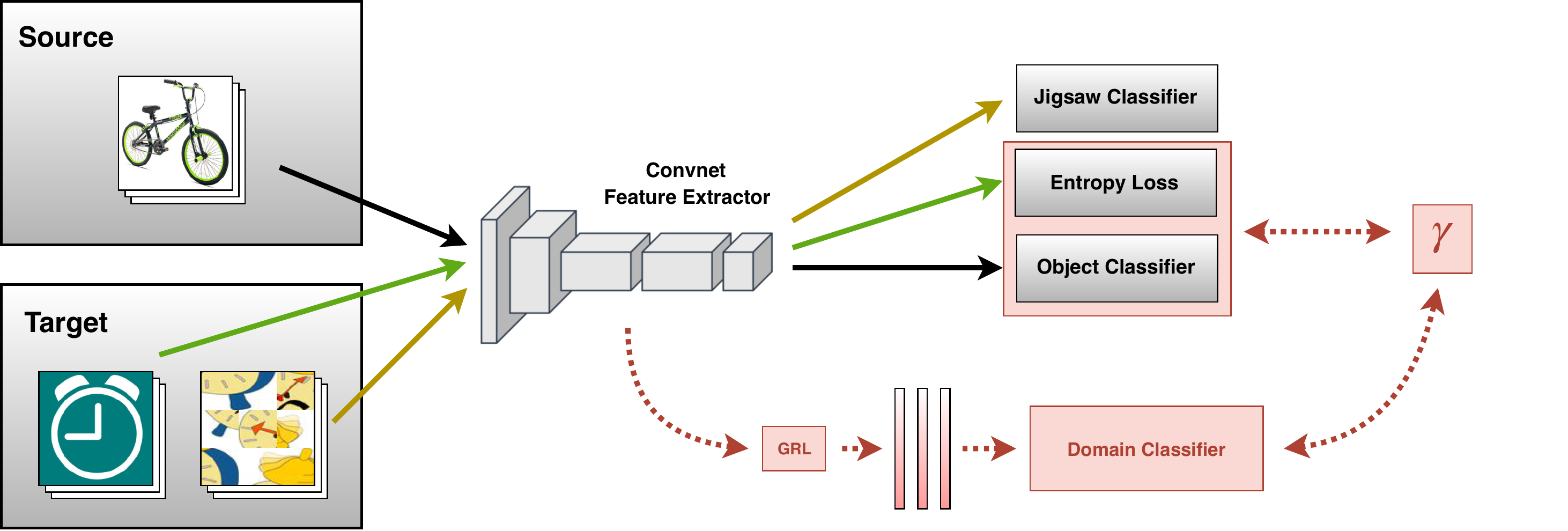}
\caption{Our PDA approach with jigsaw puzzle self-supervision. The main blocks 
of the network are in gray. The solid line arrows indicate the contribution of each group of training samples
to the corresponding final tasks. The related optimization goals appear at the end of the black/green/ocher arrows. The red blocks illustrate the domain adversarial 
classifier and source sample weighting procedure (weight $\gamma$). 
An analogous scheme holds with self-supervised rotation recognition.}\vspace{-4mm}
\label{fig:scheme}
\end{figure}

To further enforce the focus on the shared classes, we extend our approach by integrating a weighting mechanism analogous to that presented in \cite{PADA_eccv18}. The source classification output on the target data are accumulated with 
$\boldsymbol{\gamma} = \frac{1}{n^t}\sum_{j=1}^{n^t}\hat{\by}_j^t$
and normalized as $\boldsymbol{\gamma} \leftarrow \boldsymbol{\gamma}/ \max(\boldsymbol{\gamma})$, obtaining a $|\mathcal{Y}^s|$-dimensional vector that quantifies the contribution of each source class. 
Moreover, we can easily integrate a \emph{source vs target domain discriminator} $G_d$ as in \cite{Ganin:DANN:JMLR16} and adversarially maximize the related binary cross-entropy to increase the domain confusion, taking also into consideration the defined class weighting 
procedure for the source samples. In more formal terms, the final objective of our multi-task problem in the PDA setting is \vspace{-2mm}
{
\begin{align} \small 
    \arg \min_{\theta_f, \theta_c, \theta_p} \max_{\theta_d} 
    \frac{1}{n^s} \sum_{i=1}^{n^s} {\gamma_{y_i}} \Big( \mathcal{L}_c(G_c(G_f({\bx}^s_i),\by_i^s)  + \nonumber \\[-5pt] 
    \lambda \log (G_d(G_f({\bx}^s_i))) \Big)+  \nonumber \\[-5pt]
    \frac{1}{n^t} \sum_{j=1}^{n^t} \Big( \eta ~   H(G_c(G_f({\bx}^t_j))) +  
    \lambda \log (1 - G_d(G_f({\bx}^t_j)))\Big) + \nonumber \\[-10pt]
    \alpha^t \frac{1}{K^t} \sum_{k=1}^{K^t} \mathcal{L}_p(G_p(G_f({\bz}^t_k)),\bp_k^t)
    \label{equationPDA}
\end{align}}
where {$\gamma_{y_i}$ is the class weight for the ground truth label of the source point ${\bx}^s_i$} and $\lambda$ is a hyper-parameter that adjusts the importance of the introduced domain discriminator. 
When $\lambda=0$ and $\gamma_{y} = 1/|\mathcal{Y}^s| $ we fall back to the standard DA case. 
A schematic illustration of the method is presented in Fig. \ref{fig:scheme}.

\subsection{Implementation details}
\label{subsec:implementation}
We designed our multi-task network to leverage over different convolutional deep architectures: the backbone $G_f$ may inherit the structure of standard networks as AlexNet or ResNet.  
The specific object and self-supervised classifiers heads $G_c,G_p$ are respectively implemented by an ending fully connected layer.
{When including  multiple self-supervised tasks in the model (\ie Jigsaw+Rotation), a $G_p$ head is assigned to each self-supervised objective. Specifically, shuffled images are directed to the Jigsaw final head $G_p^J$ , while the rotated images to the Rotation recognition head $G_p^R$.}
In the PDA setting we introduced the domain classifier $G_d$ by adding three fully connected layers after the last pooling layer of the main backbone, and using a sigmoid function for the last activation as in \cite{Ganin:DANN:JMLR16}. For all our experiments we trained the network end-to-end by fine-tuning all the feature layers from Imagenet pre-trained models \cite{imagenet}, while $G_c, G_p$ and $G_d$ are learned from scratch. 

Overall the network for DG has two main hyper-parameters: $\alpha$ that weights the self-supervised loss, and the data bias parameter $\beta$ which regulates the data input process. The self-supervised variants of the images enter the network together with the original ones, hence each image batch contains both of them with $\beta$ specifying their relative ratio. For instance $\beta=0.6$ means that for each batch, $60\%$ of the images are standard, while the remaining $40\%$ are rotated or composed of shuffled patches. {In our experiments we chose $\alpha$ and $\beta$ by keeping a source validation set ($10\%$ of the training data) and performing model selection on it by following \cite{anonymous2021in}. 
When combining Jigsaw+Rotation we have respectively $\alpha_J$ and $\alpha_R$, while the fraction of transformed images regulated by $\beta$ are rotated or shuffled with equal probability.
In the DA setting $\alpha$ decouples in $\alpha^s$ and  $\alpha^t$ respectively for source and target data. While discussing the experimental results we will see the outcome of cross-validating $\alpha$ on the source and then setting $\alpha=\alpha^s=\alpha^t$ or fixing $\alpha^s=0$ as well as the effect on model robustness when manually tuning $\alpha^t$. 
Further parameters in DA and PDA are $\eta$ and  $\lambda$. The first is the weight assigned to the entropy loss which we safely fixed to small values: $0.1$ for DA and $0.2$ for PDA. Finally, $\lambda$ balances the importance of the gradient reversal layer when included in PDA and we adopted the same scheduling of \cite{Ganin:DANN:JMLR16} to update its value, so that the importance of the domain discriminator increases with the training epochs.}

In designing the jigsaw puzzle task we need to choose the image patch grid size $n \times n$, and the cardinality of the patch permutation subset $P$. 
As we will detail in the following section, our multi-task approach is robust to these values and for all our experiments we kept them fixed ($3 \times 3$ grid, $P=30$).

We used a simple data augmentation protocol by randomly cropping the images to retain between $80-100\%$ and randomly applied horizontal flipping. By following \cite{Noroozi_2018_CVPR}, we also randomly ($10\%$ probability) convert an image tile to grayscale.
Our DG/DA model is trained with an SGD solver, $30$ epochs, batch size $128$,  learning rate set to $0.001$ and stepped down to $0.0001$ after $80\%$ of the training epochs.
Our PDA model is trained with 
SGD with momentum set at $0.9$, weight decay  $0.0005$ and $24$ epochs. We used batch size of 64 and initial learning rate $0.0005$.
Some specific training details are used in the PrDA setting and will be  described in Sec. \ref{sec:exp_DG_6}.
We implemented our deep methods in PyTorch and the code is available at \url{https://github.com/silvia1993/Self-Supervised_Learning_Across_Domains}.

\section{Experiments}
\label{experiments}
In this section we present an extensive evaluation of using self-supervised knowledge across visual domains. 
\textbf{First of all we focus on DG (Sec. \ref{sec:exp_DG})}.
We test both the rotation and jigsaw puzzle self-supervised pretexts before using them extensively as auxiliary tasks together with supervised learning in our multi-task model.
\textbf{The second part of our analysis is dedicated to the DA scenario
(Sec. \ref{sec:exp_DA})} and its more challenging PDA setting. 
\subsection{Self-Supervision for Domain Generalization}
\label{sec:exp_DG}\vspace{-7mm}
\black{\subsubsection{Data and Setup}
For our DG analysis we used as main testbed the PACS dataset \cite{hospedalesPACS} that covers $7$ object categories and $4$ domains (Photo, Art Paintings, Cartoon and Sketches). We followed the experimental protocol in \cite{hospedalesPACS} and finetuned the Imagenet-pretrained models with three domains as source datasets and the remaining one as target test. We also considered other two data collections. The VLCS \cite{TorralbaEfros_bias} aggregates images of $5$ object categories shared by the PASCAL VOC 2007, LabelMe, Caltech and Sun datasets. 
We followed the standard protocol of \cite{DGautoencoders} dividing each domain  into a training set ($70\%$) and a test set ($30\%$) 
by random selection from the overall dataset. The Office-Home dataset \cite{venkateswara2017Deep} contains 65 categories of daily objects from 4 domains: Art, Clipart, Product and Real-World. 
For this dataset we used the same experimental protocol of \cite{Antonio_GCPR18}. Note that Office-Home and PACS are related in terms of domain types and it is useful to consider
both as testbeds to check if our multi-task self-supervised approach scales when the number of categories changes from 7 to 65. Instead VLCS offers different challenges by combining object categories from Caltech with scene images of the other domains. 
The evaluation is based on three repetitions of each run: we report the average $\pm$ standard deviation of the obtained class recognition accuracy.}

{Only for the single-source DG analysis we focused on 
digits datasets
to compare the sensitivity of our approach against a competitor method. For PrDA we considered a fine-grained car dataset.
All the details for these last two settings are described respectively in Sec. \ref{sec:exp_DG_3} and \ref{sec:exp_DG_6}.
}

\subsubsection{Self-Supervised Pretraining}
\label{sec:exp_DG_1.1}
We test here the robustness of image orientation and patch co-location knowledge across domains by using both rotation and jigsaw puzzle as pretext tasks for domain generalization. 

\noindent\emph{Baselines.} As first step we considered three jigsaw puzzles and one rotation model trained on Imagenet (ILSVRC12, \cite{imagenet}) data without original labels. For the jigsaw puzzle, we used the two Context-Free-Network (CFN) models provided by the authors of \cite{NorooziF16,Noroozi_2018_CVPR}. The CFN has 9 AlexNet-based siamese branches that extract features separately from each image patch and then recompose them before entering the final classification layer. 
We indicate these models respectively as J-CFN \cite{NorooziF16} and J-CFN+ \cite{Noroozi_2018_CVPR}. The third puzzle-based model is obtained by training an AlexNet on whole images recomposed from disordered patches, which we call J-AlexNet. 
Inspired by \cite{gidaris2018unsupervised}, we also trained an AlexNet model for rotation recognition that we dub R-AlexNet.

\noindent\emph{Results.} The obtained results are collected in the top part of Table \ref{table:resultsDG_PACS_one} and show that using a patch-based (p) jigsaw method provides  on average  a more reliable pretext model than dealing with the whole (w) recomposed image. 
The rotation pretext model shows the best results with a small advantage over the patch based jigsaw approaches. In summary, we find that moving the jigsaw puzzle task from the feature to the image level when training a pretext model does not appear as a good choice and that the rotation task is the simplest and most effective solution.

\begin{table}[tb]
\begin{center} \small
\caption{Test on different tasks and architectures: DG classification accuracy. 
The target is indicated as column title. Best results are in bold. 
\emph{Top}: self-supervised pretraining on Imagenet, followed by fine-tuning on the source. (p) indicates the methods that use patch-based networks, while (w) the ones that use whole-images networks.
\emph{Bottom}: supervised pretraining on Imagenet followed by the multi-task combination of self-supervised objective and supervised fine-tuning.}\vspace{-2mm}
\begin{tabular}{@{}c@{}c@{}c@{~~}c@{~~}c@{~~}c|@{~~}c}
\hline
\multicolumn{2}{c}{\textbf{PACS}}  & \textbf{art\_paint.} & \textbf{cartoon} &  \textbf{sketches} & \textbf{photo} &   \textbf{Avg.}\\ \hline
\multicolumn{7}{@{}c@{}}{{Self-Supervised Pretraining}}\\
\hline
\multicolumn{2}{@{}c@{}}{\footnotesize{J-CFN (p)}} & 47.23  & 62.18  & 58.03  & 70.18  & 59.41\\
\multicolumn{2}{@{}c@{}}{\footnotesize{J-CFN+ (p)}} &  51.14  &  58.83  &  54.85  &  73.44  &  59.57\\
\multicolumn{2}{@{}c@{}}{\footnotesize{J-AlexNet (w)}} & 38.93 & 53.75 & 49.00 & 64.23 & 51.48   \\ 
\multicolumn{2}{@{}c@{}}{\footnotesize{R-AlexNet (w)}} & 52.08 & 59.24 & 56.54 & 72.91 & \textbf{60.19}  \\
\hline 
\multicolumn{7}{@{}c@{}}{{Supervised Pretraining and  Multitask}}\\
\hline
\multicolumn{2}{@{}c@{}}{\footnotesize{C-CFN-DeepAll (p)}} &  59.69  &  59.88  &  45.66  &  85.42  &  62.66\\
\multicolumn{2}{@{}c@{}}{\footnotesize{C-CFN-Jigsaw (p)}} &  60.68  &  60.55  &  55.66  &  82.68  &  {64.89}\\
\multicolumn{2}{@{}c@{}}{\footnotesize{AlexNet-DeepAll (w)}} & 66.50  & 69.65  & 61.42  & 89.68 &  71.81\\
\multicolumn{2}{@{}c@{}}{\footnotesize{AlexNet-Jigsaw (w)}} & 67.79 & 70.79 & 64.01 & 89.64 & 73.05\\
\multicolumn{2}{@{}c@{}}{\footnotesize{AlexNet-Rotation (w)}} & 69.43  & 69.40  & 65.20 &  89.17  & \textbf{73.30}\\
\hline
\end{tabular}
\label{table:resultsDG_PACS_one}
\end{center}\vspace{-5mm}
\end{table}

\subsubsection{Supervised Pretraining and Multi-task Learning}
\label{sec:exp_DG_1.2}
In designing our multi-task approach which combines supervised and self-supervised learning we have several options, both in terms of the architecture to use and of the best self-supervised task. 

\noindent\emph{Baselines.} We compare the CFN multi-branch architecture with a plain AlexNet backbone. 
To differentiate the classification-aware CFN model with respect to the self-supervised pretraining discussed in the previous section we name it C-CFN. 
Regardless of the specific architecture used, {we indicate with \emph{DeepAll} the single-task supervised model trained on all the original source images (\ie $\alpha=0$)}, while we use  \emph{Jigsaw (Puzzle)} or \emph{Rotation} to specify the multi-task case where each of those self-supervised tasks was trained jointly with the object classification.

\noindent\emph{Results.} From the results in the bottom part of Table \ref{table:resultsDG_PACS_one} we can draw two conclusions. First, combining supervised and self-supervised learning provides better results than a single-task supervised model across domains. This is true regardless of the chosen architecture, as indicated by the comparison between the DeepAll and Jigsaw/Rotation variants. Second, a single branch architecture is better suited for the multi-task problem at hand. In this case, moving the jigsaw puzzle task from feature to image level simplifies the self-supervised task and its combination with the supervised objective. The whole-image Rotation auxiliary task supports generalization even slightly better than Jigsaw.

\subsubsection{Multi-Source Domain Generalization}
\label{sec:exp_DG_2}
Here we provide an extensive evaluation of our multi-task approach against state-of-the-art multi-source DG methods.

\noindent\emph{Baselines:}
We consider different families of DG approaches\footnote{We are aware of recent DG solutions based on data augmentation. In \cite{zhang2020learning},  MSCOCO (\url{http://cocodataset.or}) and WikiArt (\url{https://www.kaggle.com/c/painter-by-number}) are used for style transfer. None of the other considered references exploit those extra data collections so we do not include this method.}. 
The first is based on low-rank constraints applied on network parameters: {TF} \cite{hospedalesPACS}, {SLRC} \cite{Ding2017DeepDG}. The second exploits domain-specific component aggregation: {Epi-FCR} \cite{episodic_hospedales}, {D-SAM} \cite{Antonio_GCPR18}.
The third builds on meta-learning strategies: {MLDG} \cite{MLDG_AAA18}, {MetaReg} \cite{NIPS2018_metareg}, {MASF} \cite{dou2019domain}.
Finally, the fourth family leverages adversarial classifiers in different ways: {DDAIG} \cite{zhou2020deep}, {PAR} \cite{wang2019learning}, {MMLD} \cite{dg_mmld}. 
{We carefully report the DeepAll reference for each method to have an overview on their relative advantage\footnote{
The differences between the DeepAll results, are likely due to small undocumented inconsistencies and/or different library implementations of these baseline methods. Reporting them all is the only fair way of showing the relative improvement brought by each approach and highlighting possible inconsistencies.}.}

\begin{table}[tb]
\begin{center} \small
\caption{Comparison with DG-sota methods on PACS.
The target is indicated as column title.
We report the used hyper-parameters, obtained through source cross-validation. The top result is highlighted in bold.
Only to get fair comparison with MASF we computed the max target accuracy over the training period: the results are indicated with $\diamond$. The top result in this case is underlined.
}
\vspace{-5mm}
\resizebox{\columnwidth}{!}{\begin{tabular}{@{}c@{~~~}c@{~~~}c@{~~~}c@{~~~}c@{~~~}c|@{~~~}c}
\hline
\multicolumn{2}{c}{\textbf{PACS}}  & \textbf{art\_paint.} & \textbf{cartoon} &  \textbf{sketches} & \textbf{photo} &   \textbf{Avg.}\\ \hline
\multicolumn{7}{c}{\textbf{Alexnet}}\\
\hline
\multirow{2}{*}{\cite{hospedalesPACS}}  & DeepAll & 63.30 & 63.13 & 54.07 & 87.70 & 67.05\\
& TF & 62.86 & 66.97 & 57.51  & 89.50 & 69.21\\
\hline
\multirow{2}{*}{\cite{Antonio_GCPR18}} & DeepAll  & 64.44 & 72.07 & 58.07 &  87.50 & 70.52 \\
& D-SAM  & 63.87 & 70.70 & 64.66 & 85.55 & 71.20\\
\hline
\multirow{2}{*}{\cite{episodic_hospedales}}  & DeepAll & 63.40  & 66.10  & 56.60  & 88.50  &   68.70\\
& Epi-FCR &  64.70  &  72.30  &  65.00  & 86.10  &     72.00\\
\hline
\multirow{2}{*}{\cite{MLDG_AAA18}} & DeepAll & 64.91 & 64.28 & 53.08 & 86.67 & 67.24\\
 & MLDG   & 66.23 & 66.88 & 58.96 &  88.00& 70.01\\
\hline
\multirow{2}{*}{\cite{NIPS2018_metareg}}  & DeepAll &  67.21  &  66.12  &  55.32  & 88.47   &  69.28  \\
&  MetaReg &  69.82   &  70.35 &  59.26   &  91.07 &  72.62   \\
\hline
\multirow{2}{*}{\cite{wang2019learning}} & DeepAll  & 63.30 & 63.10 & 54.00 & 87.70 & 67.03 \\
& {{PAR}}  & 68.70 & 70.50 & 64.60 & 90.40 & 73.54\\
\hline
\multirow{2}{*}{\cite{dg_mmld}} & DeepAll  & 68.09 & 70.23 & 61.80 & 88.86 & 72.25 \\
& {{MMLD}}  & 66.99 & 70.64 & 67.78 & 89.35 & {73.69}\\
\hline
\multicolumn{2}{c}{DeepAll} & 66.50 & 69.65 & 61.42 & 89.68  & 71.81$\pm$0.26\\
\multicolumn{2}{c}{{{Jigsaw}$_{\alpha=0.9,\beta=0.6}$}} & 67.76 & 70.79 & 64.01 & 89.64 & 73.05$\pm$0.20\\
\multicolumn{2}{c}{{{Rotation}$_{\alpha=0.4,\beta=0.4}$}} & 69.43  & 69.40  & 65.20  &  89.17  & 73.30$\pm$0.47\\
\multicolumn{2}{c}{{{Jigsaw+Rotation$_{\alpha_{J}=0.9, \alpha_{R}=0.9, \beta=0.4}$}}} & 69.70 & 71.00  & 66.00  &  89.60  & \textbf{74.08}$\pm$0.32\\
\hline
\multirow{2}{*}{\cite{dou2019domain}}  & DeepAll$^\diamond$ & 67.60   & 68.87   & 61.13  &  89.20  &  71.70 \\
& MASF$^\diamond$ & 70.35  & 72.49   & 67.33   &  90.58  &  75.21 \\
& {Jigsaw}$^\diamond$ & 69.76 & 72.27 & 66.41 & 90.97 & 74.86$\pm$0.64\\
& {Rotation}$^\diamond$ & 69.80  & 71.10  & 66.57  & 90.13 & 74.40$\pm$0.67\\
& {Jigsaw+Rotation}$^\diamond$ & 70.23 & 73.33  &  67.23 & 90.40 & \underline{75.30}$\pm$0.50\\
\hline
\multicolumn{7}{c}{\textbf{Resnet-18}}\\
\hline
 \multirow{2}{*}{\cite{Antonio_GCPR18}} & DeepAll & 77.87 & 75.89 & 69.27 &  95.19 & 79.55\\
 &  D-SAM & 77.33 & 72.43 & 77.83 & 95.30 & 80.72\\
\hline
\multirow{2}{*}{\cite{episodic_hospedales}}  & DeepAll & 77.60  & 73.90  & 70.30  & 94.40  &  79.10\\
&  Epi-FCR  &  82.10  &  77.00  &  73.00  &  93.90  &   81.50\\
\hline
\multirow{2}{*}{\cite{NIPS2018_metareg}}  & DeepAll & 79.90   & 75.10   & 69.50  &  95.20  &  79.90 \\
& MetaReg & 83.70   &  77.20   &  70.30  & 95.50  &  81.70  \\
\hline
\multirow{2}{*}{\cite{zhou2020deep}} & DeepAll  & 77.00 & 75.90 &69.20  &96.00  &  79.50 \\
& {{DDAIG}}  & 84.20 & 78.10 & 74.70 &  95.30 & \textbf{83.10}\\
\hline
\multirow{2}{*}{\cite{dg_mmld}} & DeepAll  & 78.34 & 75.02 & 65.24 & 96.21 & 78.70 \\
 & {{MMLD}}  & 81.28 & 77.16 & 72.29 & 96.09 & 81.83\\
\hline
\multicolumn{2}{c}{DeepAll} & 77.83  & 74.26  & 65.81  & 95.71  & 78.40$\pm$0.28 \\
\multicolumn{2}{c}{{{Jigsaw}$_{\alpha=0.7, \beta=0.9}$}}    & 79.28  & 75.74  & 68.31  & 95.71 &  79.80$\pm$0.55\\
\multicolumn{2}{c}{{{Rotation}$_{\alpha=0.8, \beta=0.4}$}}    &  81.07  &  74.13  & 76.17  &  96.10  &  {81.87}$\pm$0.49 \\
\multicolumn{2}{c}{{{Jigsaw+Rotation}$_{\alpha_{J}=0.7,\alpha_{R}=0.7, \beta=0.8}$}}    &  81.07  & 73.97  & 74.67  & 95.93  & 81.41$\pm$0.50 \\
\hline
\multirow{2}{*}{\cite{dou2019domain}}  & DeepAll$^\diamond$ & 77.38  & 75.68    & 69.64  &  94.35  &  79.26 \\
& MASF$^\diamond$ & 80.29  & 77.17   & 71.69   & 94.99  &  81.04 \\
 & {Jigsaw}$^\diamond$    & 80.00  & 76.52 & 70.70  & 96.03 &  80.81$\pm$0.31\\
 & {Rotation}$^\diamond$    & 82.40  & 75.27  & 77.20 & 96.53  &  \underline{82.85}$\pm$0.55 \\
 & {Jigsaw+Rotation}$^\diamond$ & 81.40   & 75.03   & 76.47  & 96.40  &   82.33$\pm$0.47 \\
\hline
\end{tabular}}
\label{table:resultsDG_PACS}
\end{center}\vspace{-5mm}
\end{table}
\begin{table}[tb]
\begin{center} \small
\caption{Comparison with DG-sota methods on VLCS. Refer to Table \ref{table:resultsDG_PACS} for notation details. 
} \vspace{-5mm}
\resizebox{\columnwidth}{!}{\begin{tabular}{@{}c@{~~~}c@{~~~}c@{~~~}c@{~~~}c@{~~~}c|c}
\hline
\multicolumn{2}{c}{\textbf{VLCS}}  & \textbf{Caltech} & \textbf{Labelme} &  \textbf{Pascal} & \textbf{Sun} &   \textbf{Avg.}\\ \hline
\multicolumn{7}{c}{\textbf{Alexnet}}\\
\hline
\multirow{2}{*}{\cite{hospedalesPACS}} & DeepAll & 93.40 & 62.11 & 68.41 & 64.16 & 72.02\\
 & TF & 93.63 & 63.49 & 69.99 & 61.32 & 72.11\\
\hline
\multirow{2}{*}{\cite{Ding2017DeepDG}} & DeepAll &  86.67 & 58.20 & 59.10 & 57.86 & 65.46 \\
&  SLRC &  92.76	& 62.34	& 65.25	& 63.54	& 70.97 \\
\hline
\multirow{2}{*}{\cite{Antonio_GCPR18}} & DeepAll  & 94.95  & 57.45  & 66.06  & {65.87}  &  71.08 \\
 & D-SAM & 91.75  & 56.95  & 58.59  & 60.84  & 67.03\\
\hline
\multirow{2}{*}{\cite{episodic_hospedales}} & DeepAll  &  93.10  &  60.60  &  65.40  & 65.80   & 71.20   \\
 &  Epi-FCR &  94.10  & 64.30  & 67.10  &  65.90 & 72.90\\
\hline
\multirow{2}{*}{\cite{dg_mmld}} & DeepAll  & 95.89 & 57.88 & 72.01 & 67.76 & 73.39 \\
& {{MMLD}}  & 96.66 & 58.77 & 71.96 & 68.13 & \textbf{73.88}\\ 
\multicolumn{2}{c}{DeepAll} & 96.15 & 59.05 &  70.84 & 63.92 & 72.49$\pm$0.21\\
\multicolumn{2}{c}{{{Jigsaw}$_{\alpha=0.5, \beta=0.8}$}} & 96.46 & 59.51 & 72.95 & 64.40 & 73.33$\pm$0.16\\
\multicolumn{2}{c}{{{Rotation}$_{\alpha=0.9, \beta=0.6}$}} & 97.30  & 60.30  & 71.93  & 65.97 & \textbf{73.88}$\pm$0.62 \\
\multicolumn{2}{c}{{{Jigsaw+Rotation}$_{\alpha_{J}=0.9,\alpha_{R}=0.5, \beta=0.7}$}}    &  96.30  & 59.20   & 70.73  & 66.37  & 73.15$\pm$0.36 \\
\hline
\multirow{2}{*}{\cite{dou2019domain}}
& DeepAll$^\diamond$ & 92.86 & 63.10  & 68.67 & 64.11 & 72.19\\
& MASF$^\diamond$ & 94.78& 64.90  & 69.14  & 67.64 & {74.11} \\
& {Jigsaw}$^\diamond$ & 98.27 & 61.44 & 73.61 & 66.53 & 74.96$\pm$0.21\\
& {Rotation}$^\diamond$ & 98.40  & 62.80  & 73.03  & 67.40 & \underline{75.41}$\pm$0.63 \\
& {Jigsaw+Rotation}$^\diamond$    & 98.10   &  60.20  & 72.60  & 68.87  & 74.94$\pm$0.20 \\
\hline
\end{tabular}}
\label{table:resultsDG_VLCS}\vspace{-5mm}
\end{center}
\end{table}
\begin{table}[tbp]
\begin{center}\small
\caption{Comparison with DG-sota methods on Office-Home.  Refer to Table \ref{table:resultsDG_PACS} for notation details. 
} \vspace{-5mm}
\resizebox{\columnwidth}{!}{\begin{tabular}{@{}c@{~~}c@{~~}c@{~~}c@{~~~}c@{~~~}c@{~}|@{~~}c@{~}}
\hline
 \multicolumn{2}{c}{\textbf{Office-Home}}  & \textbf{Art} & \textbf{Clipart} &  \textbf{Product} & \textbf{Real-World} &  \textbf{Avg.}\\ \hline
\multicolumn{7}{c}{\textbf{Resnet-18}}\\
\hline
 \multirow{2}{*}{\cite{Antonio_GCPR18}} & DeepAll  & 55.59 & 42.42 & 70.34 & 70.86 & 59.81 \\
 & D-SAM  & 58.03 & 44.37 & 69.22 & 71.45 & 60.77\\
 \hline
  \multirow{2}{*}{\cite{zhou2020deep}} & DeepAll  & 58.90 & 49.40 & 74.30 & 76.20 & {64.70} \\
 & {{DDAIG}}  & 59.20 & 52.30 & 74.60 & 76.00 & \textbf{65.50}\\
\hline
\multicolumn{2}{c}{DeepAll} & 52.15 & 45.86 &  70.86 & 73.15 & 60.51$\pm$0.12\\
\multicolumn{2}{c}{{{Jigsaw}$_{\alpha=0.9,\beta=0.8}$}} & 53.04 & 47.51 & 71.47 & 72.79 & 61.20$\pm$0.11\\
\multicolumn{2}{c}{{{Rotation}$_{\alpha=0.8,\beta=0.4}$}} & 57.80  & 48.73  &  72.70 & 74.87  & 63.53$\pm$0.25 \\
\multicolumn{2}{c}{{{Jigsaw+Rotation}$_{\alpha_{J}=0.4,\alpha_{R}=0.5, \beta=0.9}$}}     &  58.33  &  49.67  & 72.97  & 75.27  &  64.06$\pm$0.31 \\
\hline
\end{tabular}}
\label{table:resultsDG_officehome}
\end{center}\vspace{-4mm}
\end{table}

\noindent\emph{Results:}
Table \ref{table:resultsDG_PACS} shows the results of our multi-task approach on the dataset PACS. 
We tested Jigsaw, Rotation and their combination. 
On average our approach produces results equal or better than all the competitors with the only exception of DDAIG which got the top results on Resnet-18.
We highlight that DDAIG 
needs domain annotation for each source sample. In many practical conditions this information might not be available \cite{mancini2018boosting}, and our multi-task method does not rely on it. Moreover, DDAIG benefits from a tailored per-domain model parameter selection, different from our approach for which the parameters are fixed and shared by all the domain pairs of each dataset.
{Analogous observations hold for the VLCS results (Table \ref{table:resultsDG_VLCS}). For Office-Home (Table \ref{table:resultsDG_officehome}), Rotation appears more suitable than Jigsaw as auxiliary task with a gain larger than three percentage points over the DeepAll baseline and with even higher advantage in the Jigsaw+Rotation case.} DDAIG, although producing apparently the top average result, improves slightly more than one percentage point over its DeepAll reference. 
\subsubsection{Single-Source Domain Generalization}
\label{sec:exp_DG_3}
\begin{figure*}[!t]
\centering
\resizebox{0.9\textwidth}{!}{
\begin{tabular}{c@{~}c@{~}c@{~}c@{~}}
\includegraphics[width=0.246\textwidth]{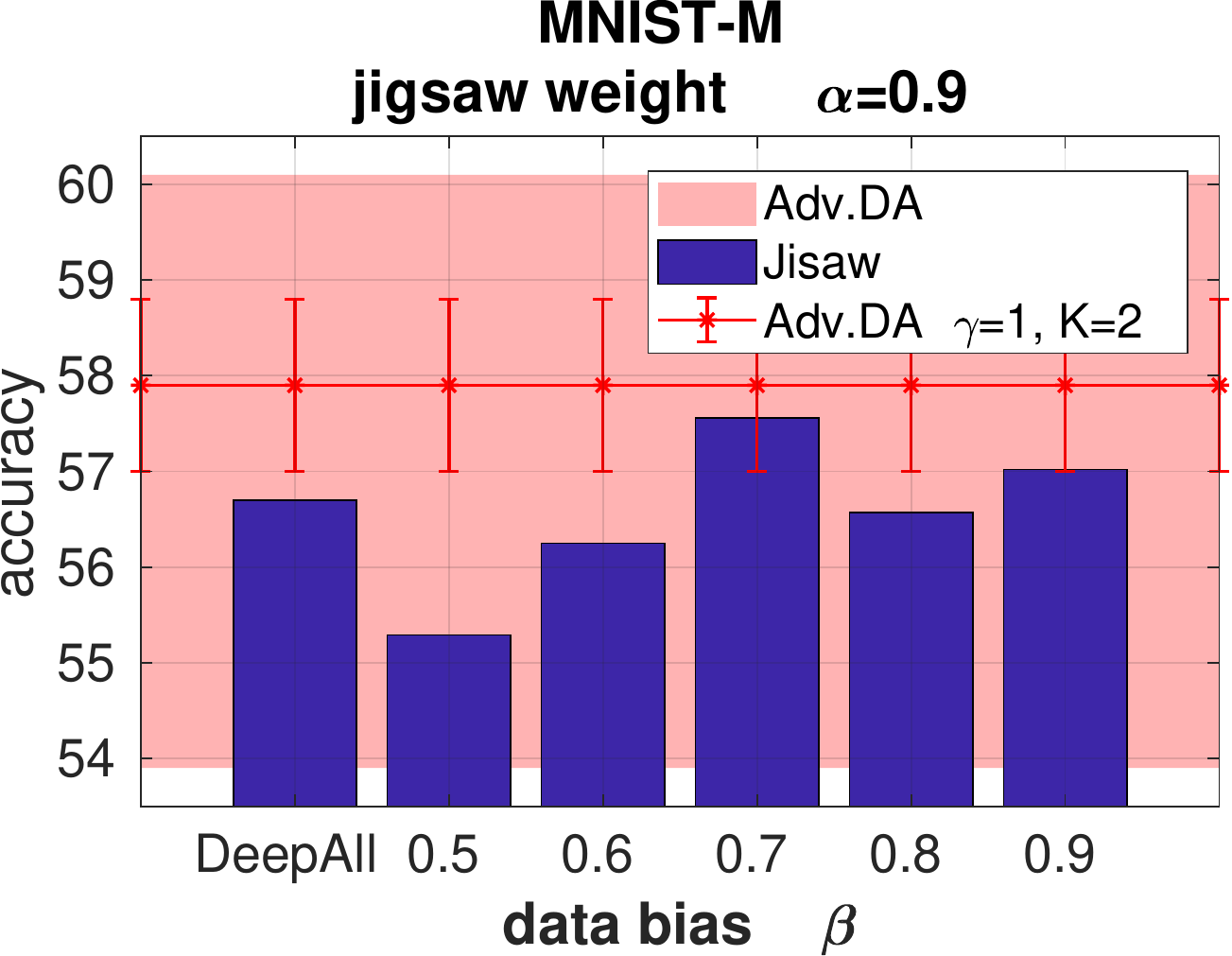} & \includegraphics[width=0.246\textwidth]{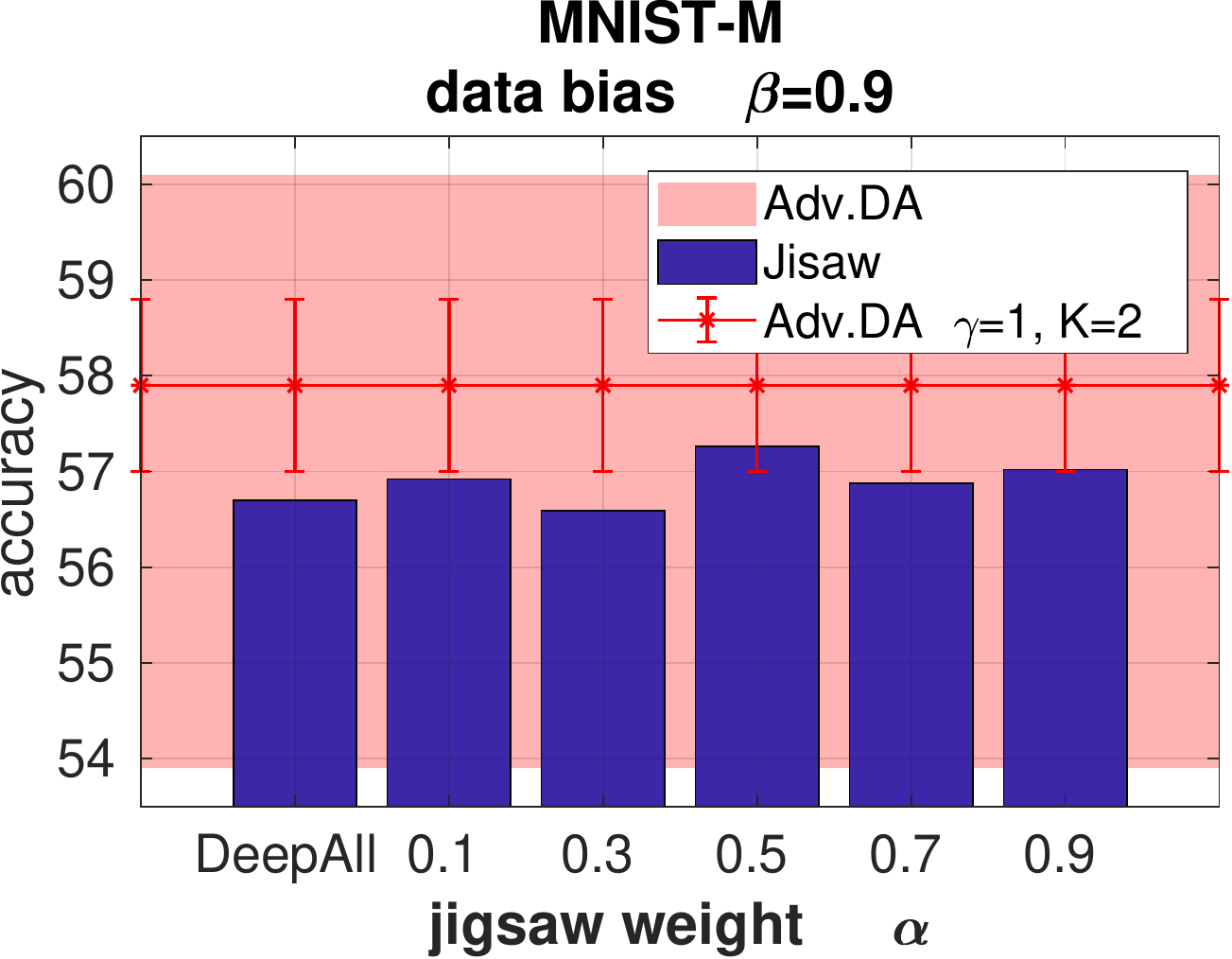} & \includegraphics[width=0.246\textwidth]{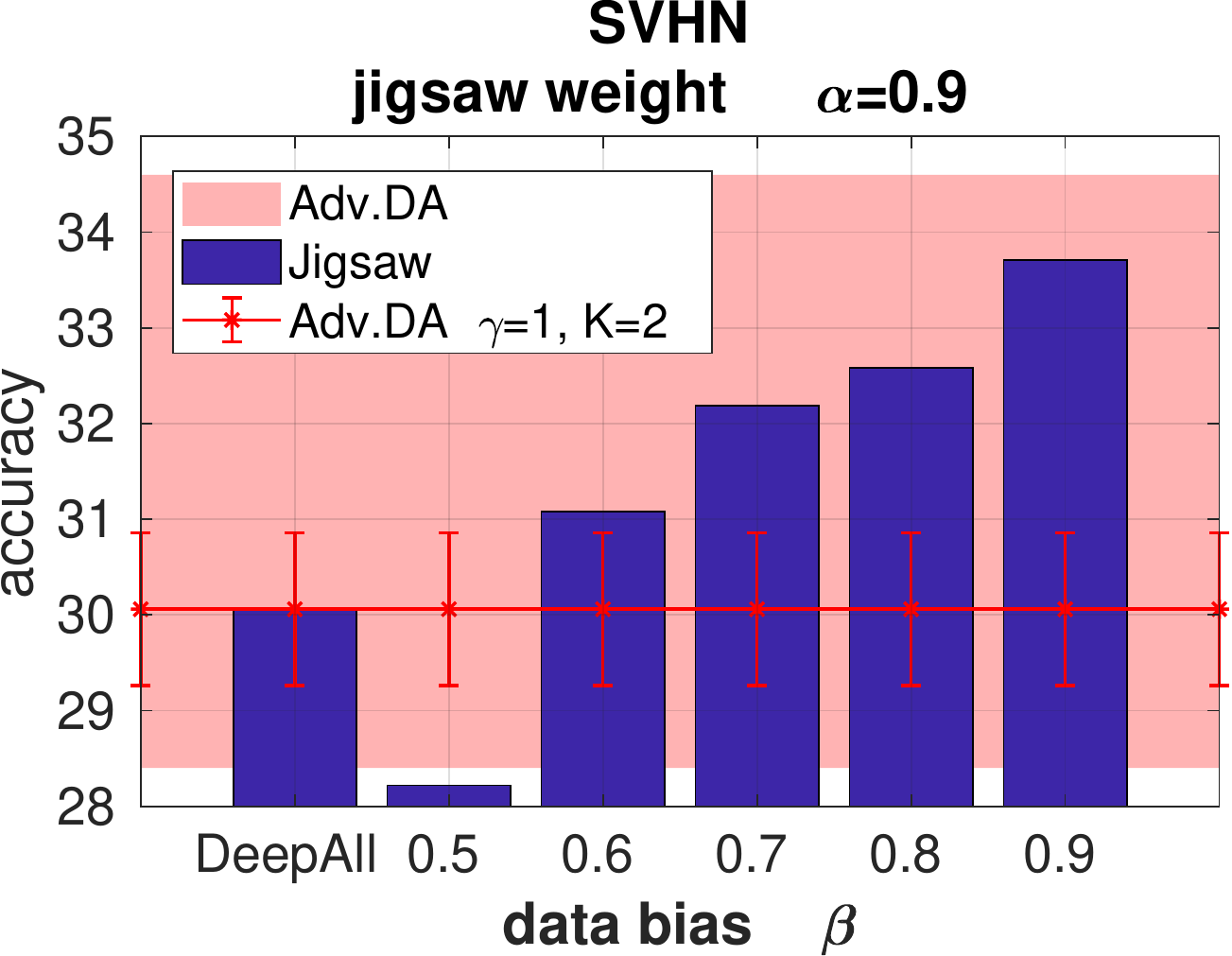}   & \includegraphics[width=0.246\textwidth]{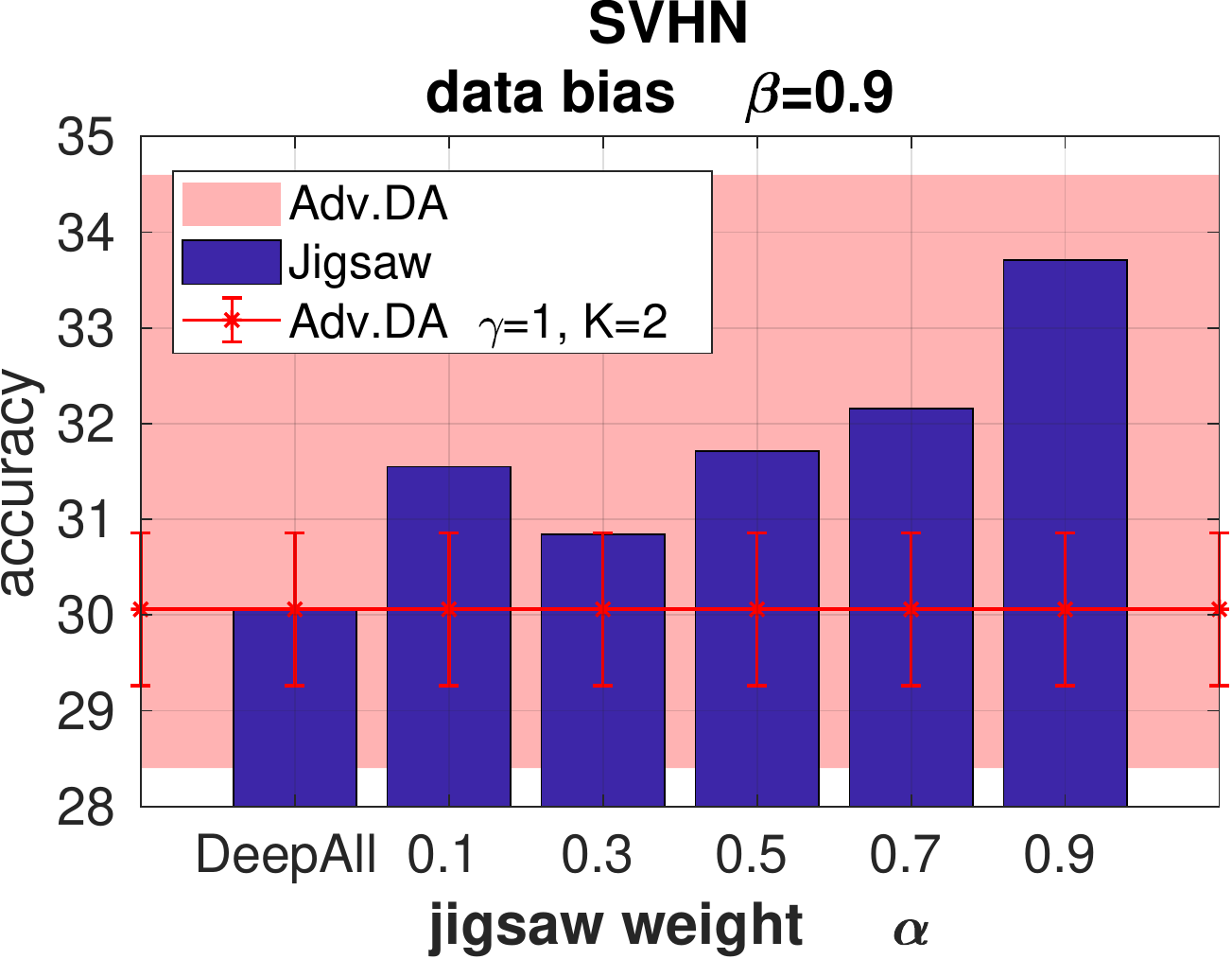}\\
\vspace{0.2cm}
\includegraphics[width=0.246\textwidth]{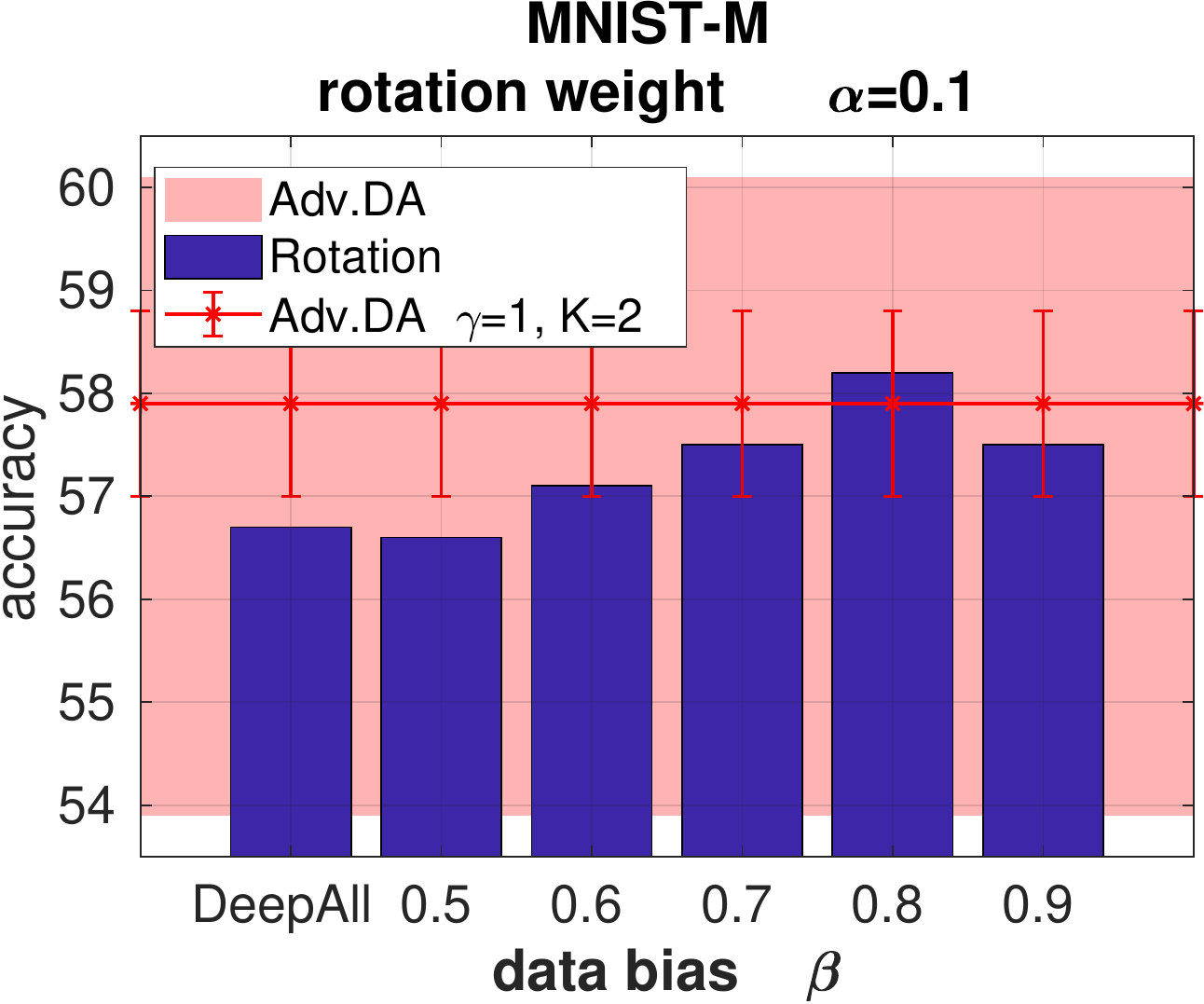} & \includegraphics[width=0.246\textwidth]{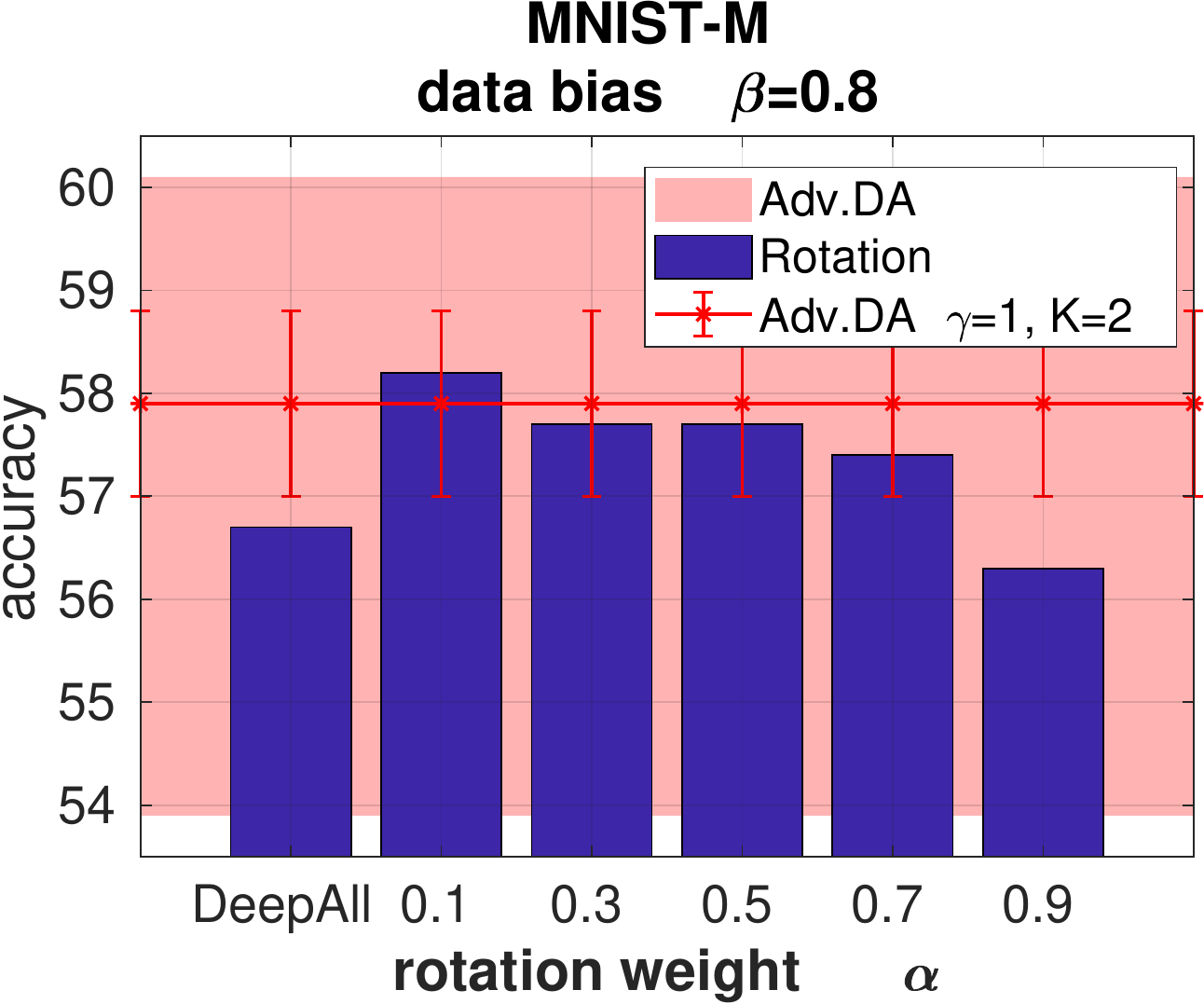} & \includegraphics[width=0.246\textwidth]{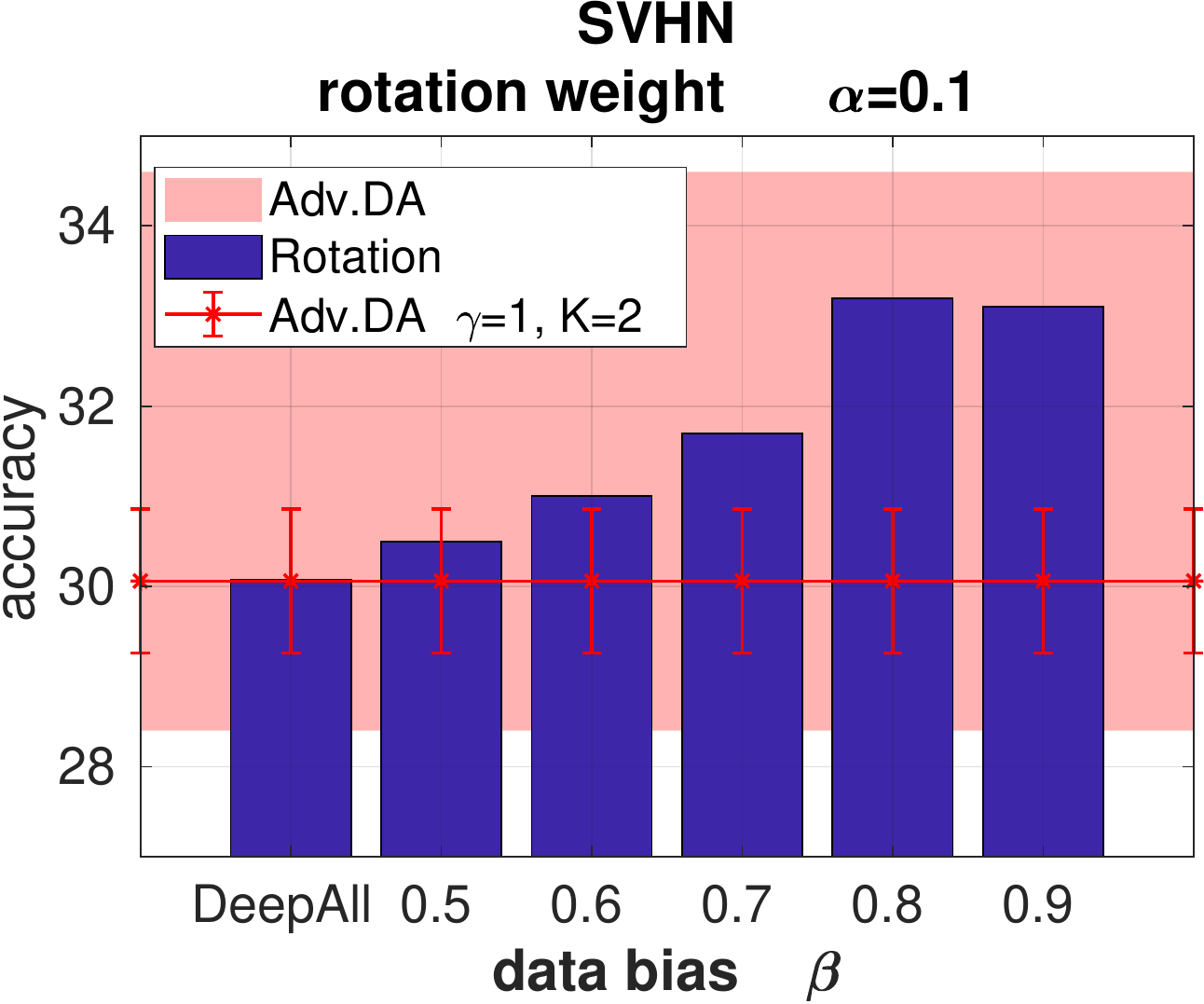}   & \includegraphics[width=0.246\textwidth]{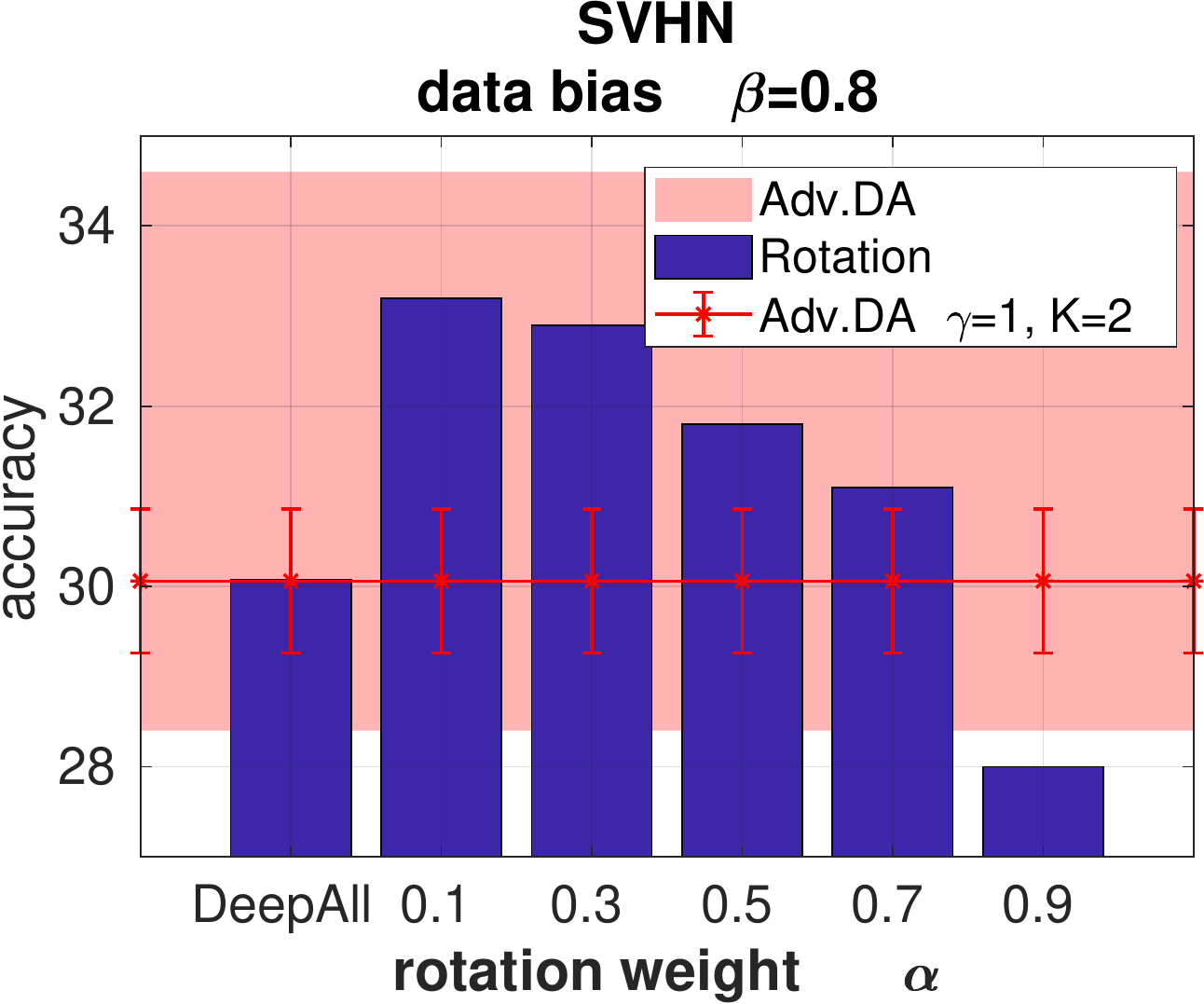}\\ 
\end{tabular}
}\vspace{-3mm}
    \caption{Single Source DG experiments. We analyze the performance
    of our multi-task Jigsaw (top row) and Rotation (bottom row) approaches in comparison with Adv.DA \cite{Volpi_2018_NIPS}. The
    shaded background area covers the overall range of results of Adv.DA obtained when changing
    the hyper-parameters of the method. The reference result of Adv.DA ($\gamma=1$, $K=2$) together 
    with its standard deviation is indicated here by the horizontal red line.
    The blue histogram bars show the performance of Jigsaw and Rotation when changing the self-supervised task weight $\alpha$ and data bias $\beta$ .}
    \label{fig:singesource}\vspace{-3mm}
\end{figure*}

\begin{figure*}[!t]
    \centering
    \resizebox{0.9\textwidth}{!}{
    \begin{tabular}{c@{~~~}c@{~~~}c@{~~~}c}
\includegraphics[height=3.5cm]{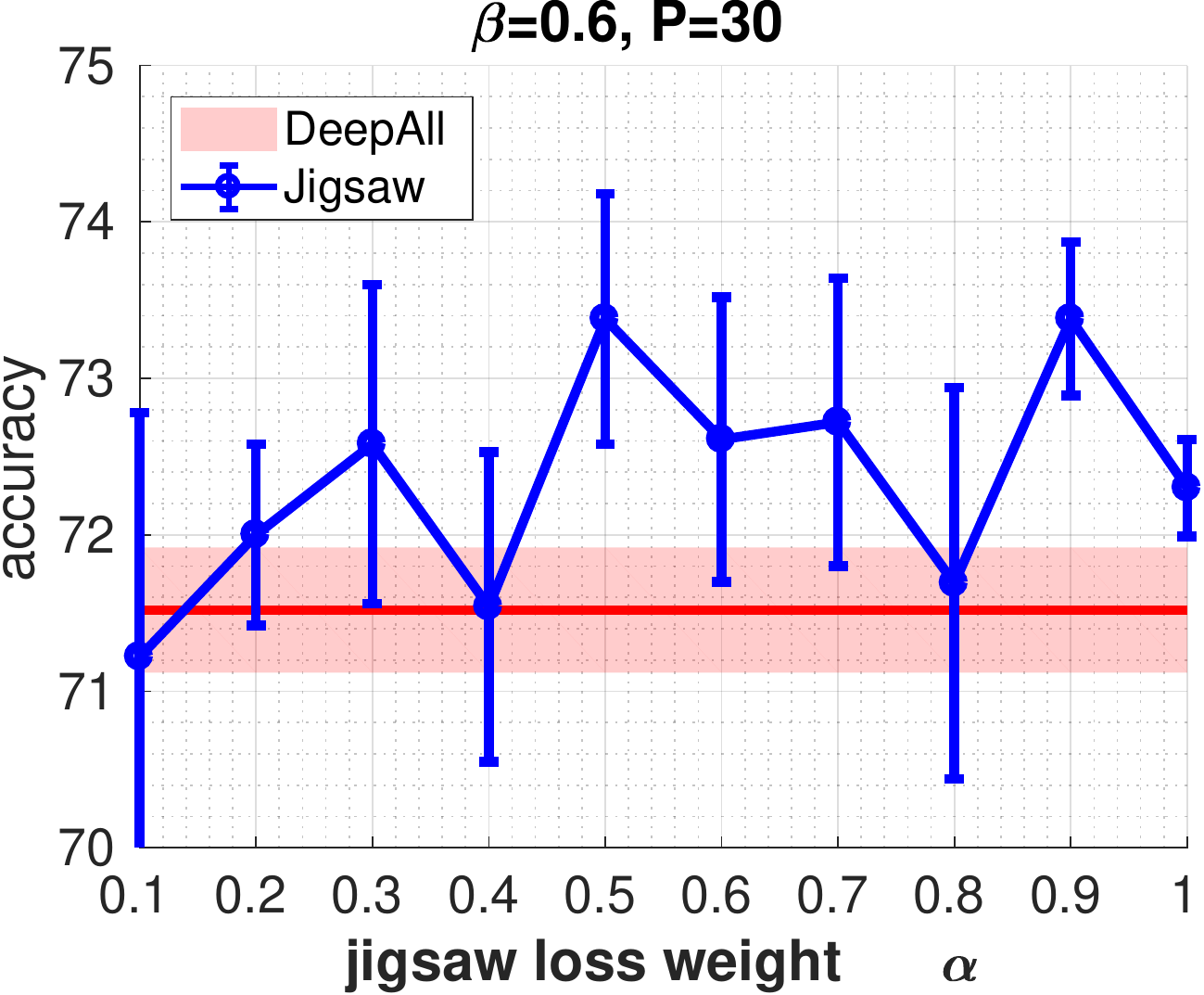}   &  \includegraphics[height=3.5cm]{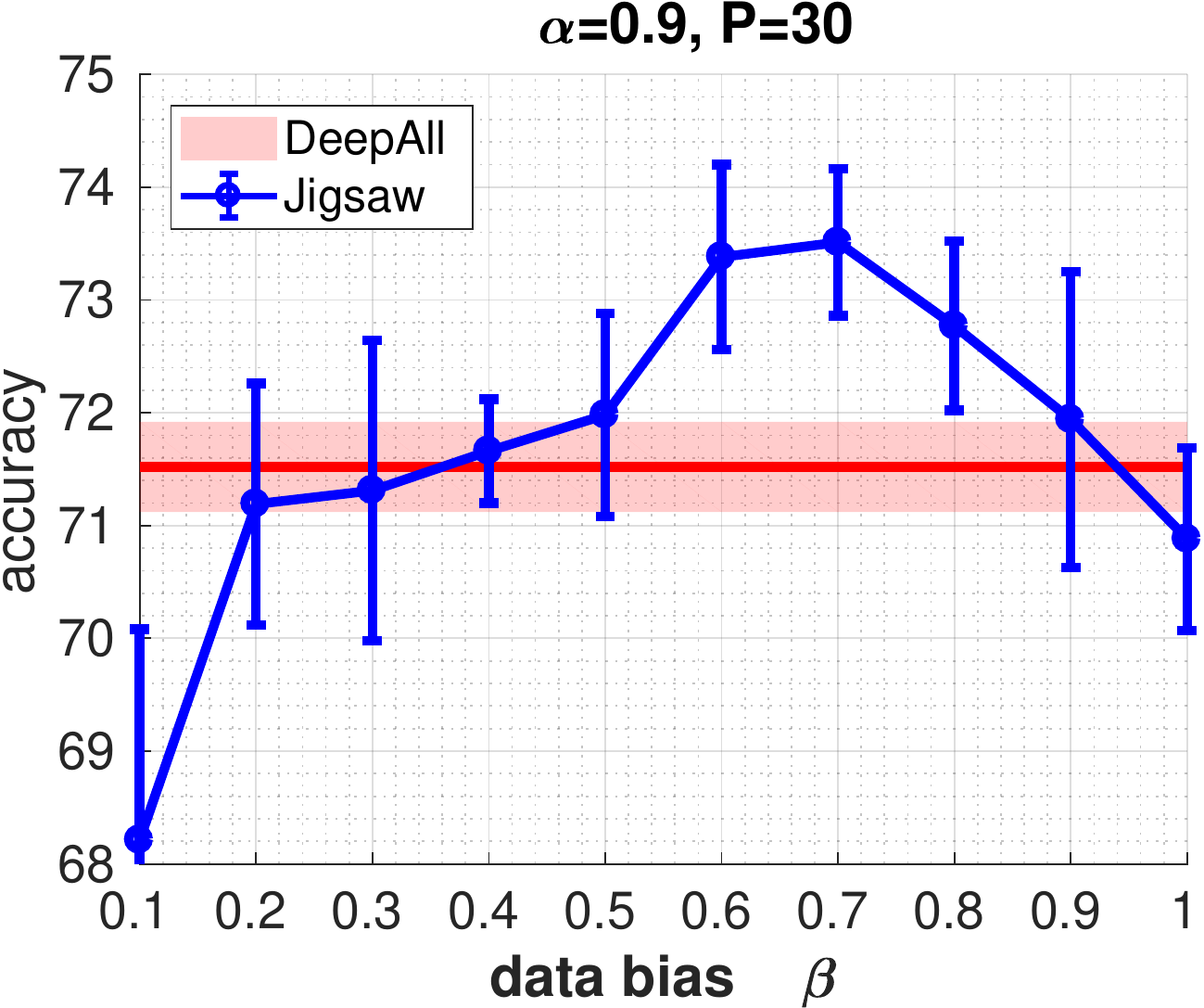} &  \includegraphics[height=3.65cm]{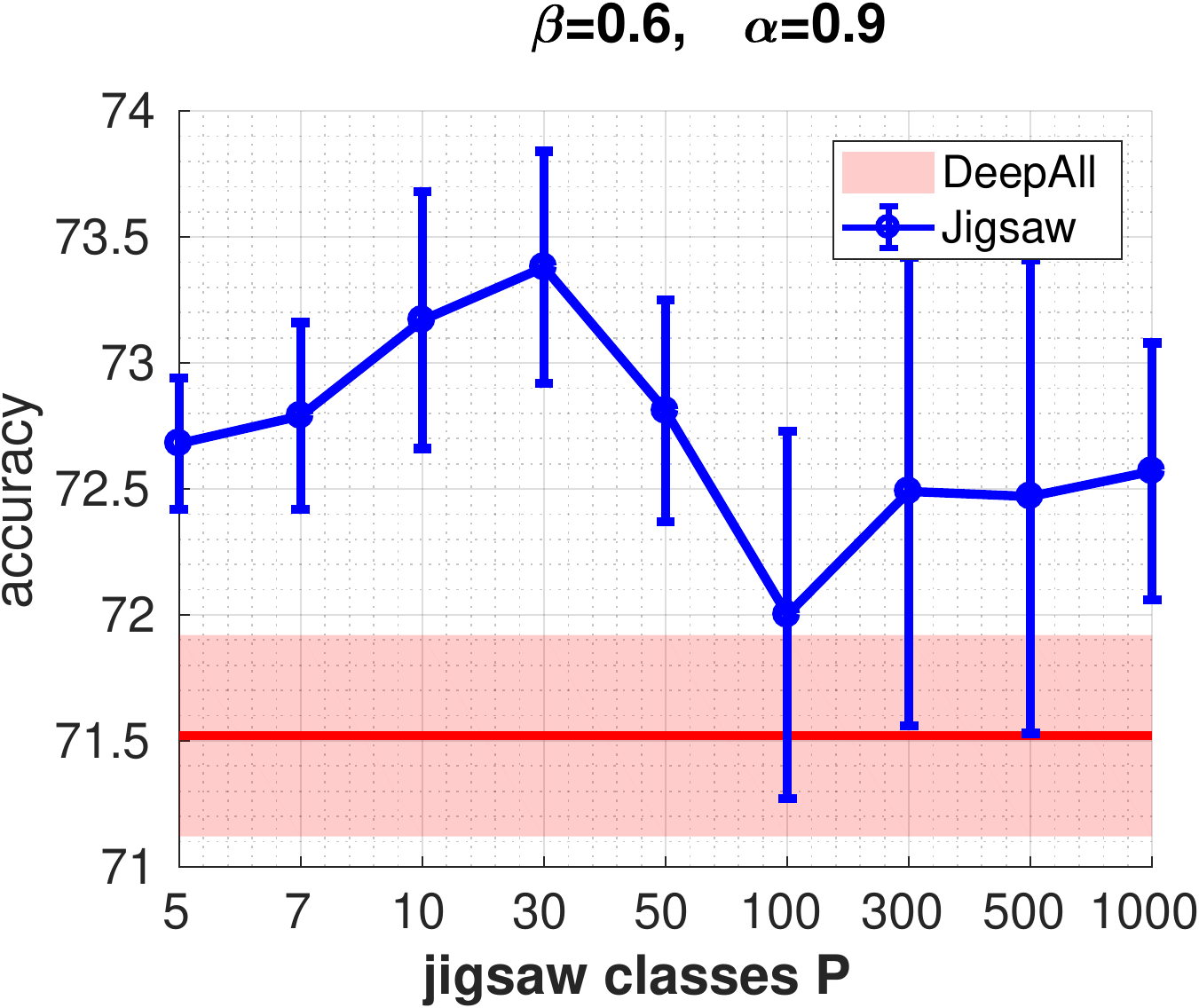}  & \includegraphics[height=3.5cm]{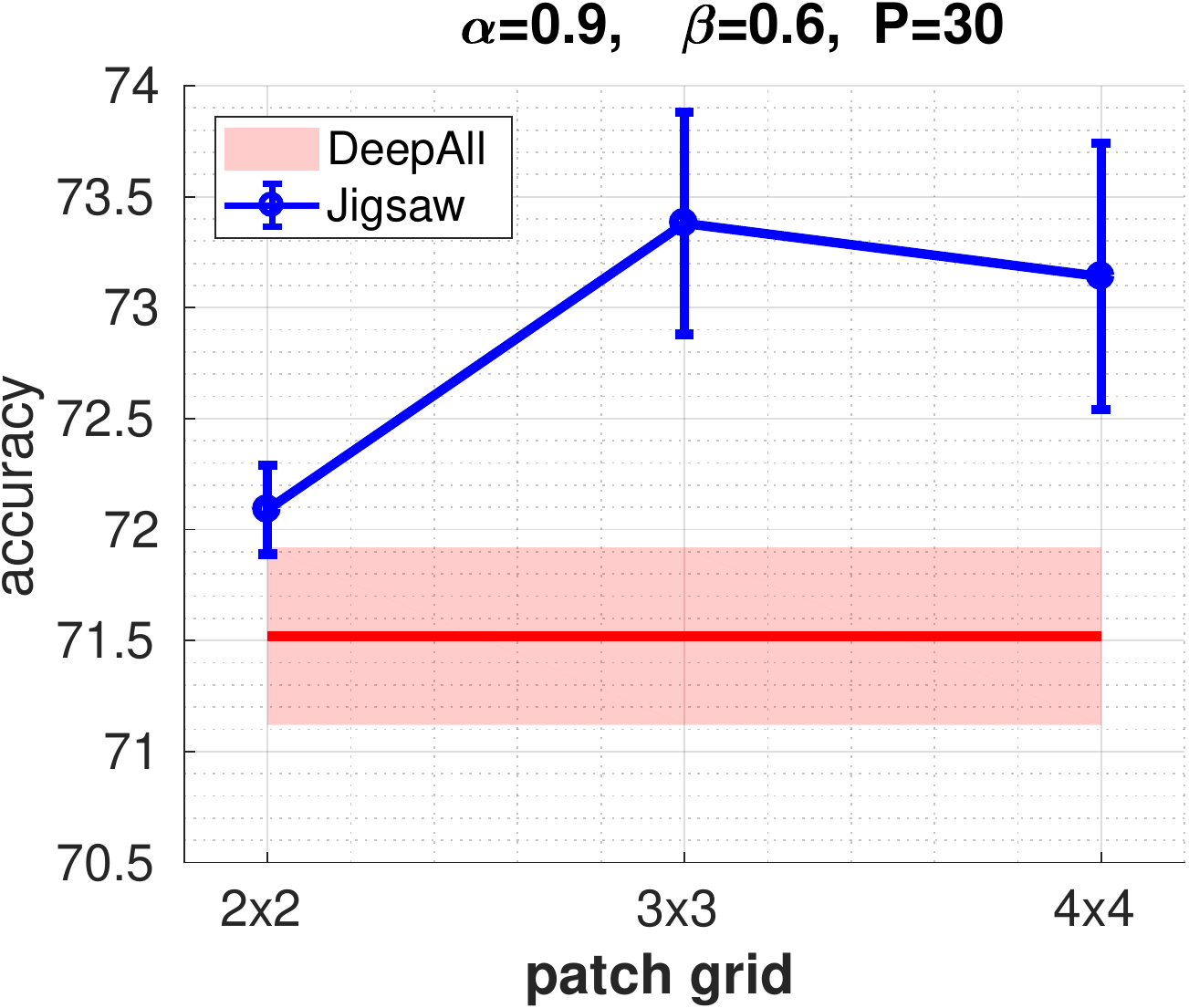}\\
\end{tabular}
}\vspace{-2mm}
\caption{Ablation results and hyper-parameter analysis on the Alexnet-PACS DG setting when using Jigsaw. The reported accuracy is the global average over all the target domains with three repetitions for each run. The red line represents our {DeepAll} average from Table \ref{table:resultsDG_PACS}.}
\label{fig:ablation}\vspace{-5mm}
\end{figure*}

\begin{figure}[!t]
\centering
\begin{tabular}{c@{~~~}c}
\hspace{-4mm}\includegraphics[height=3.3cm]{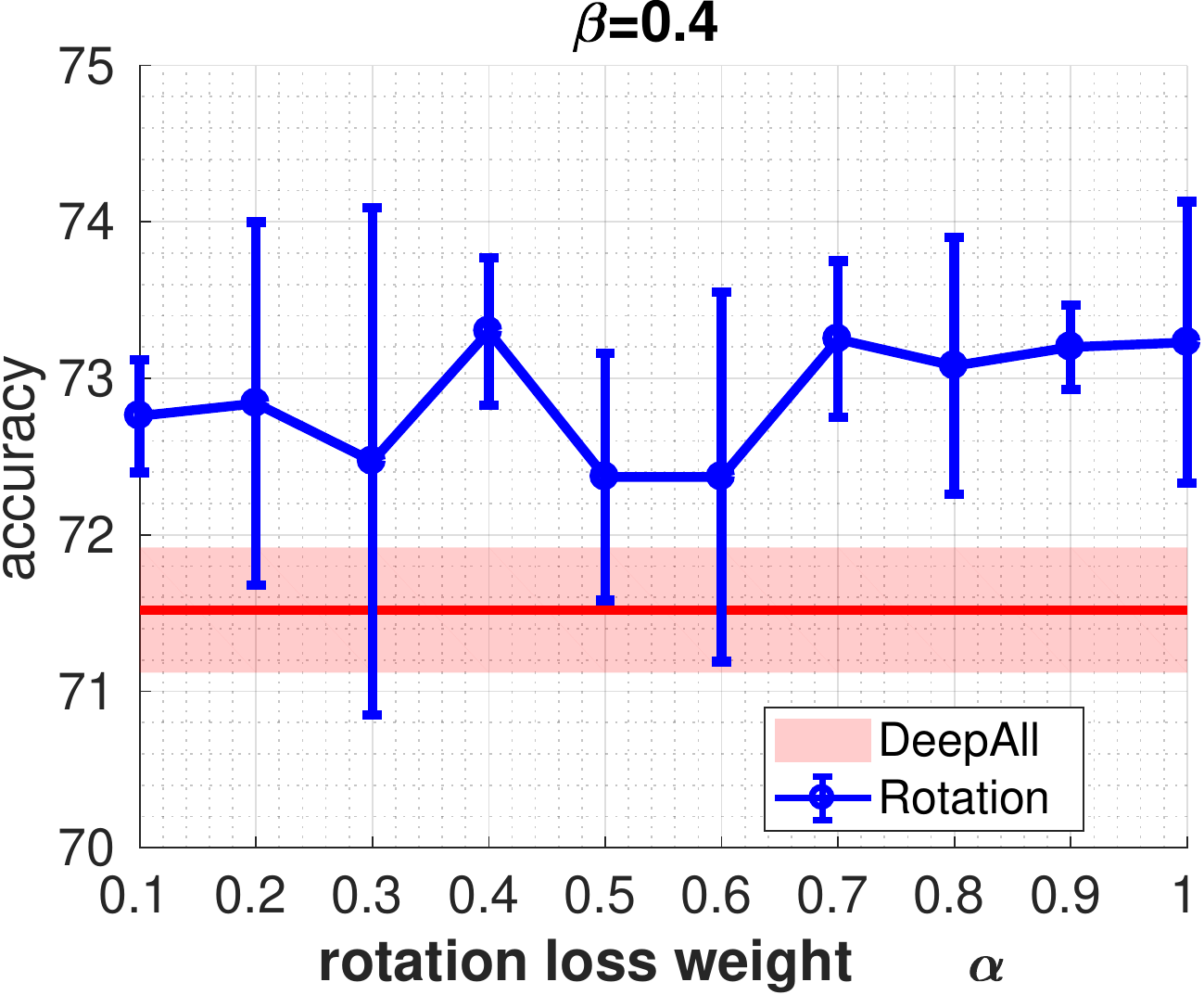} &
\includegraphics[height=3.3cm]{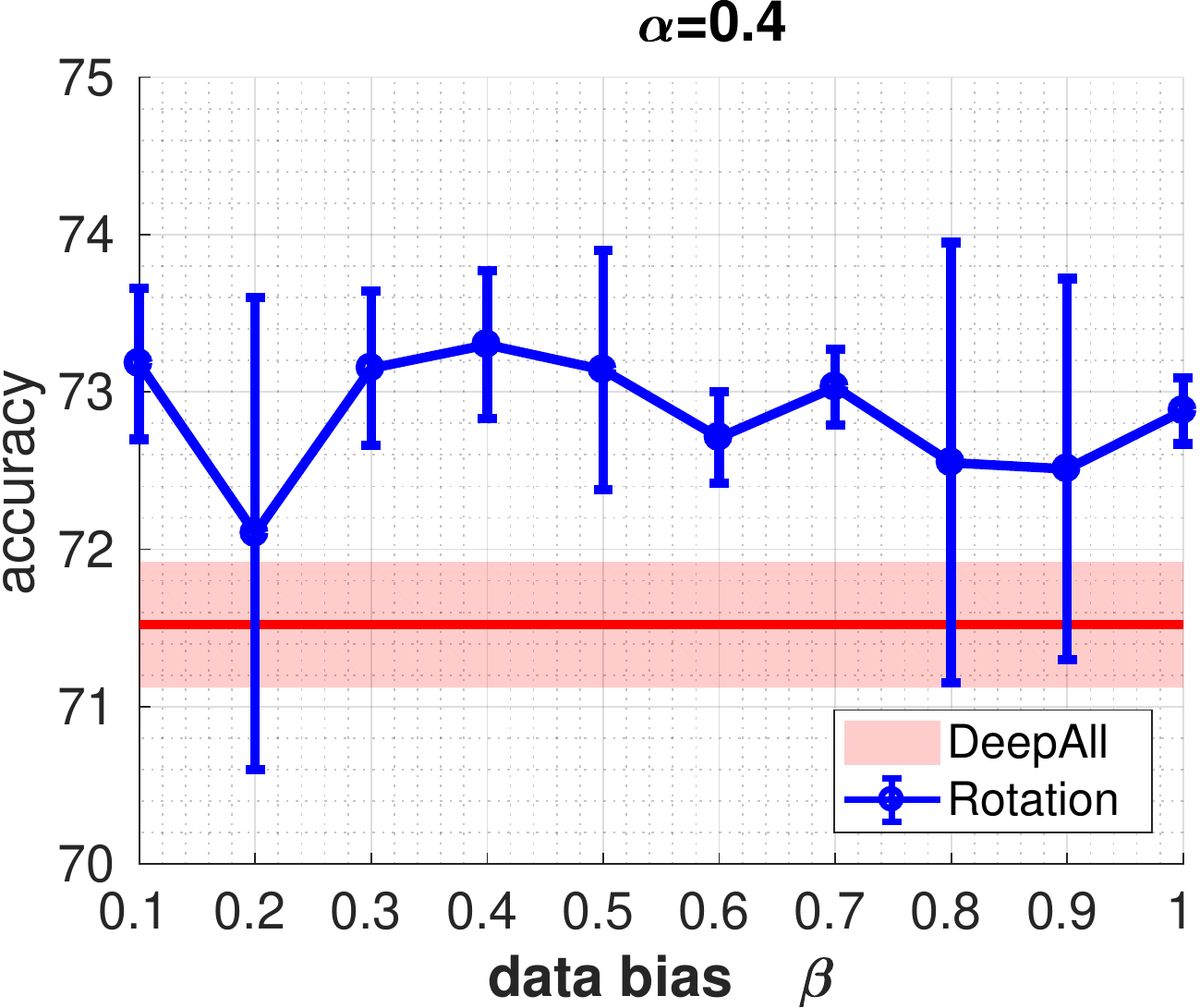} \\
\end{tabular}\vspace{-2mm}
\caption{Ablation results on the Alexnet-PACS DG setting when using Rotation. We report the average accuracy over all target domains with three repetitions for each run. The red line is our {DeepAll} from Table \ref{table:resultsDG_PACS}.}
\label{fig:rotablation}
\end{figure}
\begin{figure}[!t]\hspace{-5mm}
    \begin{tabular}{c@{}c}
\hspace{-2mm}\includegraphics[height=2.9cm]{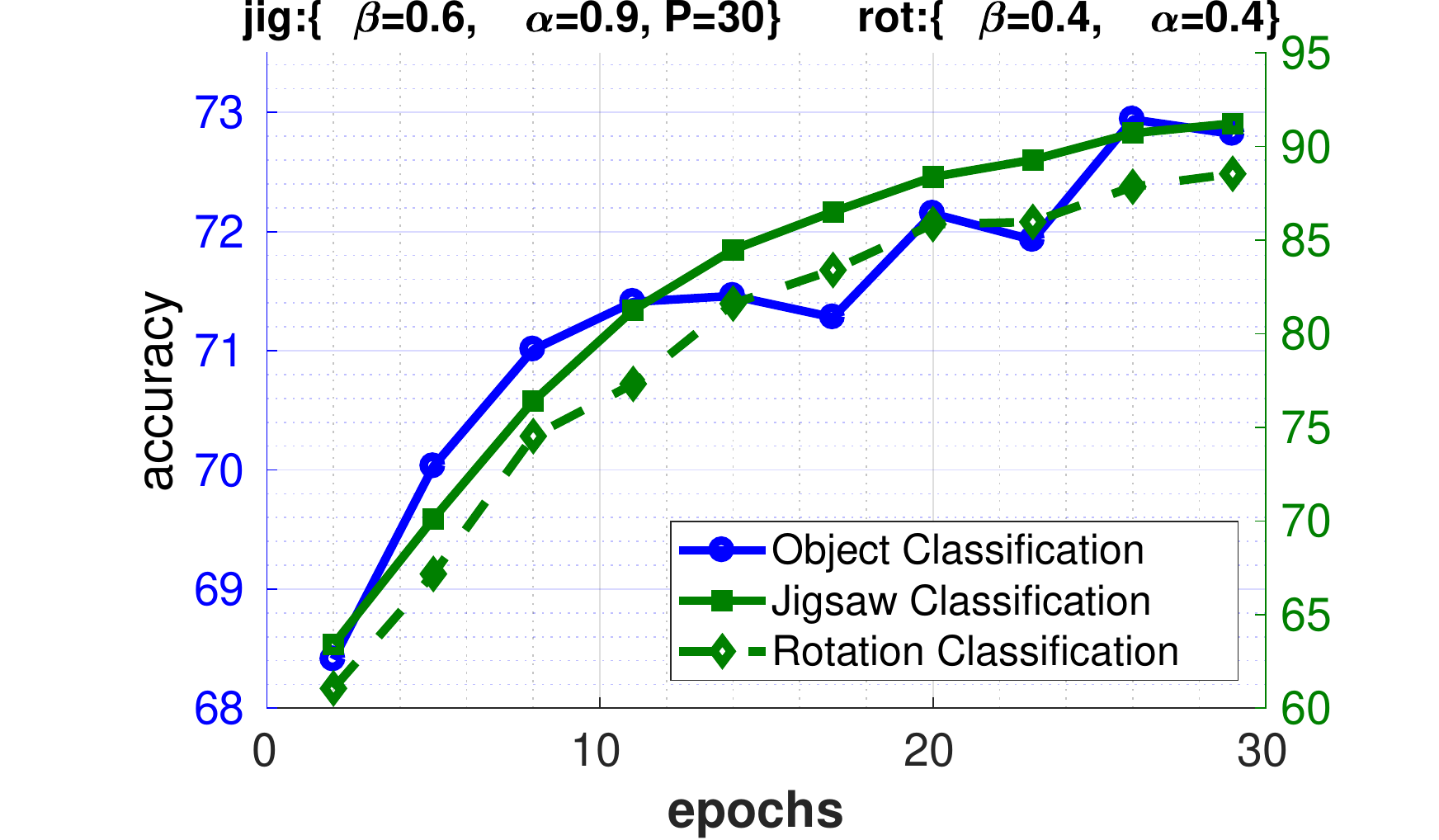}  &  \includegraphics[height=2.9cm]{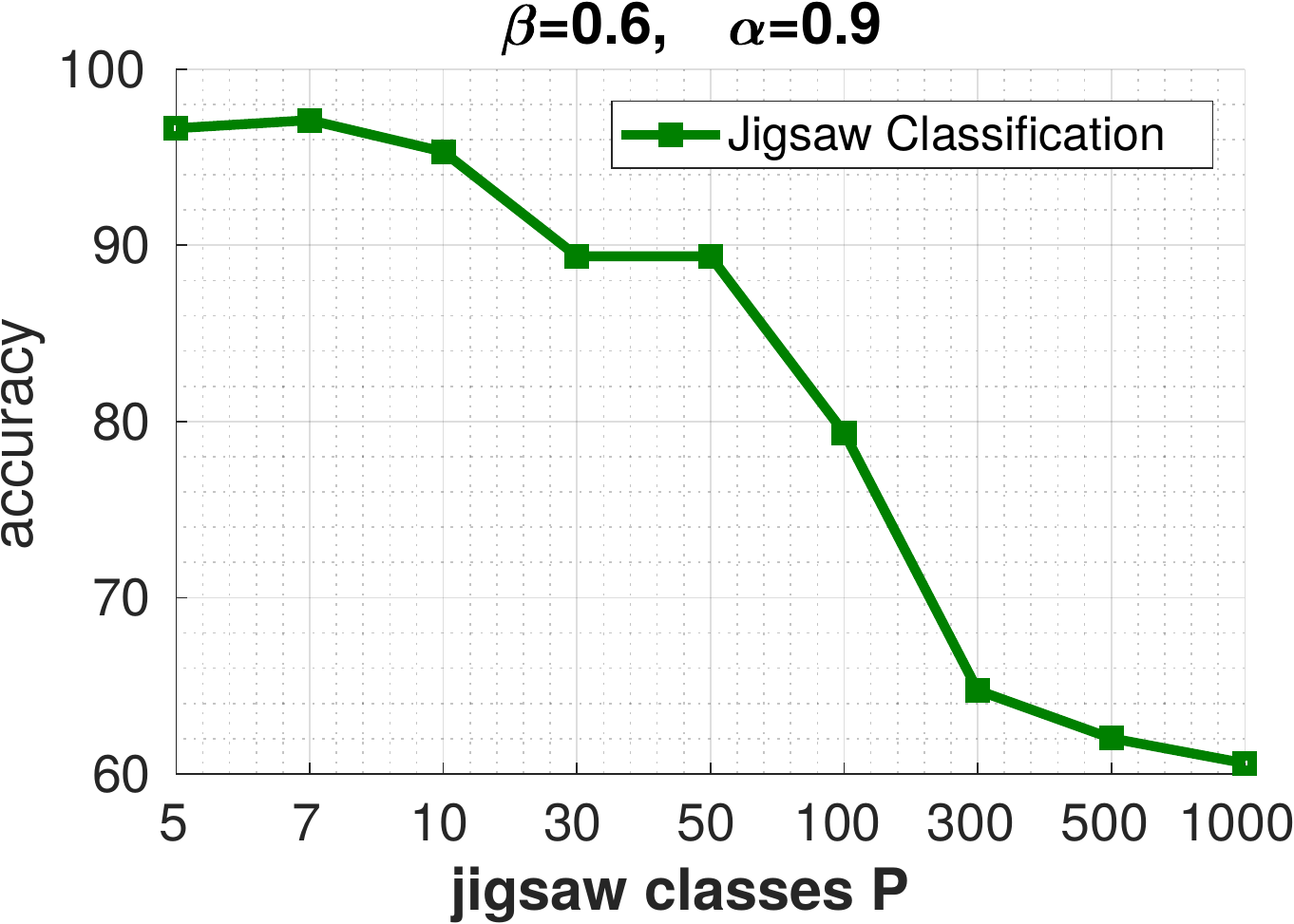}\\
    \end{tabular}\vspace{-1mm}
    \caption{Analysis of the Jigsaw classifier on Alexnet-PACS DG setting. In the left plot each axes refers to the color matching curve in the graph.}
    \label{fig:ablation_jigsaw}\vspace{-4mm}
\end{figure}

The generalization ability of a model depends both on the learning process and on the used training data. 
To better evaluate the regularization effect provided by the self-supervised tasks, we investigate the case of training data from a single source domain.

\noindent\emph{Baseline and Datasets:} For these experiments we compare against the generalization method based on adversarial data augmentation ({Adv.DA}) presented in  \cite{Volpi_2018_NIPS}. 
We based our model on their same backbone (conv-pool-conv-pool-fc-fc-softmax), we reproduced their experimental setting and adopted a similar 
result display style with bar plots. 
We trained a model on 10k digit samples of the MNIST dataset 
\cite{lecun1998gradient} and evaluated on the respective test sets of MNIST-M~\cite{Ganin:DANN:JMLR16} and SVHN~\cite{netzer2011reading}.  The digits are handwritten  on black background for MNIST and on colorful background for MNIST-M. In SVHN the images are house numbers from Google Street View. To work with comparable datasets, all the images were resized to $32\times32$ and treated as RGB.

\noindent\emph{Results:} 
In Figure \ref{fig:singesource} we show the performance of Jigsaw and Rotation when varying the data bias $\beta$ 
and the self-supervised task weight $\alpha$. With the red background shadow we indicate the overall range 
covered by Adv.DA results when changing its parameters, while the horizontal line is the reference Adv.DA results
around which the authors of \cite{Volpi_2018_NIPS} ran their ablation analysis. 
The bar plots indicates that,
although Adv.DA can reach high peak values, it is also very sensitive to the chosen hyper-parameters. 
On the other hand, our multi-task approach is much more stable and usually performs better than Adv.DA. One exception arises on SVHN, with Jigsaw  when the data bias is 0.5, and with Rotation when the self-supervised task weight is 0.9: both correspond to limit cases for the proper combination of object classification and self-supervised learning as will be discussed in the next section.
Moreover, Jigsaw and Rotation have similar performance to Adv.DA on MNIST-M and significantly outperform it on SVHN.

\subsubsection{Ablation and hyper-parameter tuning}
\label{sec:exp_DG_4}
\begin{figure*}[!t]
\vspace{-2mm}
\centering
\resizebox{\textwidth}{!}{
\begin{tabular}{c@{~}c@{~}c@{~}c@{~}c@{~}c@{~}c@{~}c@{~~}|@{~~} c@{~}c@{~}c@{~}c@{~}c@{~}c@{~}c@{~}c} 
& {DeepAll  \ding{55}} & {Jigsaw \ding{51}} &
{Rotation \ding{51}} & &
{DeepAll \ding{55}} & {Jigsaw \ding{51}} & {Rotation \ding{51}} 
& & {DeepAll \ding{51}} & {Jigsaw \ding{51}}& 
{Rotation \ding{55}}& & {DeepAll  \ding{51}} & {Jigsaw \ding{51}} &
{Rotation \ding{55}}\\
\includegraphics[width=0.145\linewidth,frame]{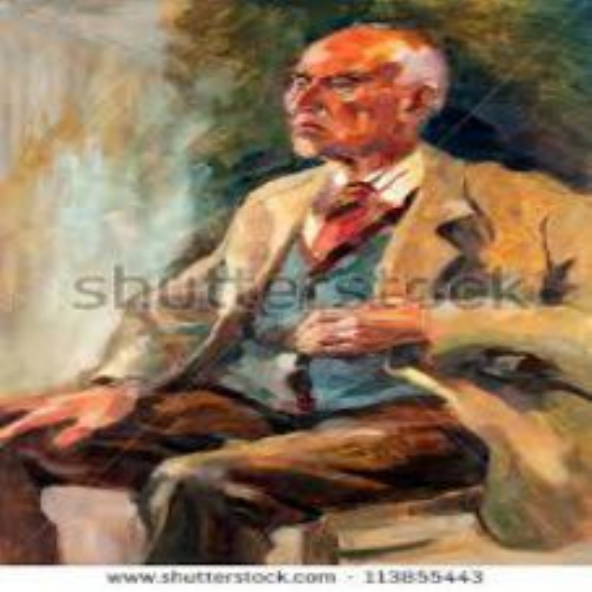} &
\includegraphics[width=0.145\linewidth,frame]{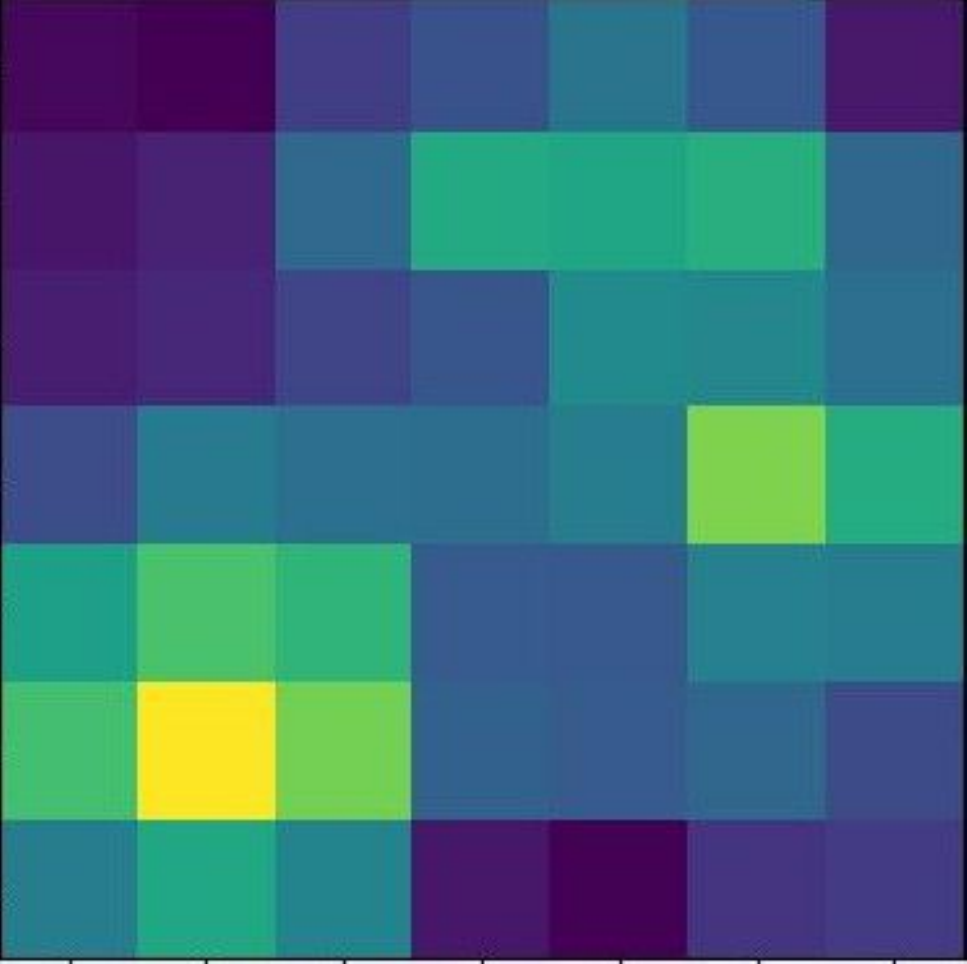}&
\includegraphics[width=0.145\linewidth,frame]{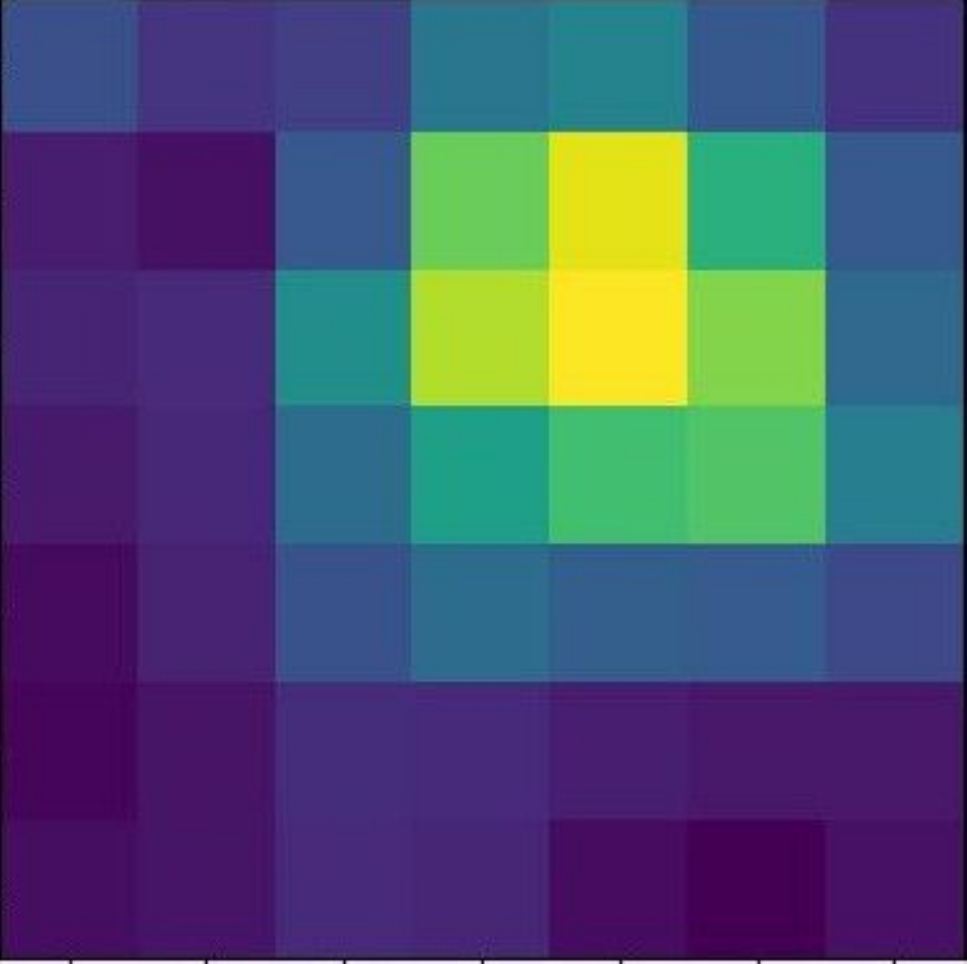} &
\includegraphics[width=0.145\linewidth,frame]{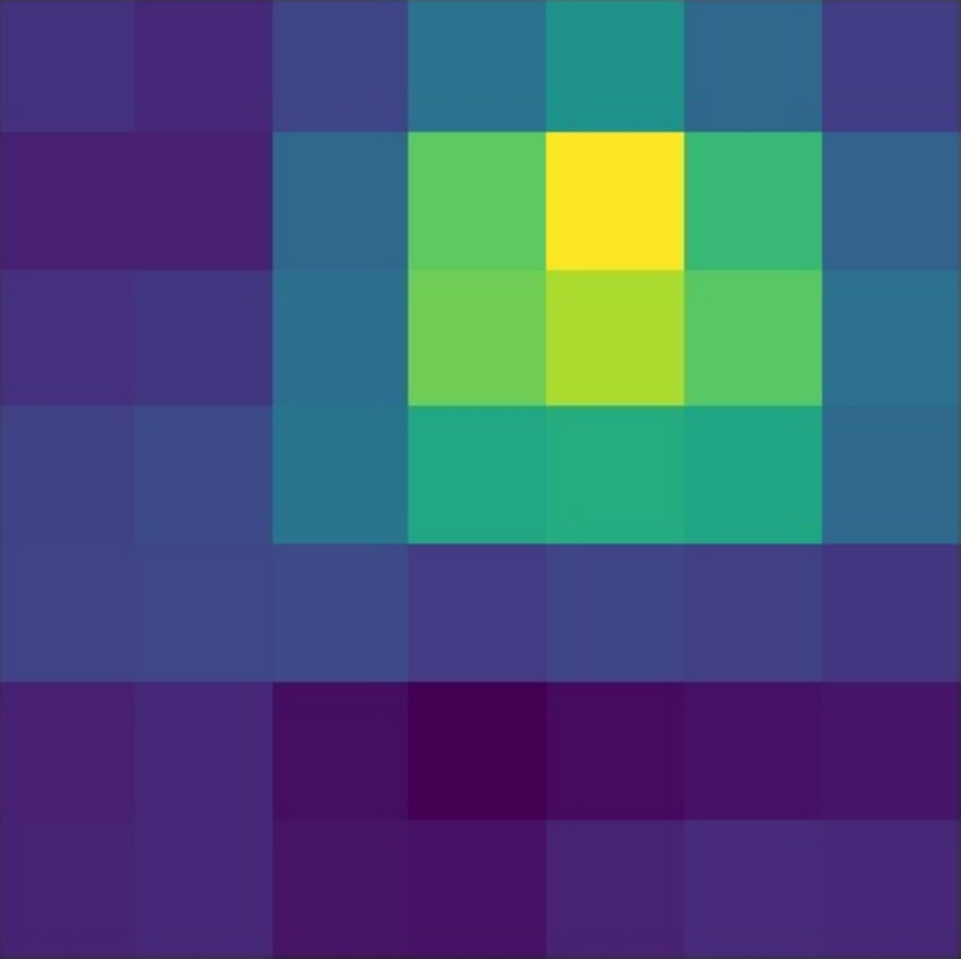}
 & \includegraphics[width=0.145\linewidth,frame]{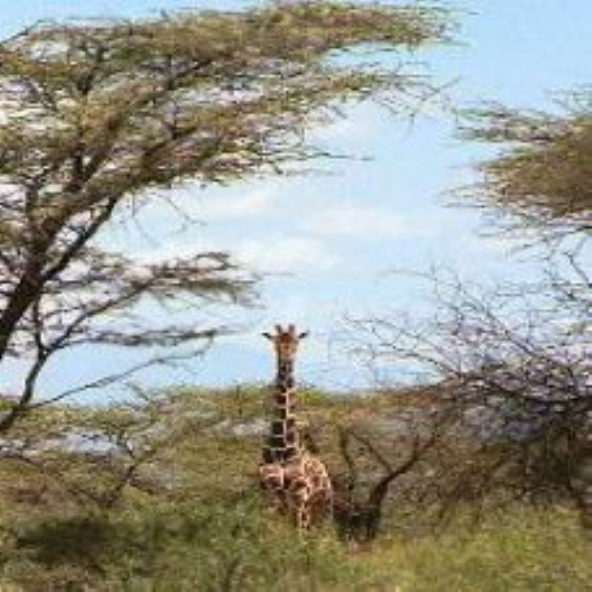} &
\includegraphics[width=0.145\linewidth,frame]{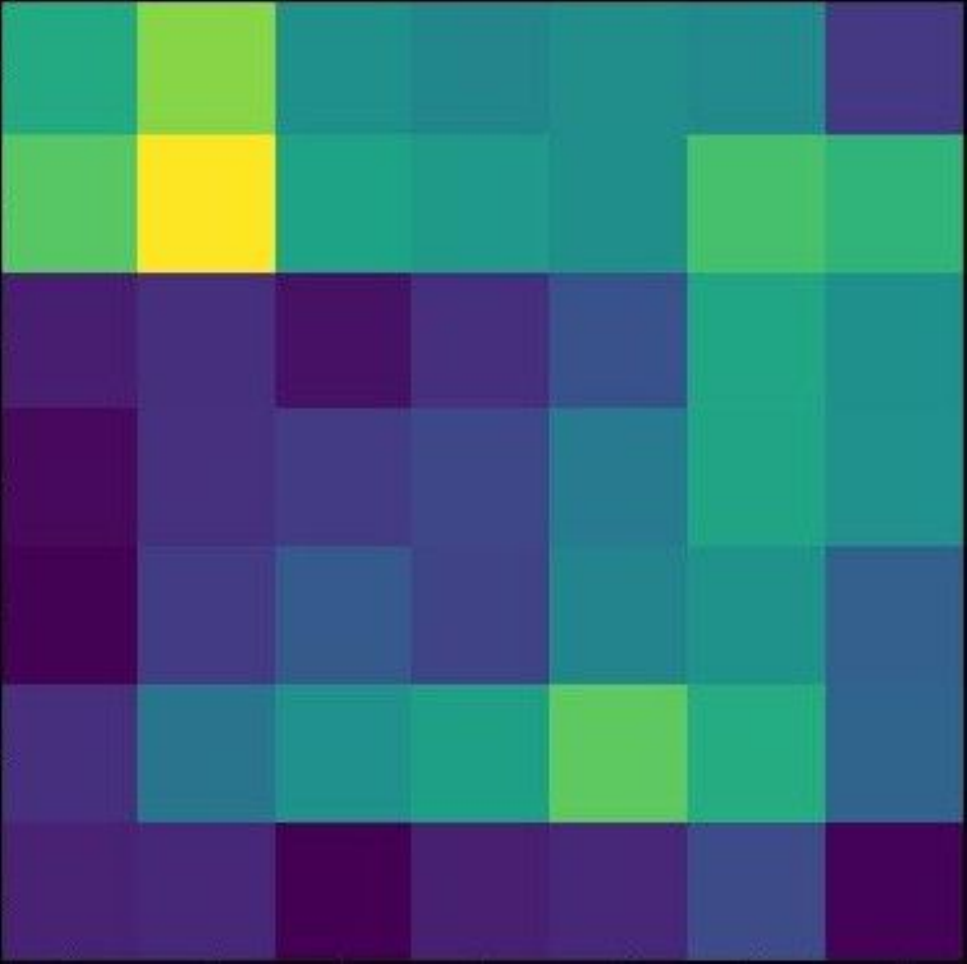}&
\includegraphics[width=0.145\linewidth,frame]{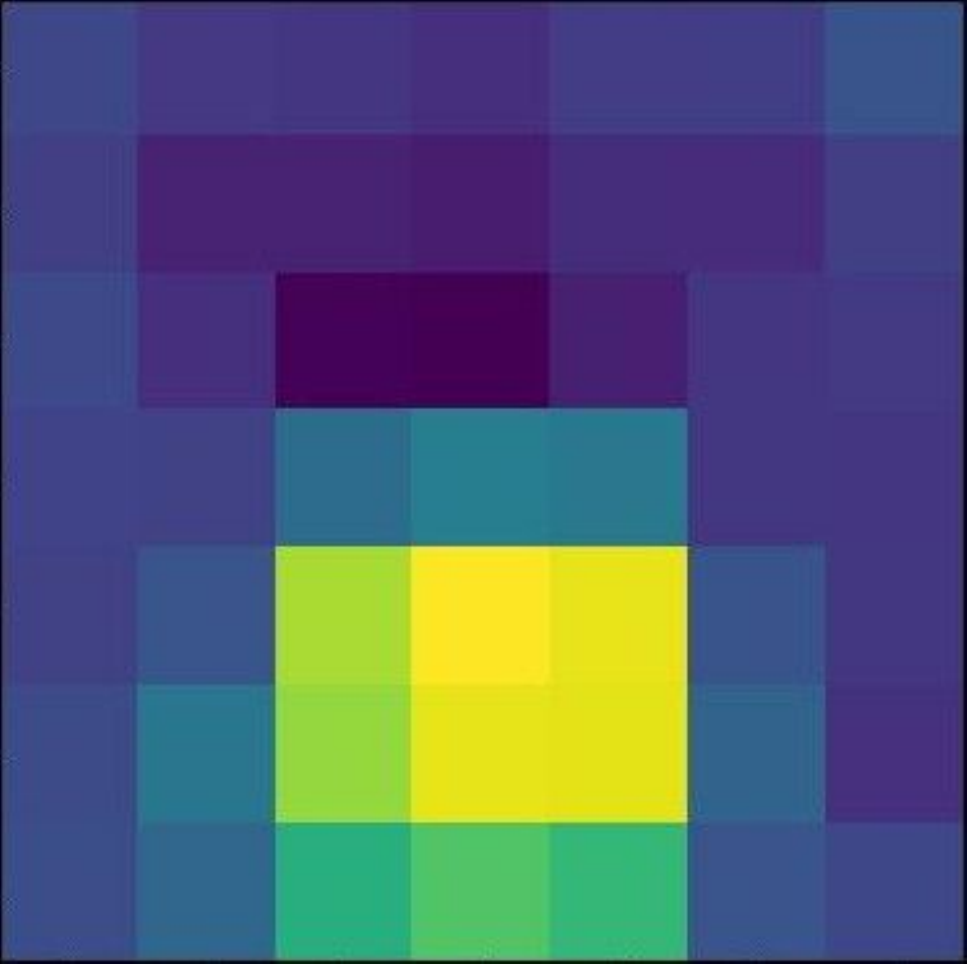} &
\includegraphics[width=0.145\linewidth,frame]{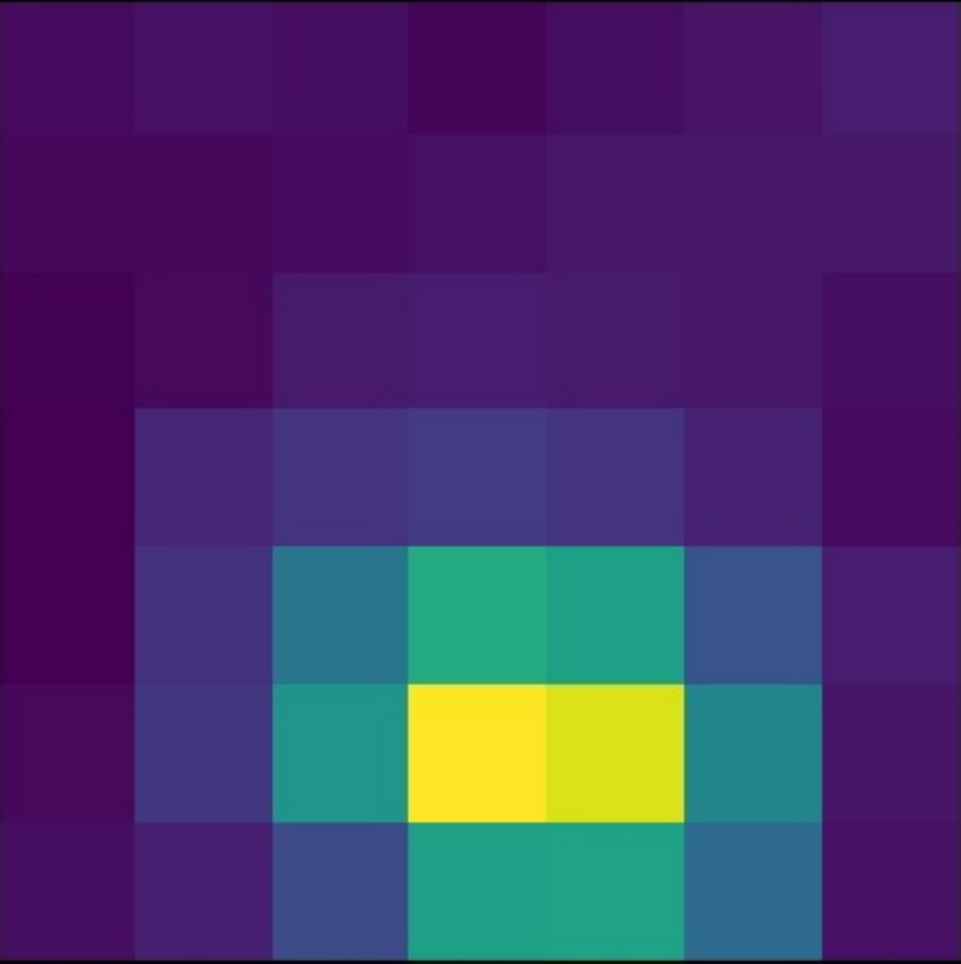}
& \includegraphics[width=0.145\linewidth,frame]{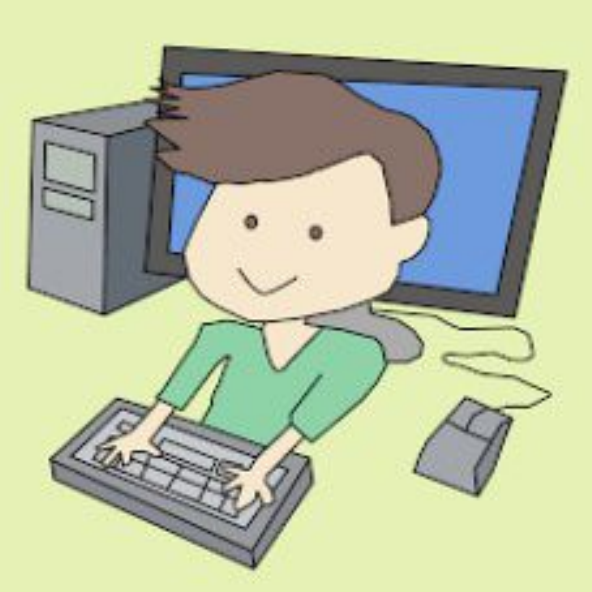} &
\includegraphics[width=0.145\linewidth,frame]{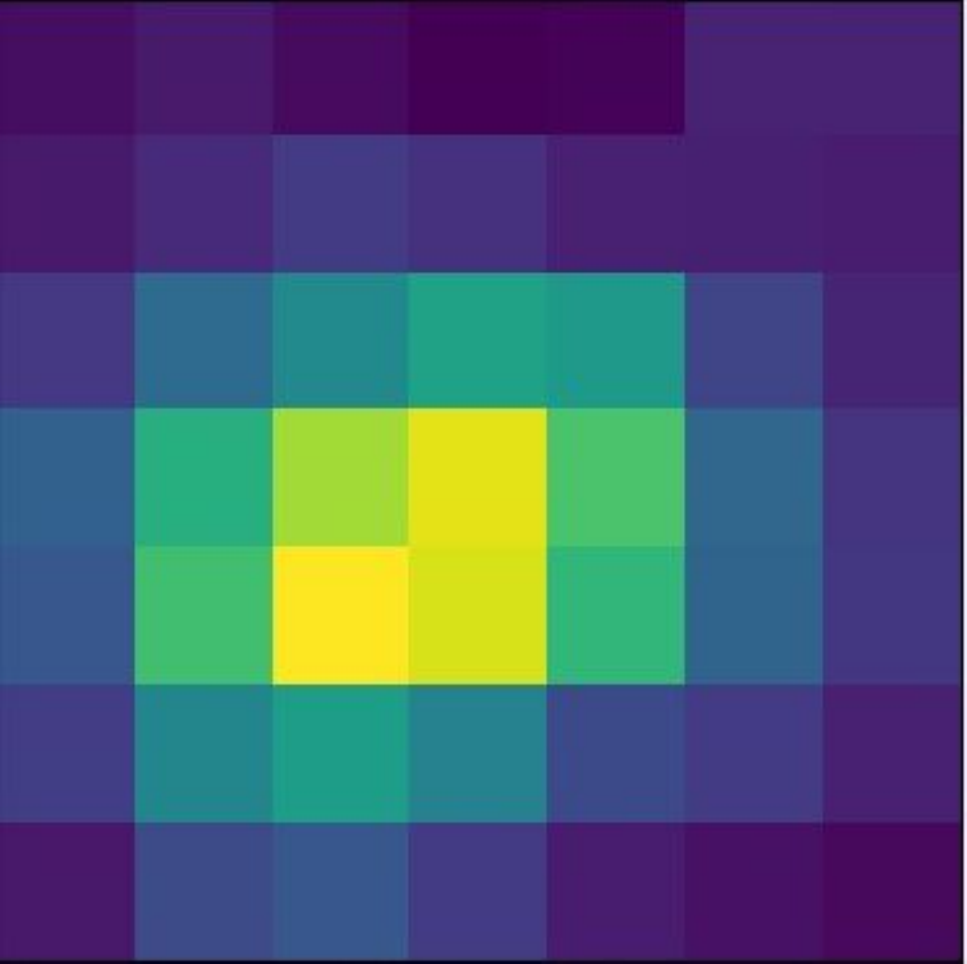} &
\includegraphics[width=0.145\linewidth,frame]{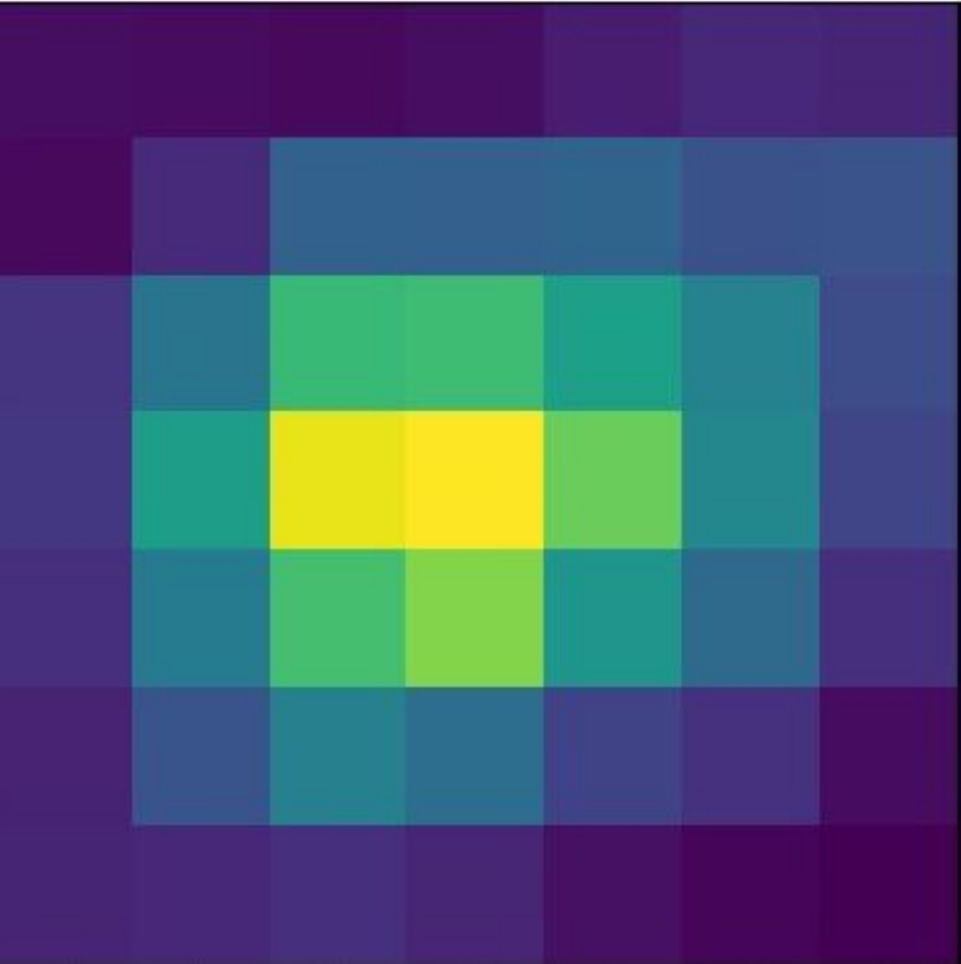} &
\includegraphics[width=0.145\linewidth,frame]{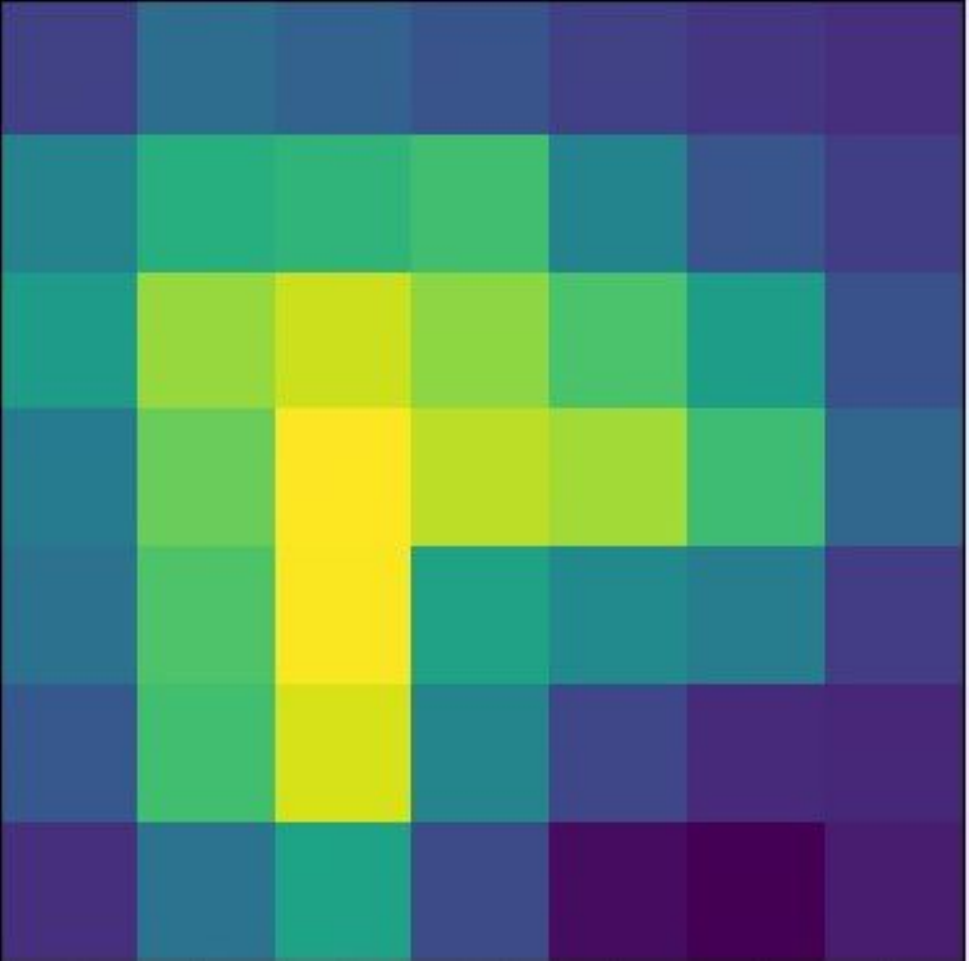} &
\includegraphics[width=0.145\linewidth,frame]{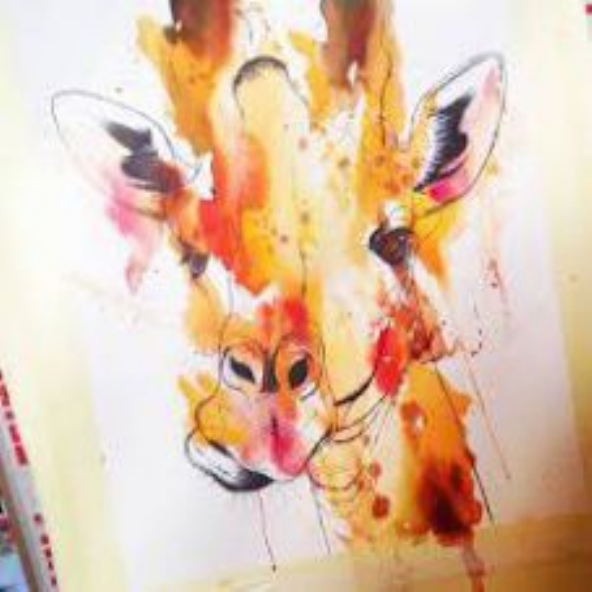} &
\includegraphics[width=0.145\linewidth,frame]{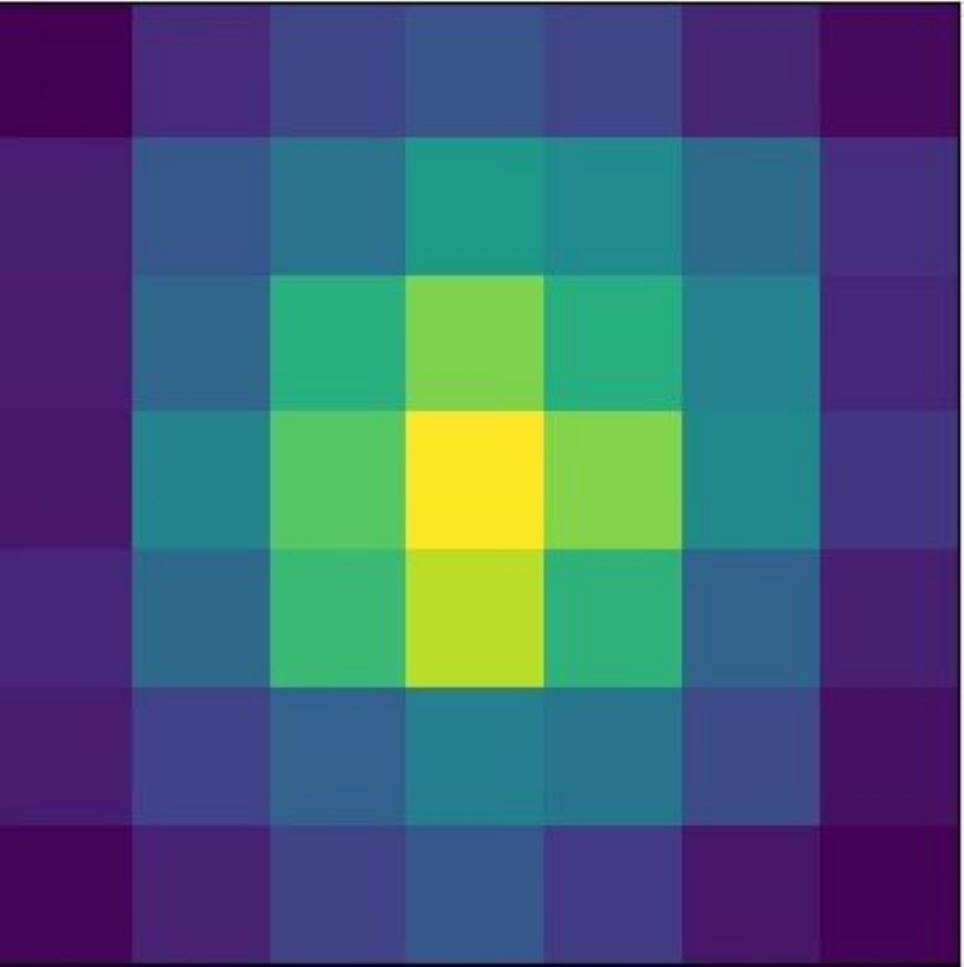}&
\includegraphics[width=0.145\linewidth,frame]{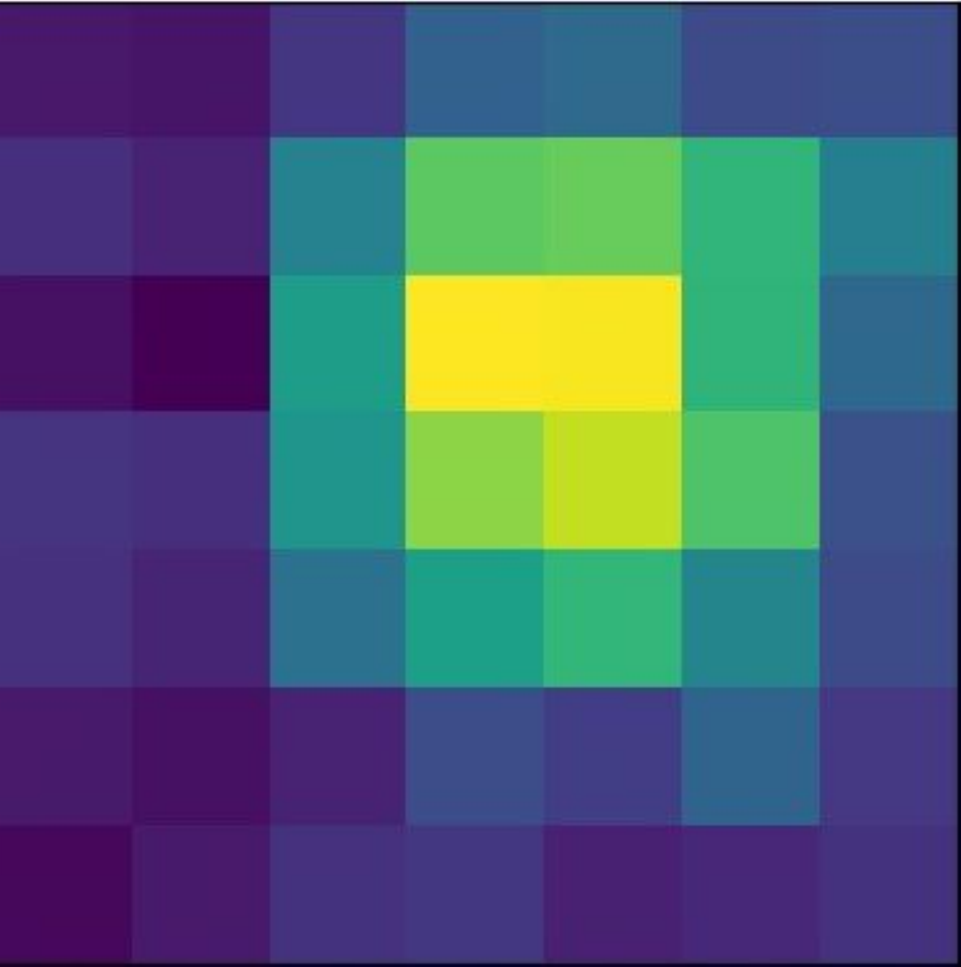} &
\includegraphics[width=0.145\linewidth,frame]{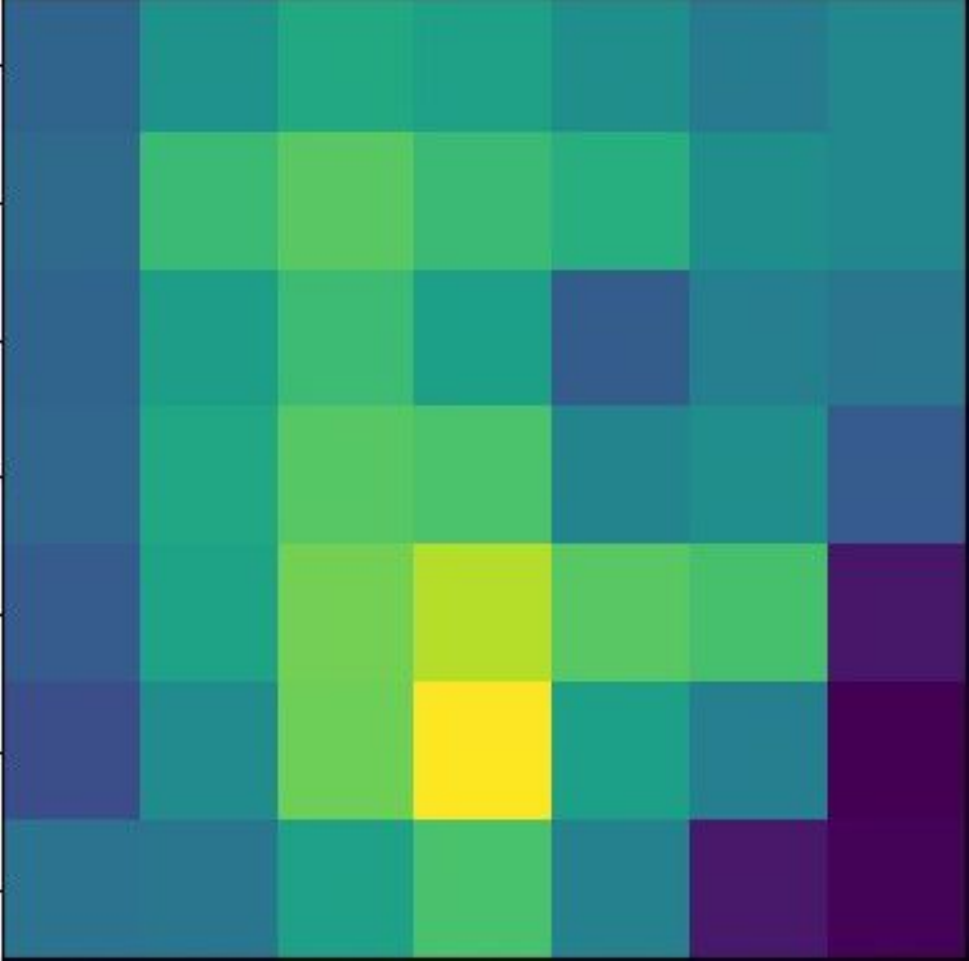}
\\
& {DeepAll \ding{55}} &
{Jigsaw \ding{51}}& 
{Rotation \ding{55}}&  &
{DeepAll \ding{55}} & {Jigsaw \ding{51}} & {Rotation \ding{55}}
& & {DeepAll \ding{51}} & {Jigsaw \ding{55}}& 
{Rotation \ding{51}}& & {DeepAll  \ding{51}} & {Jigsaw \ding{55}} &
{Rotation \ding{51}}\\
\includegraphics[width=0.145\linewidth,frame]{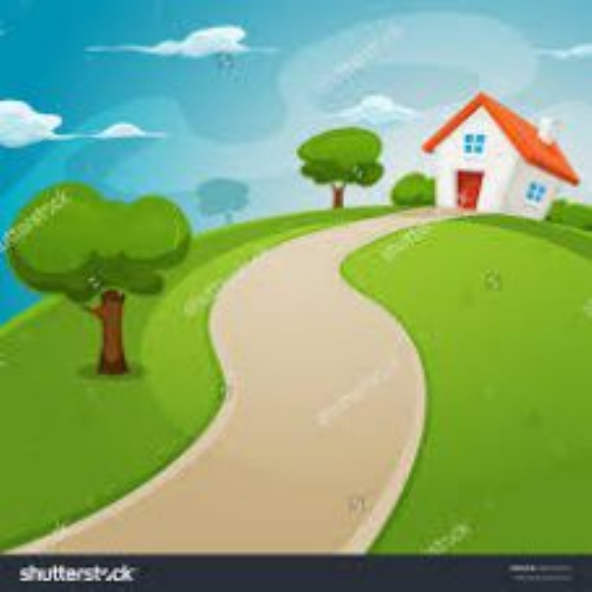} &
\includegraphics[width=0.145\linewidth,frame]{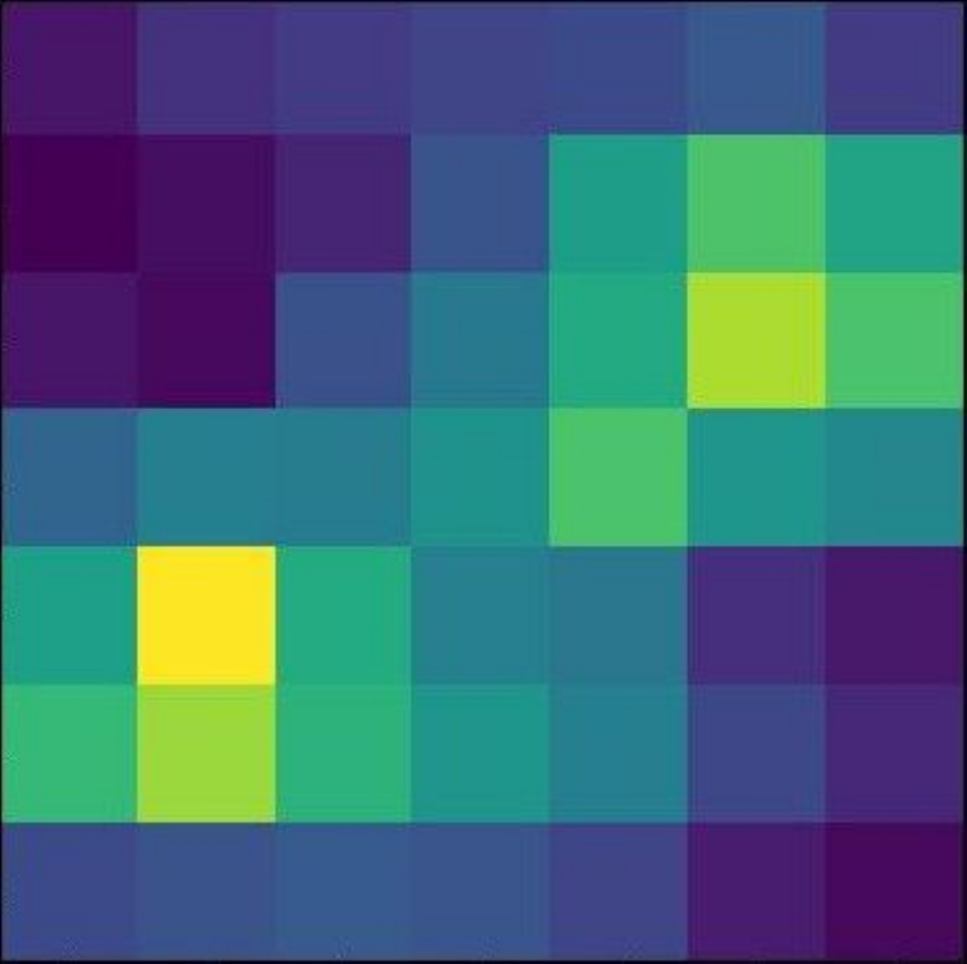} &
\includegraphics[width=0.145\linewidth,frame]{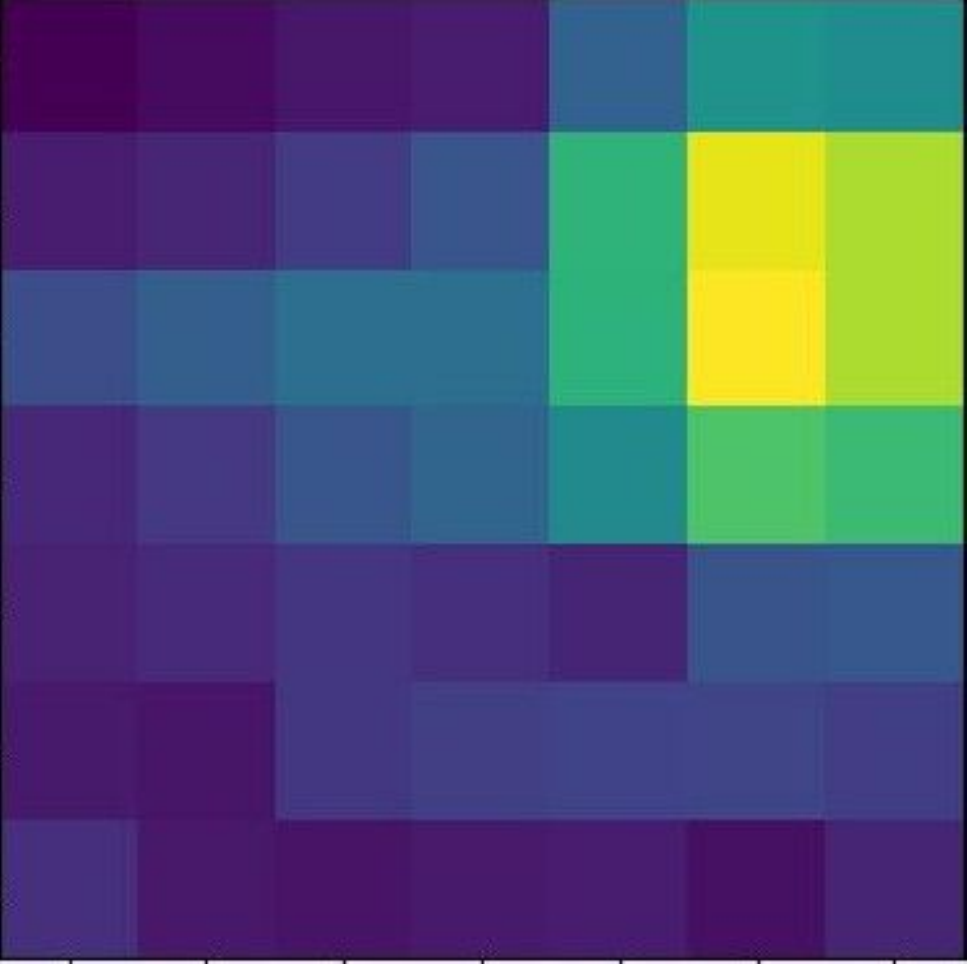} &
\includegraphics[width=0.145\linewidth,frame]{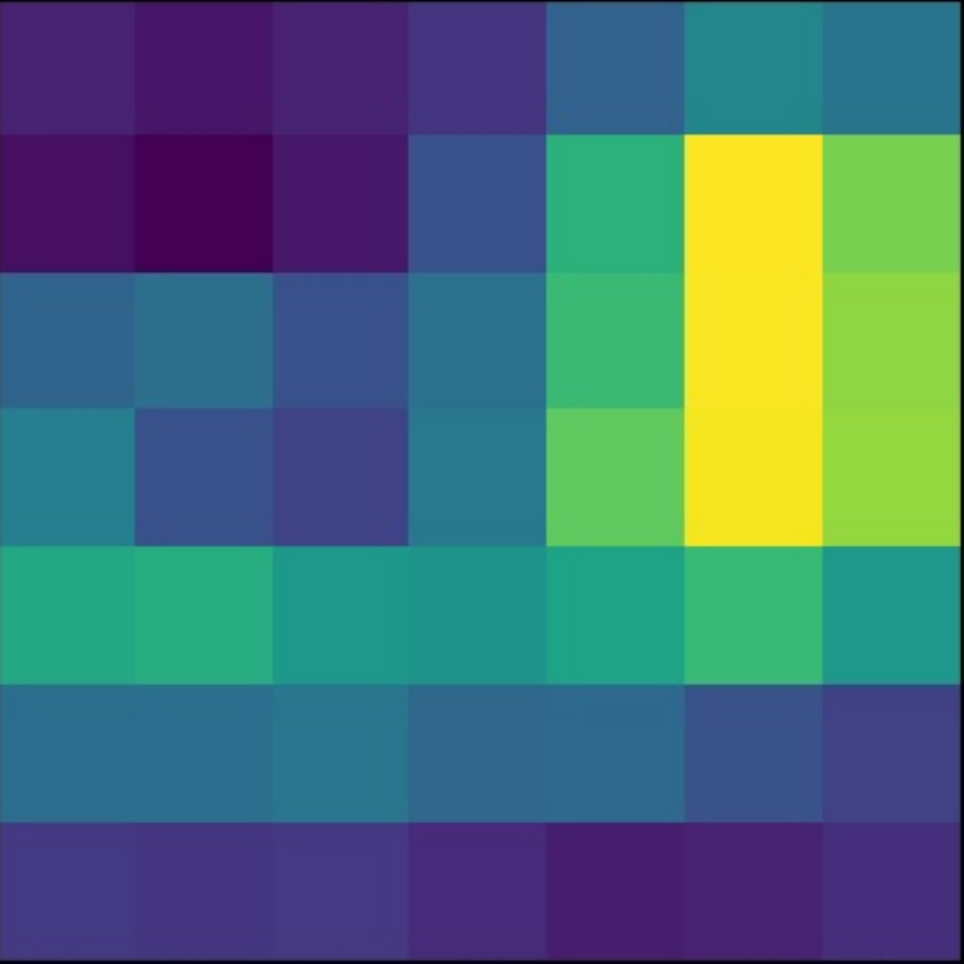} &
\includegraphics[width=0.145\linewidth,frame]{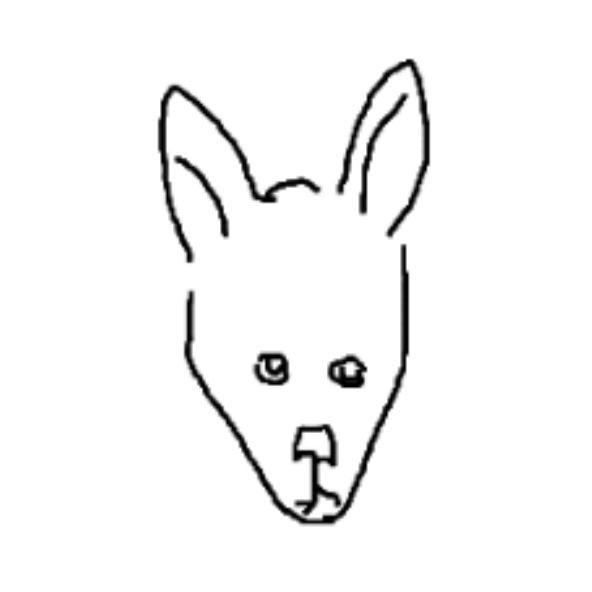} &
\includegraphics[width=0.145\linewidth,frame]{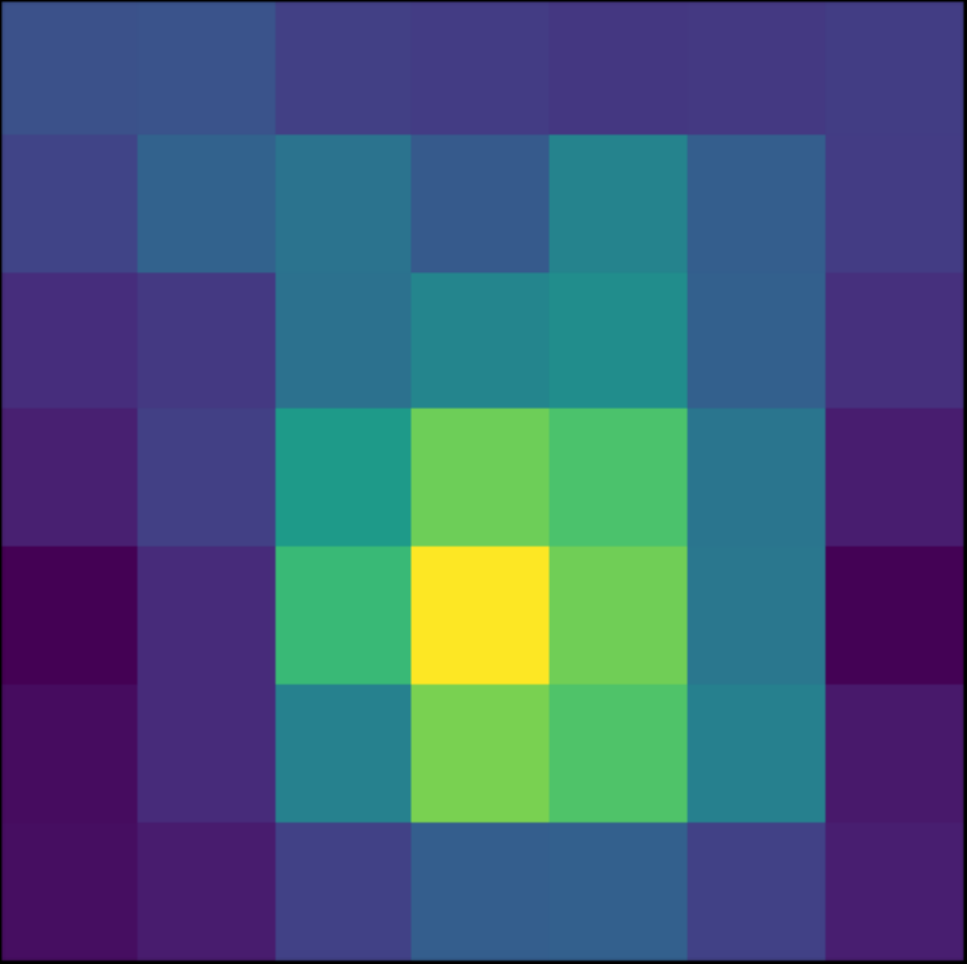} & \includegraphics[width=0.145\linewidth,frame]{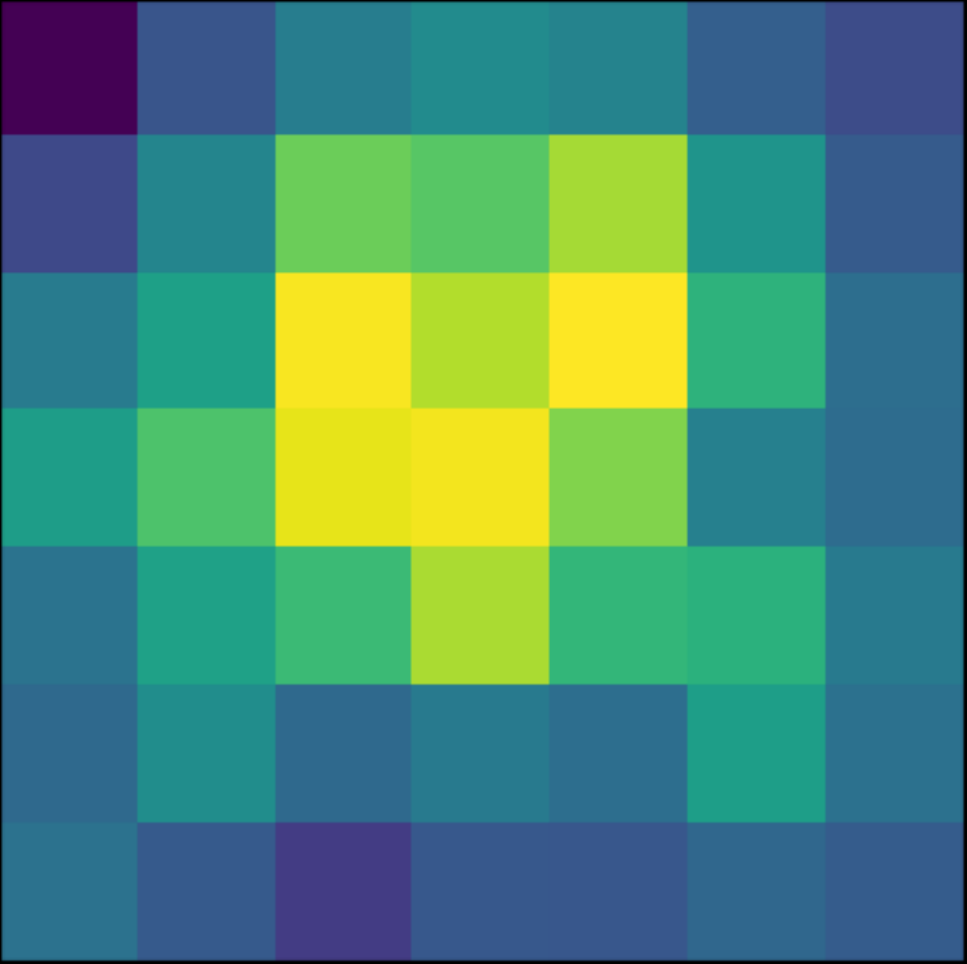} & \includegraphics[width=0.145\linewidth,frame]{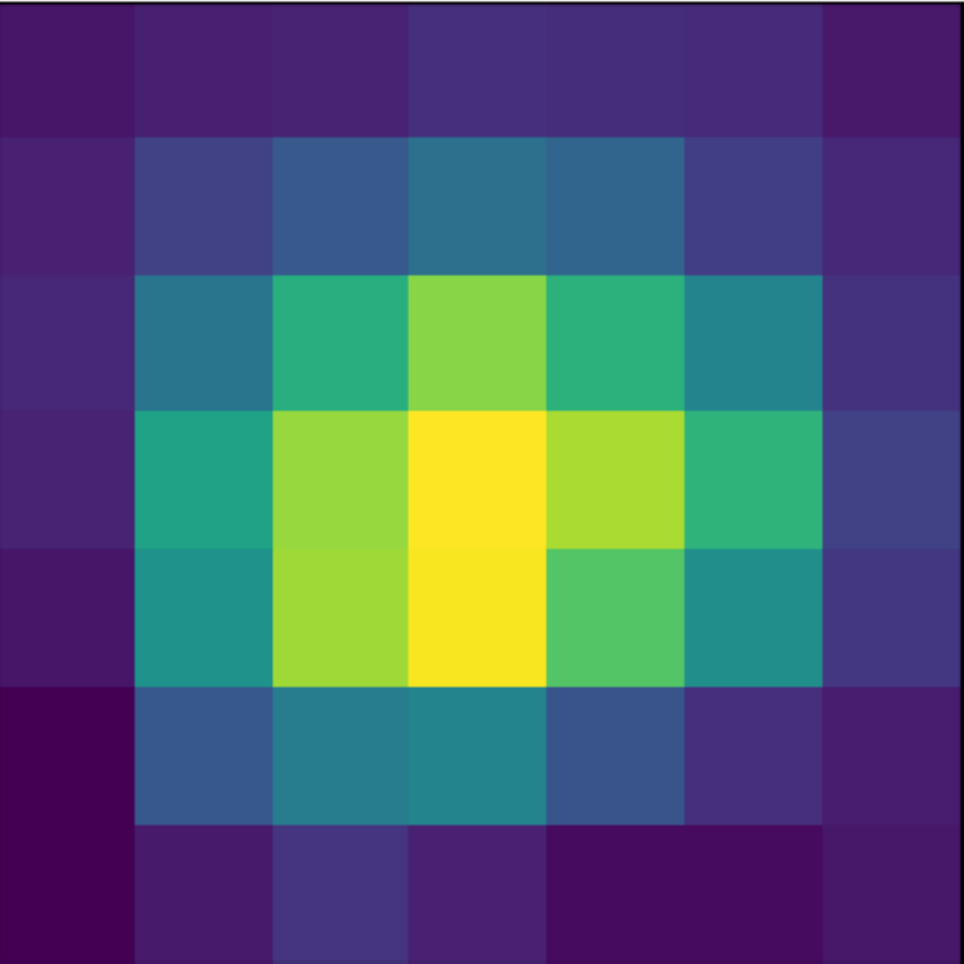}
& \includegraphics[width=0.145\linewidth,frame]{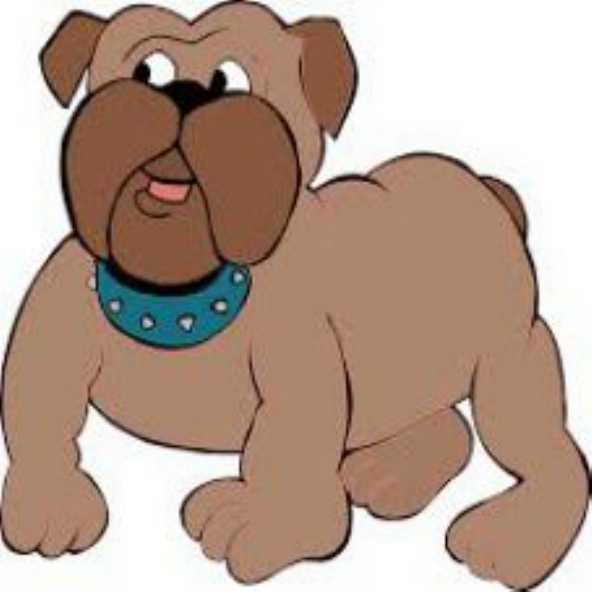} &
\includegraphics[width=0.145\linewidth,frame]{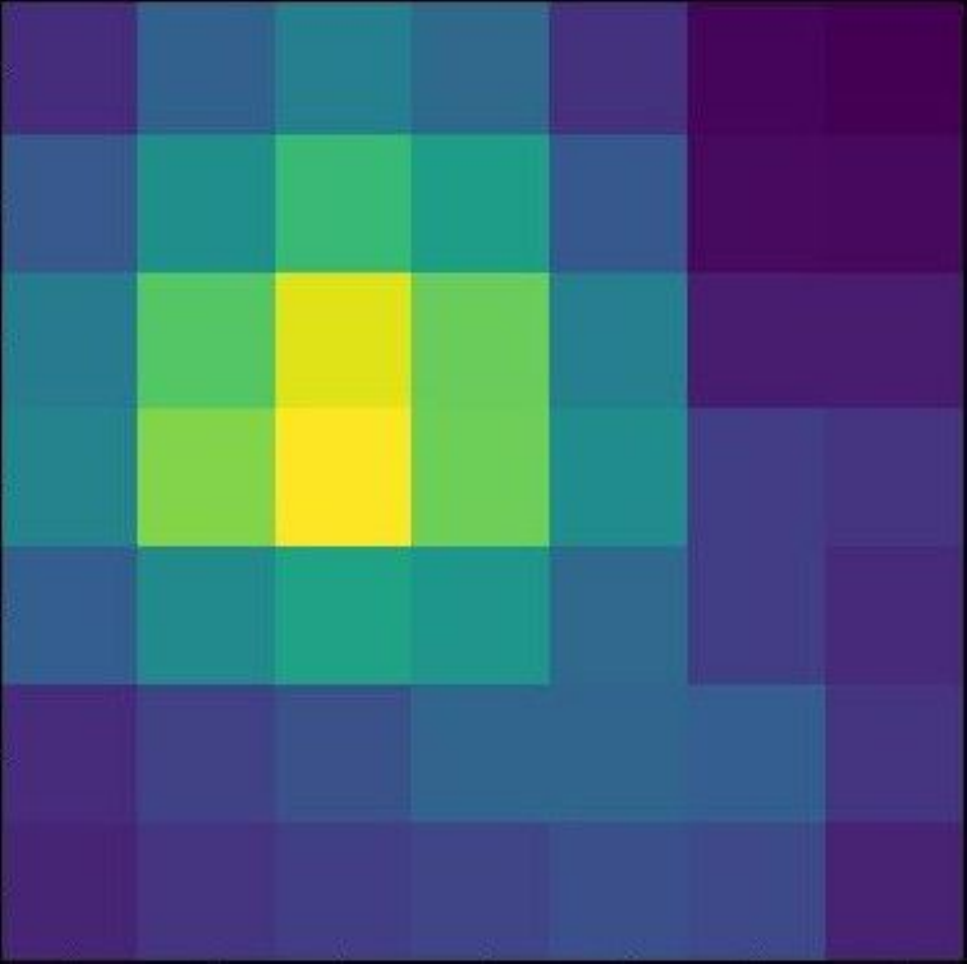} &
\includegraphics[width=0.145\linewidth,frame]{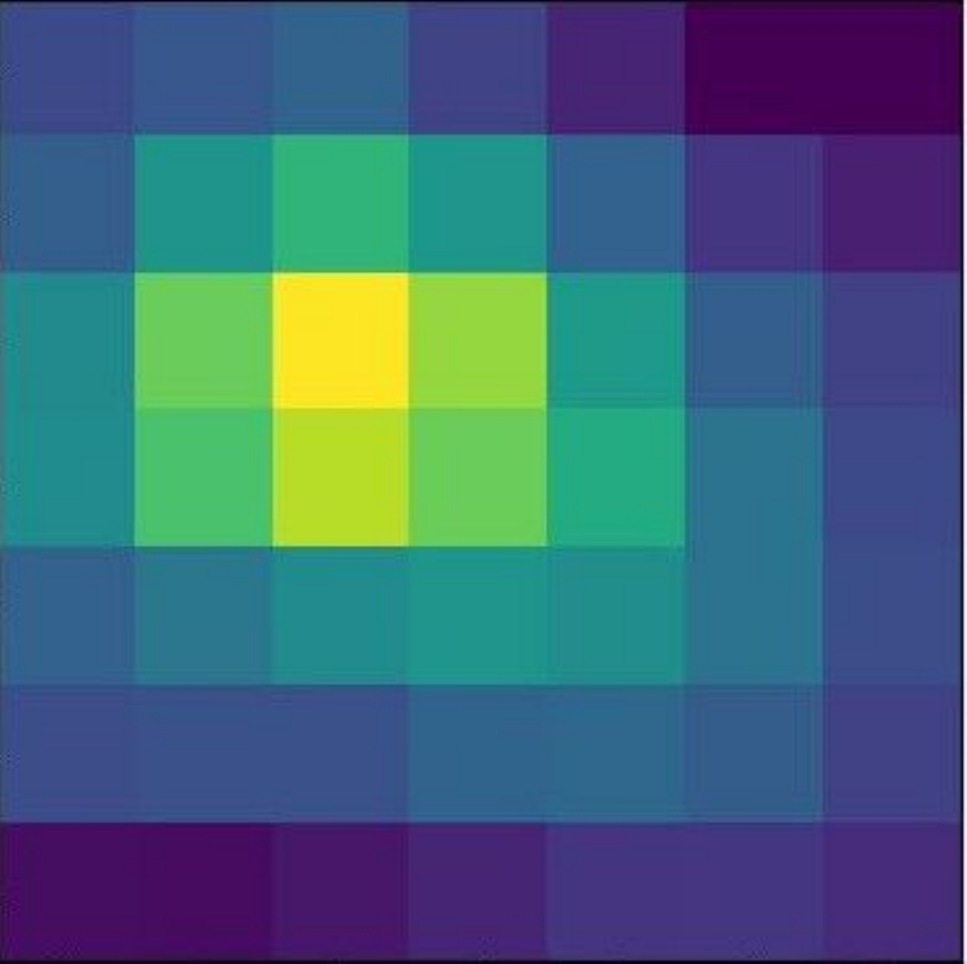} &
\includegraphics[width=0.145\linewidth,frame]{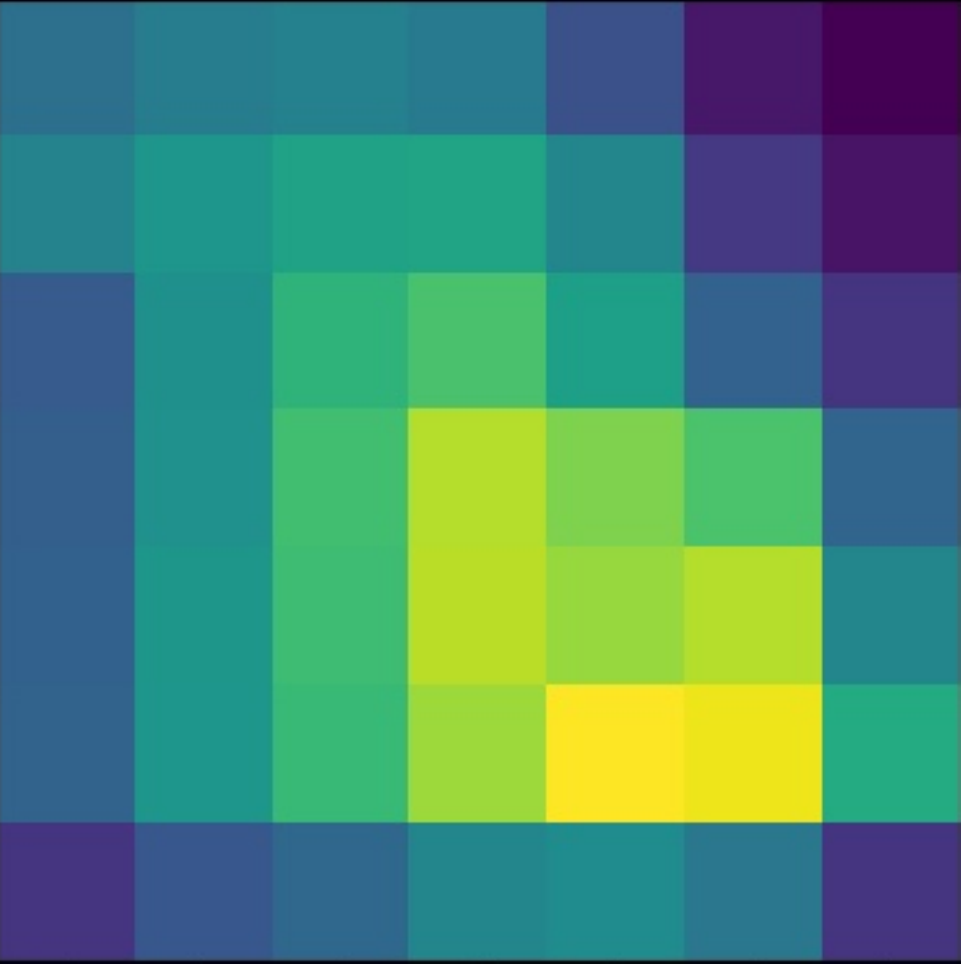} &
\includegraphics[width=0.145\linewidth,frame]{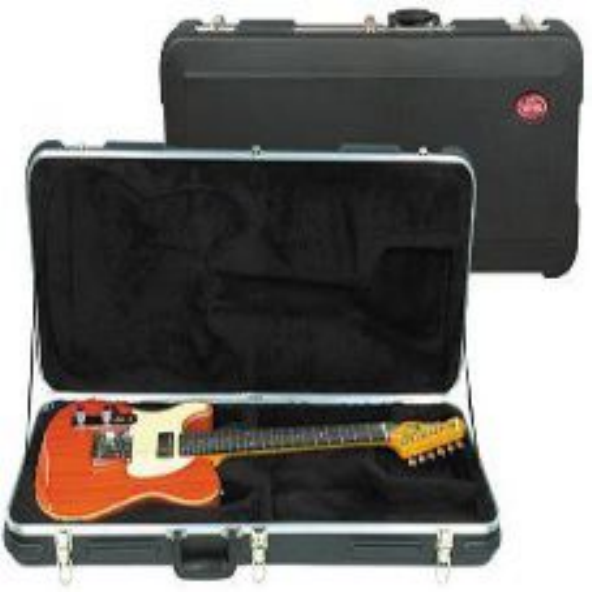} &
\includegraphics[width=0.145\linewidth,frame]{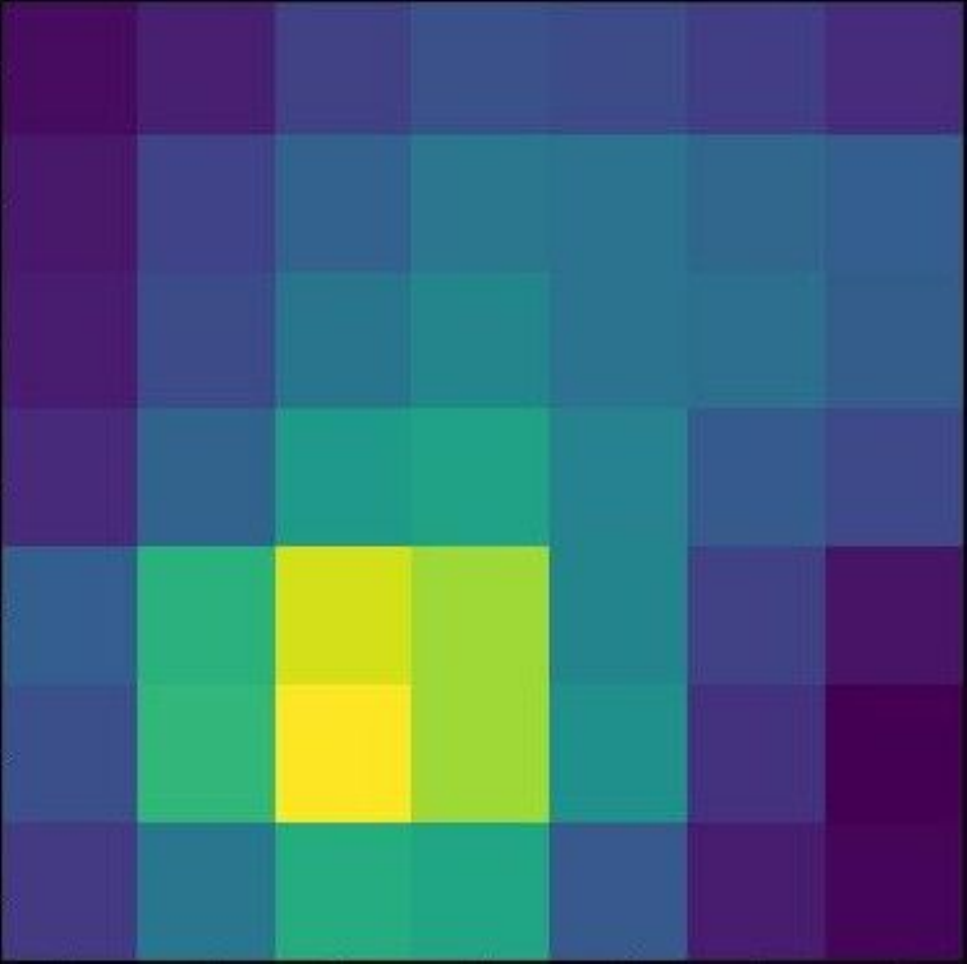}&
\includegraphics[width=0.145\linewidth,frame]{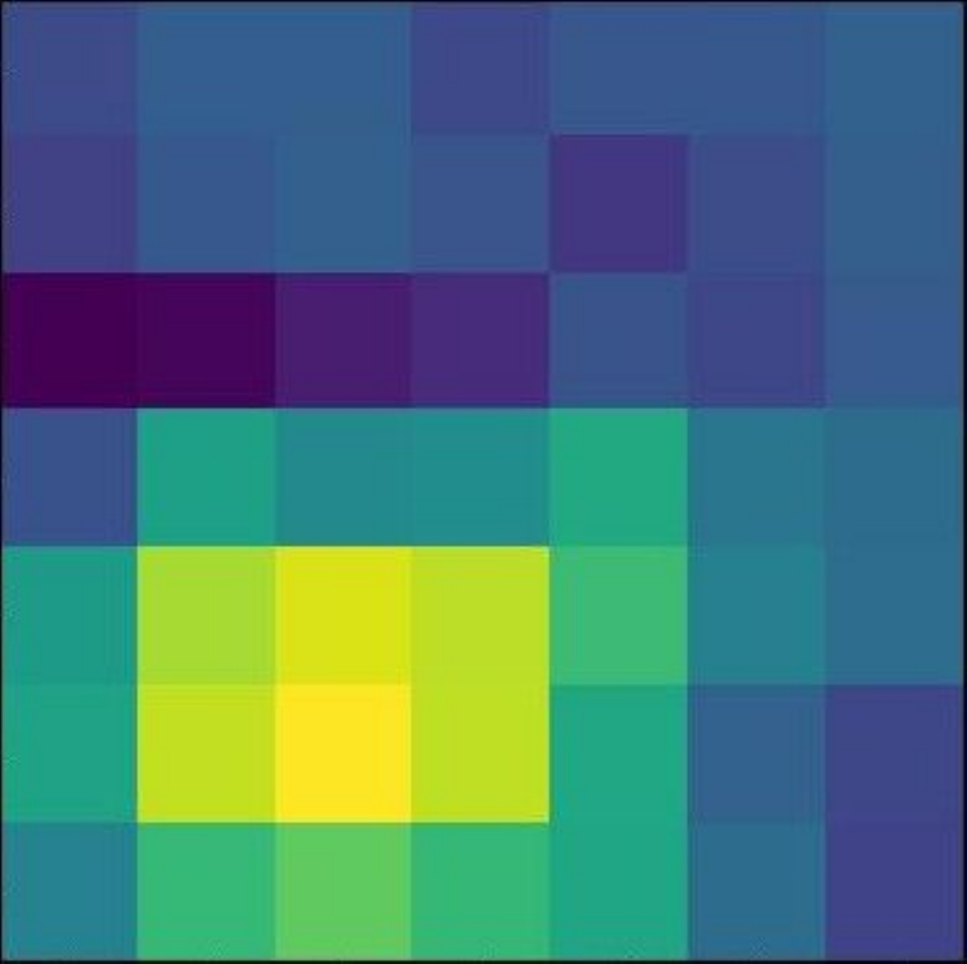} &
\includegraphics[width=0.145\linewidth,frame]{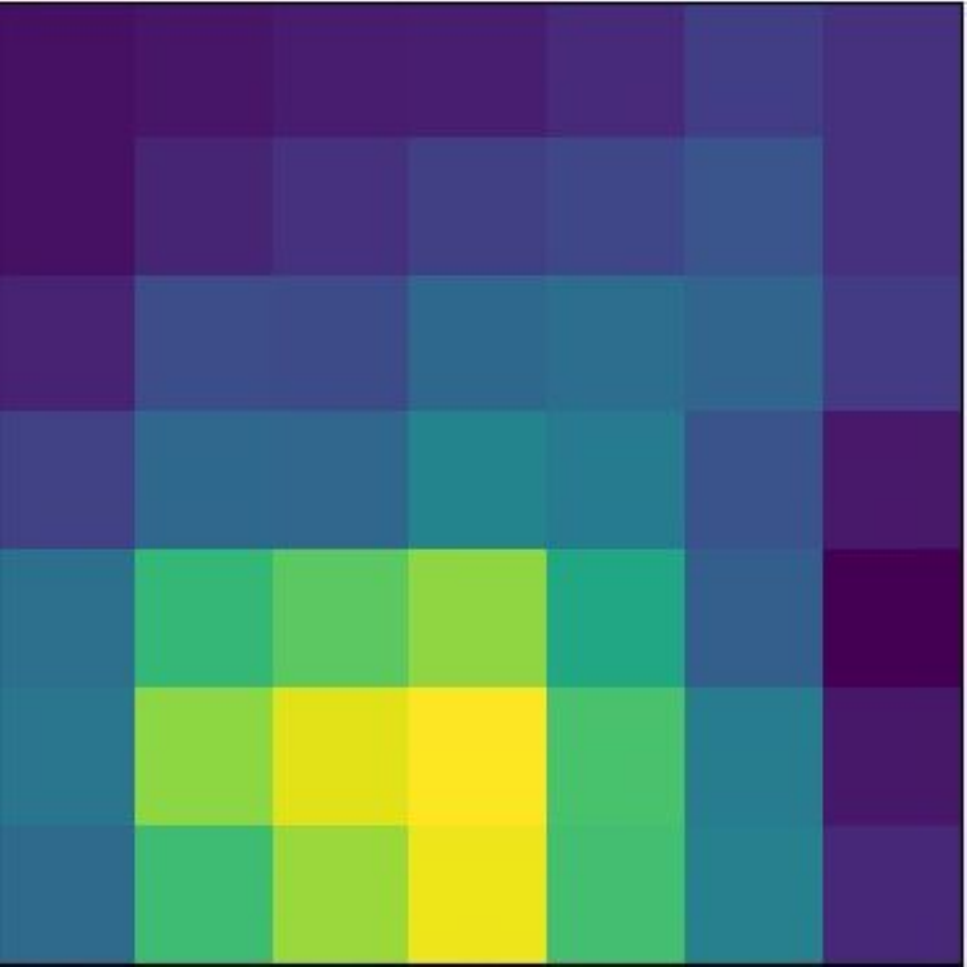}
\\ 
& {DeepAll \ding{55}} & {Jigsaw \ding{55}}& 
{Rotation \ding{51}}& & {DeepAll  \ding{55}} & {Jigsaw \ding{55}} &
{Rotation \ding{51}} 
& & {DeepAll \ding{51}} & {Jigsaw \ding{55}}& 
{Rotation \ding{55}}& & {DeepAll  \ding{51}} & {Jigsaw \ding{55}} &
{Rotation \ding{55}}
\\
\includegraphics[width=0.145\linewidth,frame]{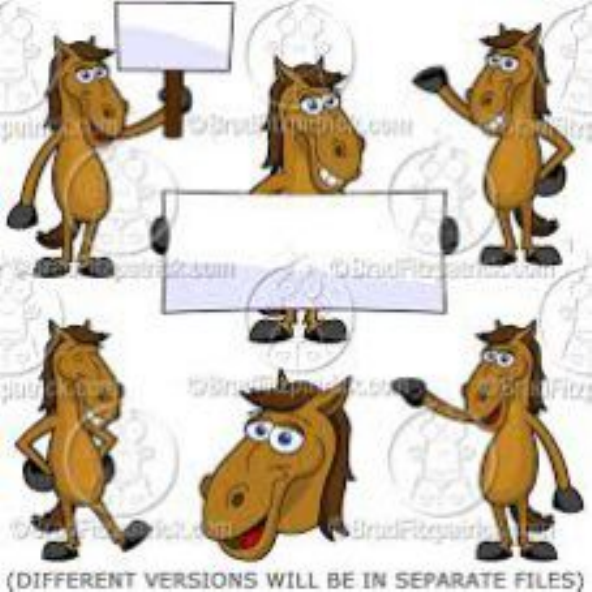} &
\includegraphics[width=0.145\linewidth,frame]{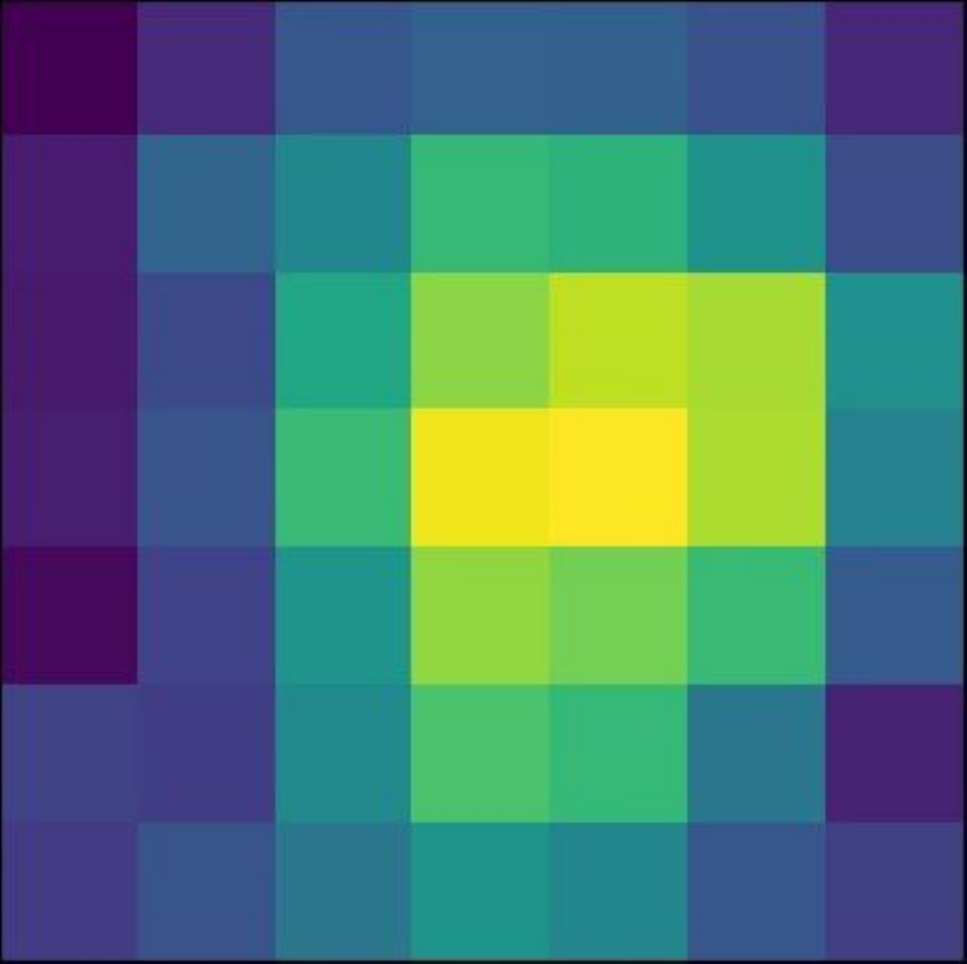} &
\includegraphics[width=0.145\linewidth,frame]{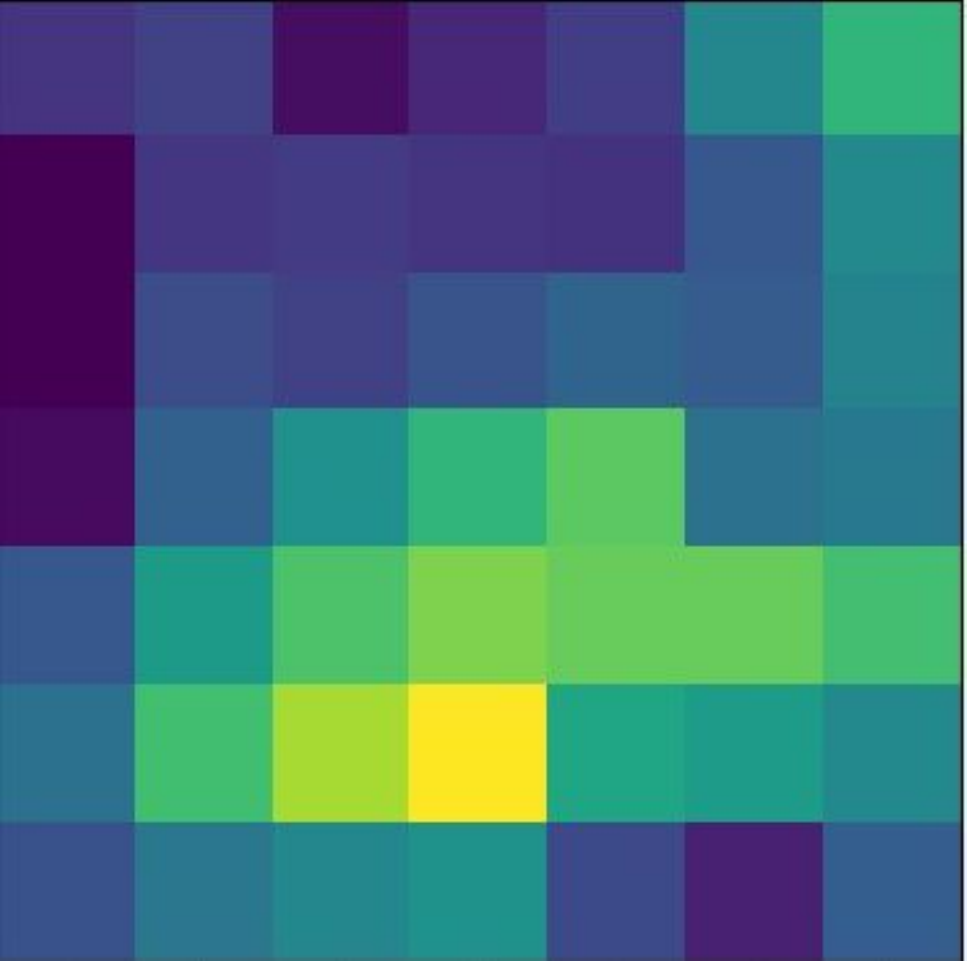} &
\includegraphics[width=0.145\linewidth,frame]{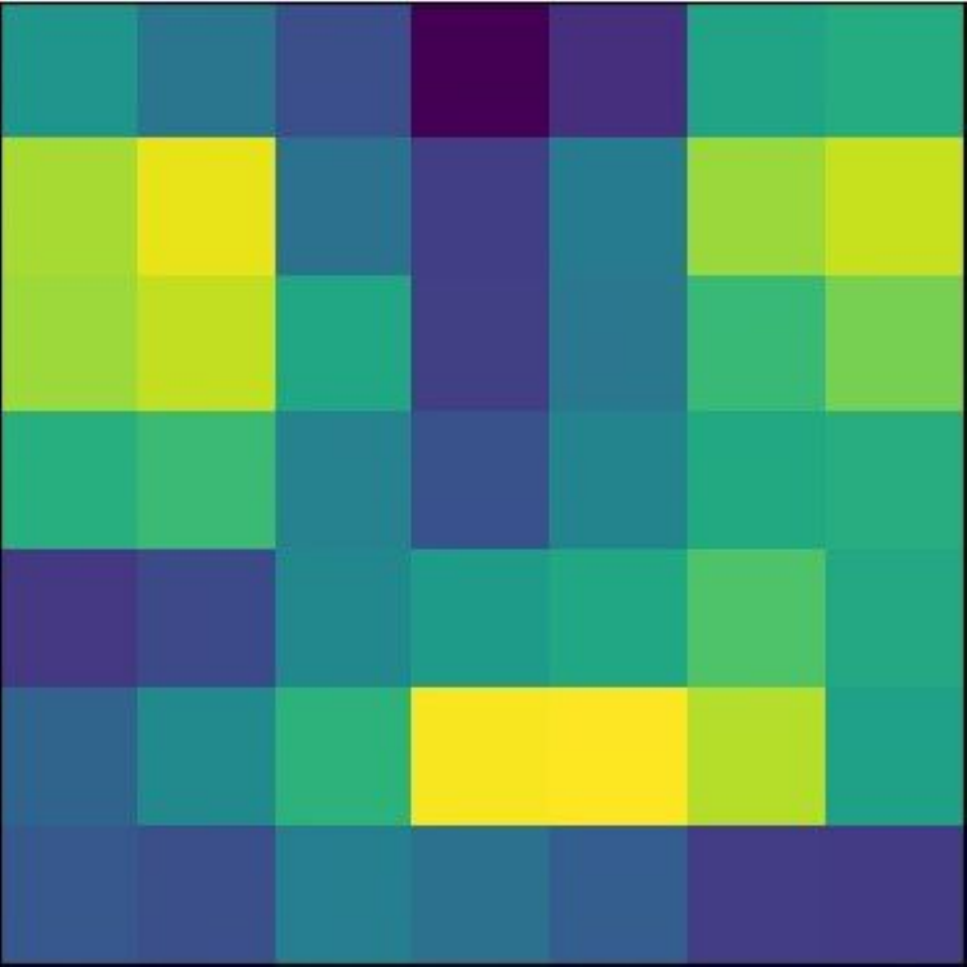} & \includegraphics[width=0.145\linewidth,frame]{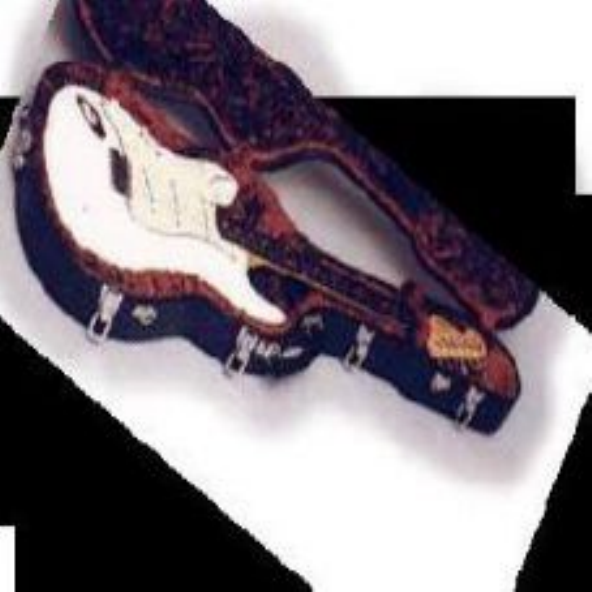} &
\includegraphics[width=0.145\linewidth,frame]{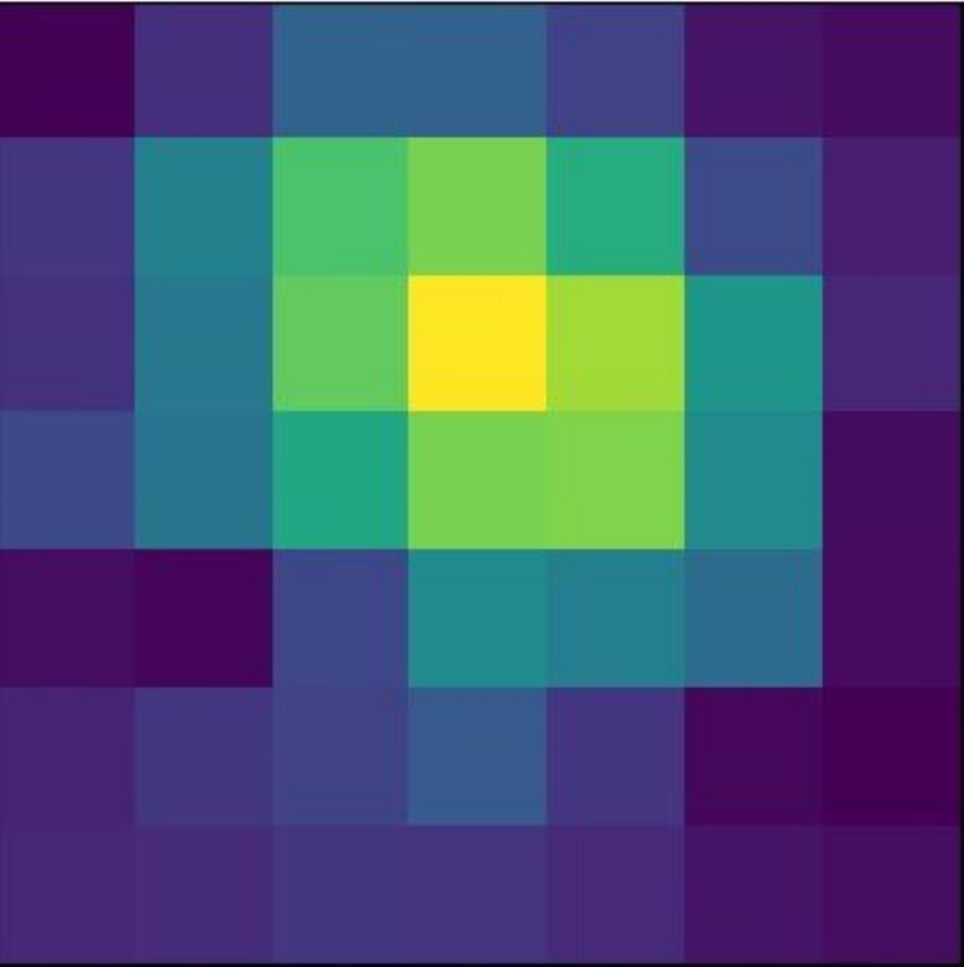}&
\includegraphics[width=0.145\linewidth,frame]{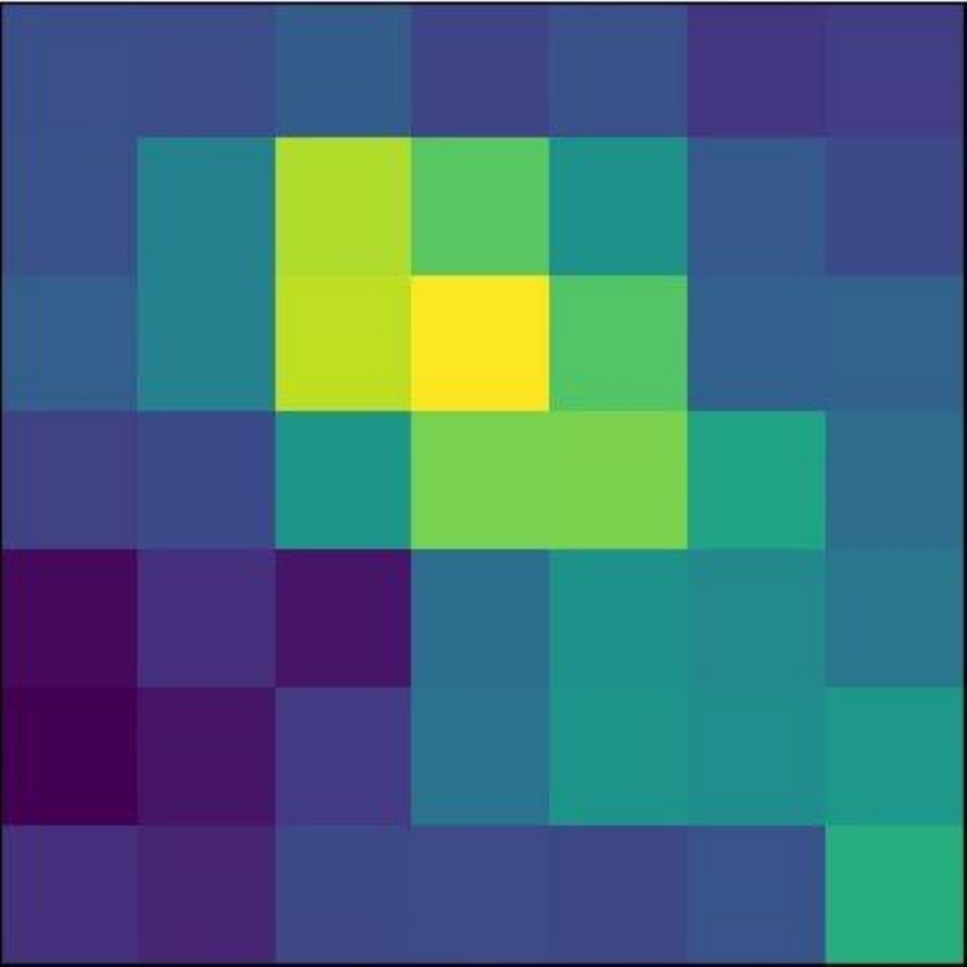} &
\includegraphics[width=0.145\linewidth,frame]{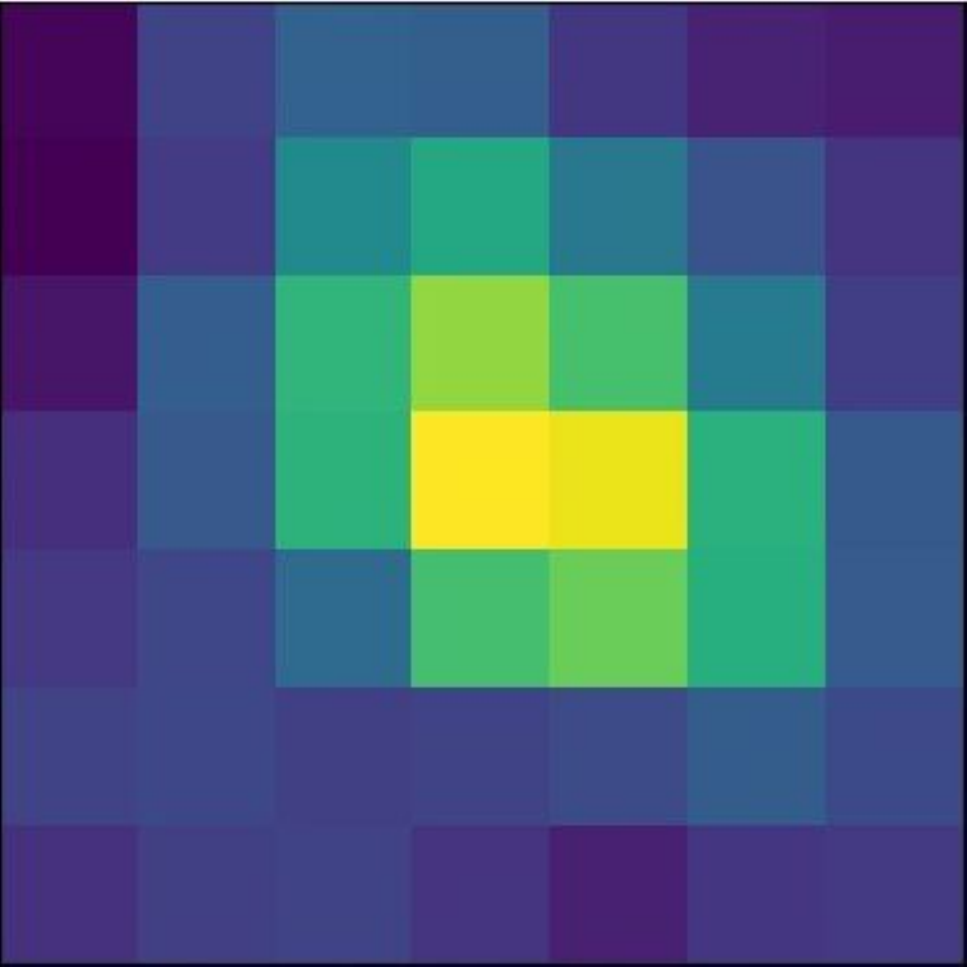}
& \includegraphics[width=0.145\linewidth,frame]{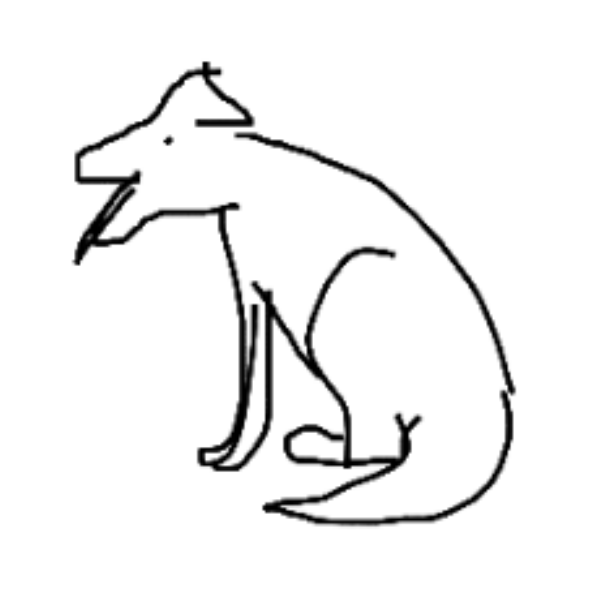} &
\includegraphics[width=0.145\linewidth,frame]{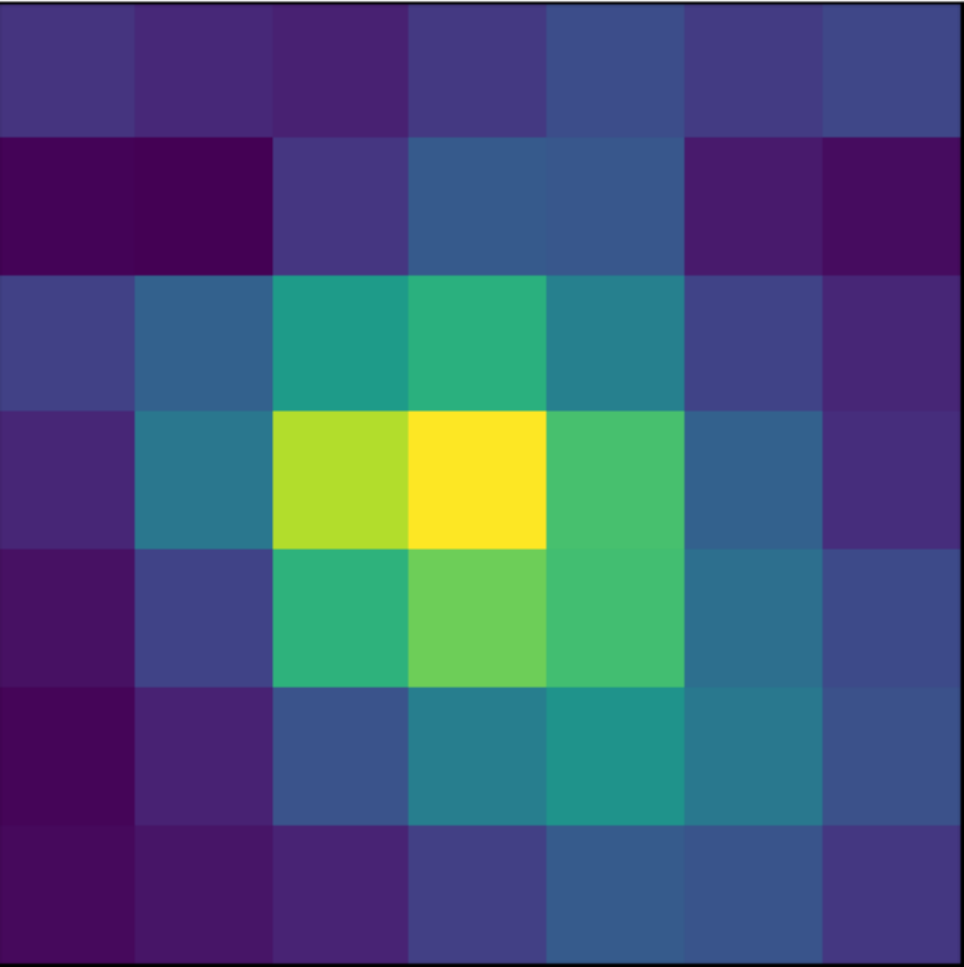} &
\includegraphics[width=0.145\linewidth,frame]{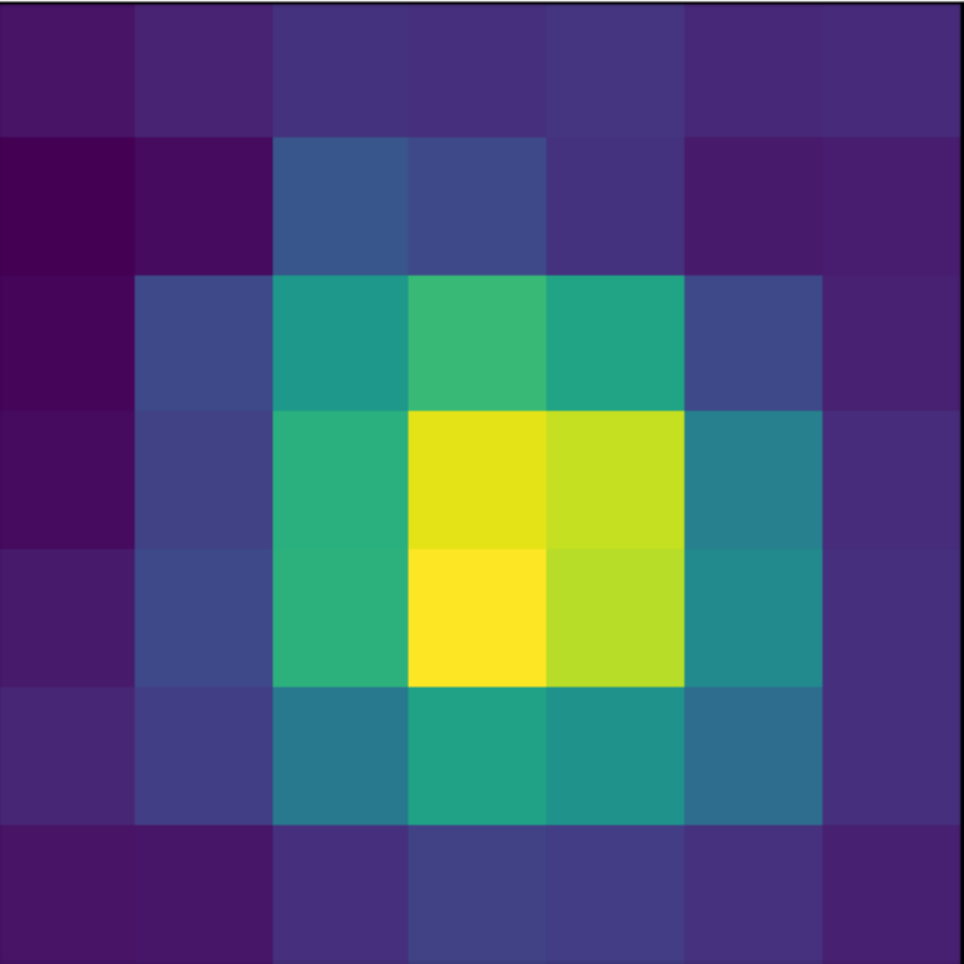} &
\includegraphics[width=0.145\linewidth,frame]{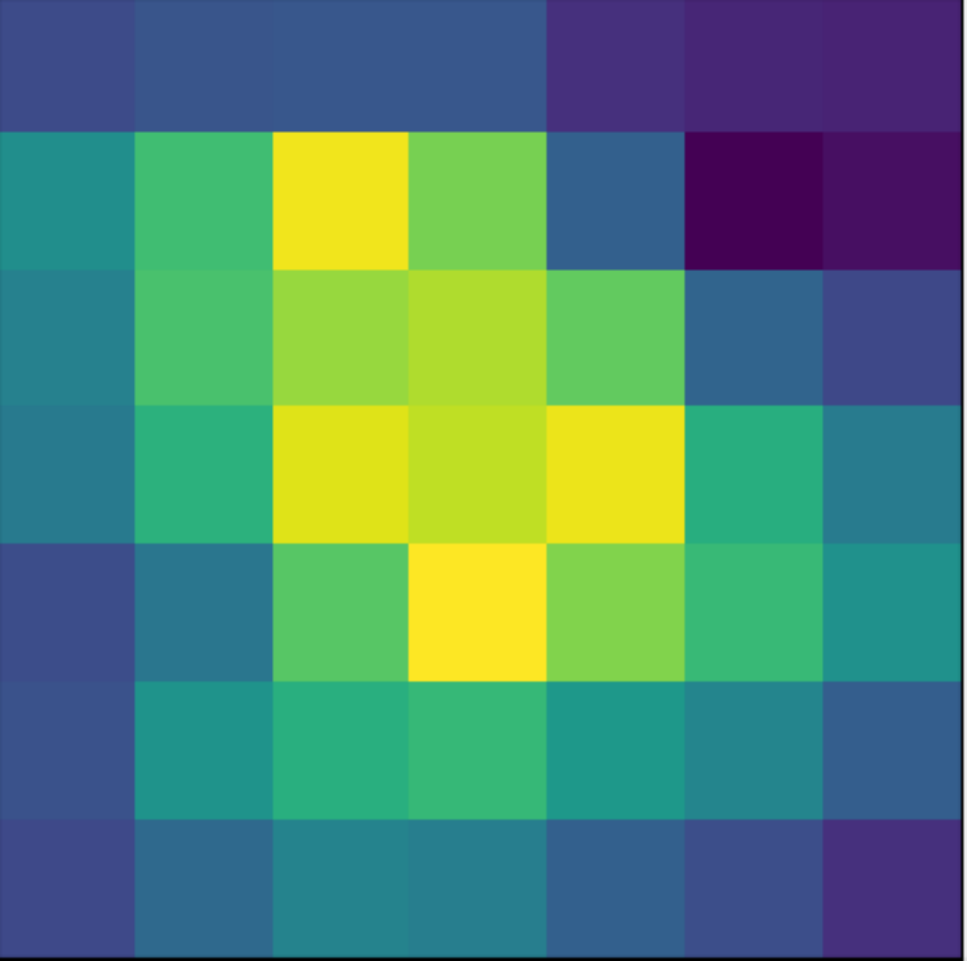} &
\includegraphics[width=0.145\linewidth,frame]{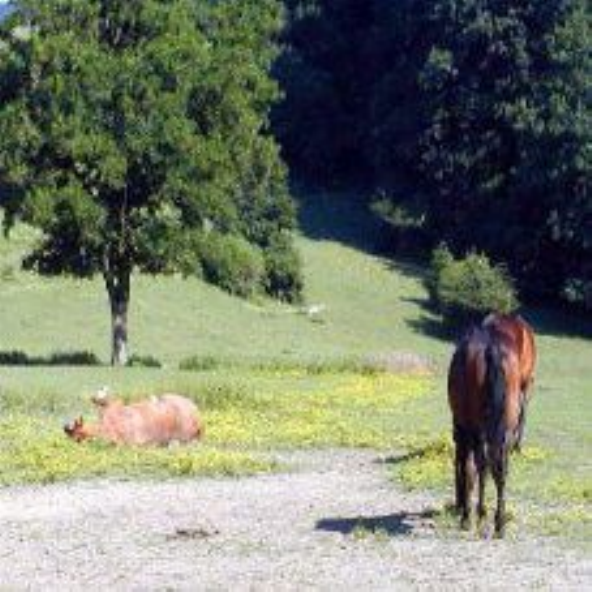} &
\includegraphics[width=0.145\linewidth,frame]{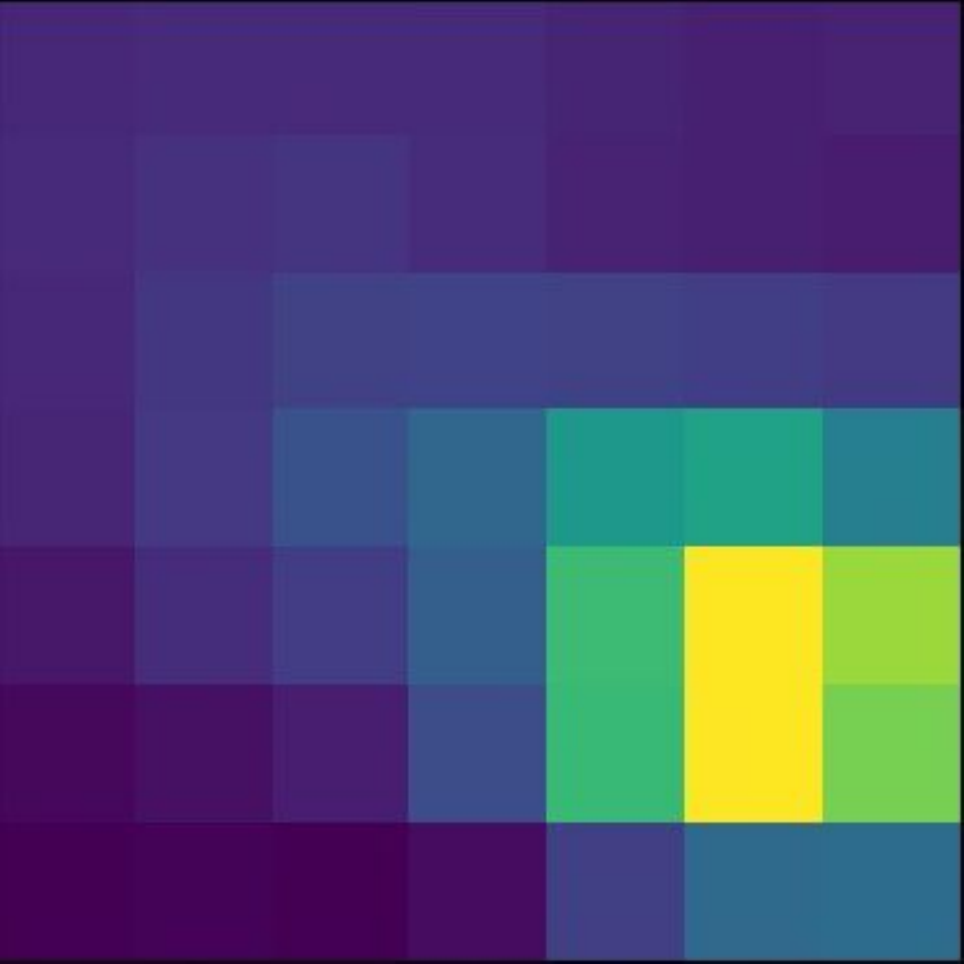}&
\includegraphics[width=0.145\linewidth,frame]{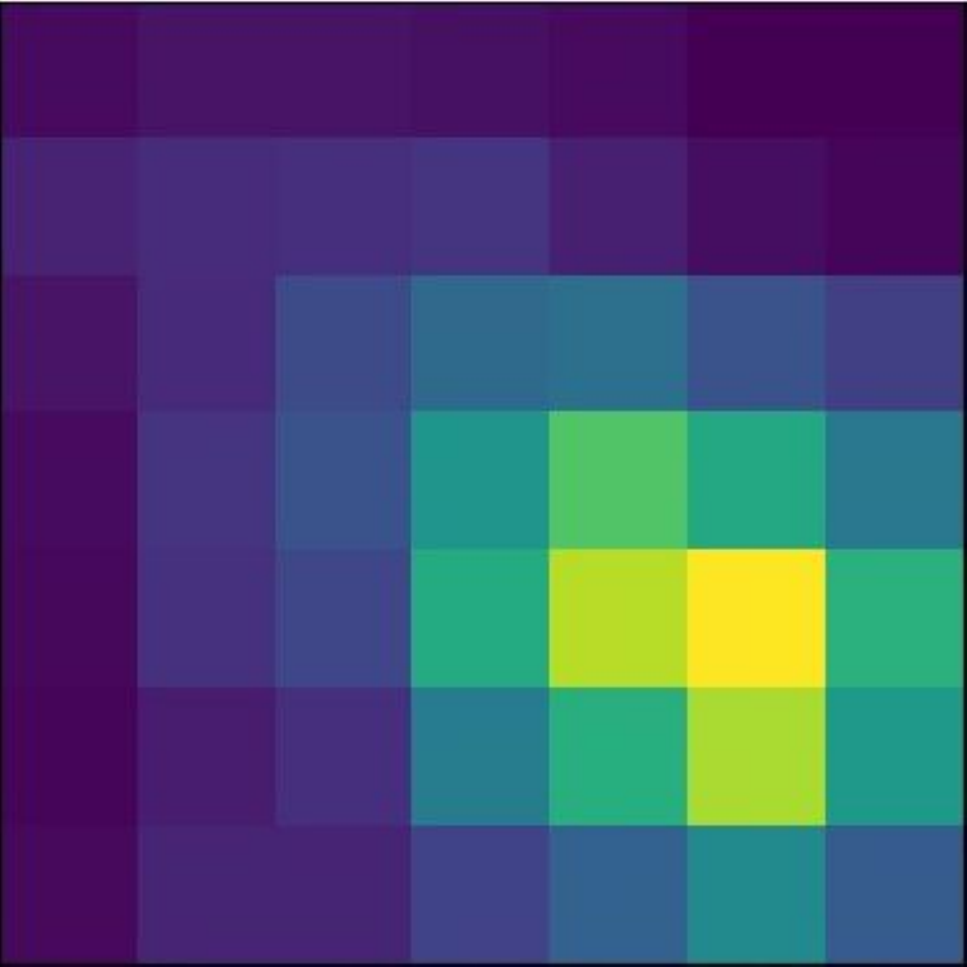} &
\includegraphics[width=0.145\linewidth,frame]{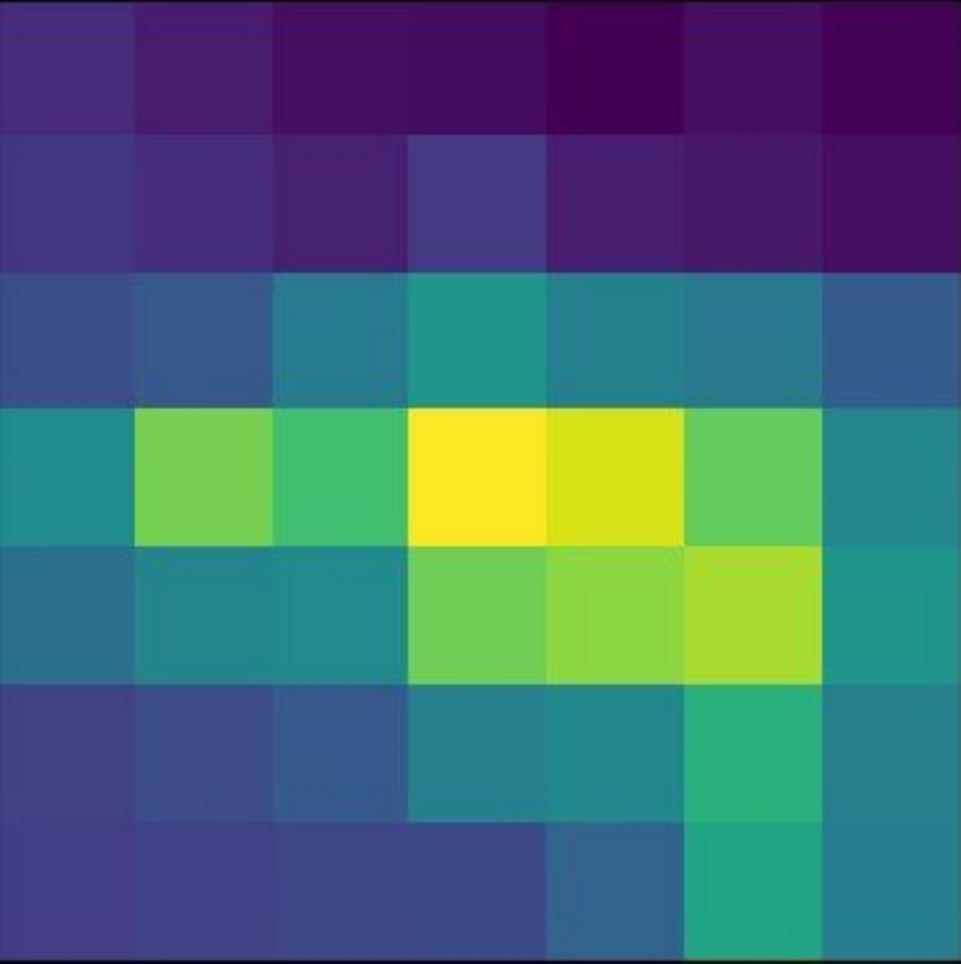}\
\\
\end{tabular}}\caption{CAM activation maps: yellow corresponds to high values, while dark blue corresponds to low values. The jigsaw puzzle task is able to localize the most informative part of the image, useful for object class prediction regardless of the visual domain. Rotation recognition has a similar effect but tend to be less precise in localization especially for sketches, cartoon and paintings.}\label{fig:examples}\vspace{-3mm}
\end{figure*}

As mentioned in Sec. \ref{subsec:implementation}, the parameters $\alpha$ and $\beta$ of our multi-task approach regulates respectively the importance of the self-supervised auxiliary loss, and the amount of samples out of each input data batch that reaches the self-supervised branch. By considering extreme cases for those parameters we obtain an ablation study on the 
respective roles of the self-supervised and of the supervised task of the learning model.
Furthermore, 
we test the robustness of our method to the number of Jigsaw classes (patch permutations) $P$, and to the dimension of the patch grid $n \times n$.

\noindent\emph{Baseline:}
for these experiments we focus on the Alexnet-PACS DG setting. We keep the Jigsaw hyper-parameters fixed with a $3 \times 3$ patch grid and $P=30$ when studying ablation. Setting $\{\alpha=0, \beta=1\}$ means that the self-supervised task is off, and the data batches contain only original ordered images, which corresponds to our DeepAll baseline.

\noindent\emph{Results - Jigsaw ablation:}
The value assigned to the data bias $\beta$ drives the training: it moves the focus from the self-supervised task when using low
values ($\beta<0.5$), to object classification when using high values ($\beta\geq0.5$). We set the data bias to $\beta=0.6$ which means that we fed the network with more ordered than shuffled images, thus keeping the classification as the primary goal of the network. In this case, when changing the loss weight $\alpha$ in $\{0.1,1\}$, we observe results
which are always either statistically equal or better than the DeepAll baseline as shown in the first plot on the left of Figure \ref{fig:ablation}.
The second plot indicates that, for high values of $\alpha$, tuning  $\beta$ has a significant effect on the overall performance.
Indeed $\{\alpha\sim1, \beta=1\}$ means that Jigsaw is on and highly relevant in the learning process, but we are feeding the network only with ordered images: in this case the puzzle task is trivial and forces the 
network to recognize always the same permutation class which, instead of regularizing the learning process,
may increase the risk of data memorization and overfitting. Further experiments confirm that, for $\beta=1$ but lower
$\alpha$ values, our multi-task method based on Jigsaw and DeepAll perform equally well. 
Setting $\beta=0$ means feeding the network only with shuffled images. For each image we have $P$ variants, only one of which has the patches in the correct order and is allowed to enter the object classifier, resulting
in a drastic reduction of the real batch size. In this condition the object classifier is unable to converge, 
regardless of Jigsaw being active ($\alpha>0$) or not ($\alpha=0$). In those cases the accuracy 
is very low ($<20\%$), so we do not show it in the plots to ease the visualization.

\noindent\emph{Results - Jigsaw hyper-parameter tuning:}
By using the same experimental setting of the previous paragraph, the third plot in Figure \ref{fig:ablation} shows the change in performance when the number of Jigsaw classes $P$ varies between 5 and 1000. We started from a low number, with the same order of 
magnitude of the number of object classes in PACS, and we grew till 1000 which is the value used
for the experiments in \cite{NorooziF16}. We observe an overall variation of 1.5 percentage points
in the accuracy which still remains almost always higher than the DeepAll baseline. 
Finally, we ran a test to check the accuracy when changing the grid size and consequently the patch number. Even in this case, the range of variation is limited when passing from a $2\times2$ to a $4\times4$ grid, confirming the conclusions of robustness already obtained for this parameter in \cite{NorooziF16} and \cite{Cruz2017}. Moreover all the results are better than DeepAll.

\noindent\emph{Results - Rotation ablation:} Changing the orientation has a milder effect on the global appearance of the image with respect to patch decomposition and puzzle reordering. One significant further difference between the Rotation and Jigsaw tasks is in the number of self-supervised classes which is $P\sim10-50$ for Jigsaw and just $4$ for Rotation, which actually reduces to $3$ if we consider that one of the classes matches with the original image orientation. In this conditions, even using a low $\beta=0.4$ does not distract the network focus from the main object classification task and, combined with $\alpha=0.4$ produces the results reported in Table \ref{table:resultsDG_PACS}. 
For the ablation analysis we keep each of the two parameters fixed while varying the other: the results are always above the DeepAll baseline and on average the performance variation is limited (around 1 percentage point) indicating low sensitivity to the specific parameter settings. 

\noindent\emph{Results - self-supervised performance:}
We have seen how the self-supervised tasks support the main supervised classifier for domain generalization, but it is also interesting to check their own internal functioning and whether those tasks get meaningful results. We show their performance when 
testing on the same target images used to evaluate the object classifier but with shuffled patches for Jigsaw and randomly changed orientation for the Rotation task. In Figure \ref{fig:ablation_jigsaw}, the first plot shows the accuracy over the learning epochs for the Object, Rotation and Jigsaw classifiers indicating that it grows for all of them simultaneously (on different scales).
The second plot shows the Jigsaw recognition accuracy when changing the number of permutation classes $P$: of course the performance decreases when the task becomes more difficult, but overall the obtained results indicate that the Jigsaw model is always effective in reordering the shuffled patches.

\subsubsection{Visual Explanation and Failure Cases}
\label{sec:exp_DG_5}
{As highlighted in \cite{jenni2020steering}, supervised deep models tend to focus too much on local image statistics, which limits the generalization and robustness properties of the learned representation. The jigsaw puzzle and the rotation recognition task, by forcing the network to use the whole image, allow to capture global information and to identify domain agnostic object shapes. By combining the supervised and self-supervised objectives we aim at learning a representation better able to capture discriminative cues, helpful in recognizing the image object content across domains.}
To analyse this behaviour we used the Class Activation Mapping (CAM, \cite{zhou2016cvpr}) method on ResNet-18 DG experiments, with which we produced the activation maps in Fig. \ref{fig:examples} for the PACS dataset.
The first two rows show that our multi-task approach with Jigsaw or Rotation self-supervision is better at localizing the object class with respect to DeepAll. 
Rotation seems slightly less precise than Jigsaw in capturing the object shapes especially when dealing with sketches (see the dog on the second and sixth row), cartoon and paintings (fourth and fifth row), while works reasonably well with photos. The last two rows indicate that for both  Jigsaw and Rotation the recognition mistakes are related to some flaw in data interpretation, while the localization remains meaningful.

\subsubsection{Predictive Domain Adaptation}
\label{sec:exp_DG_6}

\begin{figure}[!t]
    \centering
\includegraphics[width=0.5\textwidth]{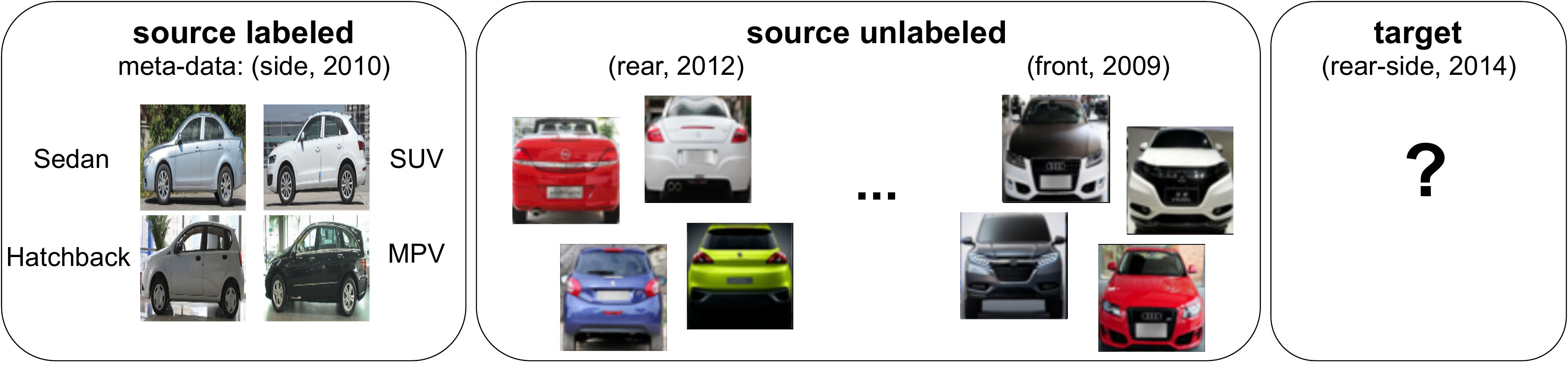}   \vspace{-6mm}
    \caption{Scheme of the Predictive DA setting. The goal is to recognize the four types of car, while the view point and the year are the meta-data.}
    \label{fig:compcars}\vspace{-3mm}
\end{figure}

Recent works have investigated intermediate settings between DG and DA. {In PrDA \cite{multivariatereg,adagraph} one labeled and several unlabeled source domains are available at training time, together with their descriptive meta-data which are a very specific kind of domain labels (see Fig. \ref{fig:compcars})}. 
The meta-data of the target is also available: they  can be used to relate the target domain to the known sources and compose a target model. Since this model is obtained without having access to the target images we are still in the DG scenario. However, the task is clearly simplified because the level of domain similarity between the sources and the target is known a priori.

{We believe it is worthwhile to evaluate our multi-task method in this scenario for two main reasons. 
(1) 
Most of the existing DG methods urge both domain and class labels of training data to work. PrDA techniques can be considered as DG methods with reduced needs in terms of class supervision, but strongly dependent on the availability of domain labels. 
By leveraging on self-supervision, our multi-task approach can work with a limited set of annotated source samples, as in AdaGraph \cite{adagraph}, but it is also completely free from the need of source (and target) domain labels. Thus, it is much cheaper in terms of manual annotations and would still be reliable in case of missing or noisy domain labels.
(2) The existing PrDA testbeds focus on fine-grained classification tasks, thus allowing us to evaluate our method on a recognition problem significantly different with respect to that offered by the standard DG datasets.
}

\noindent\emph{Baseline and Dataset:}
We use as baseline the source-only case which learns from the single labeled source and cannot exploit unlabeled data. {We also consider a state of the art semi-supervised method based on Label Propagation (LP, \cite{Iscen_2019_CVPR}) and the Minimum Class Confusion multi-target approach (MCC, \cite{jin2020minimum}). LP uses the unlabeled images in the learning process via pseudo-labeling, while they are considered as a temporary target data for MCC that adapts on them and finally uses the obtained model on the real target.} Finally, {AdaGraph} \cite{adagraph} is our main PrDA reference. It is a very recent approach that 
exploits domain-specific batch-normalization layers to learn models for each source domain in a graph, where the graph is provided on the basis of the source auxiliary meta-data. 
We follow the experimental protocol described in \cite{adagraph} on the Comprehensive Cars (CompCars) dataset \cite{Yang_2015_CVPR}. We used a subset of 24,151 images with 4 categories (MPV, SUV, sedan and hatchback) which are type of cars  produced  between  2009
and  2014  and  taken  under  5  different  view  points  (front, front-side,  side,  rear,  rear-side). 
Each  view point  and  each  manufacturing  year  defines a  separate  domain and specifies its meta-data, leading to a total of 30 domains. We selected a pair of domains as source and target and use the remaining 28 as auxiliary unlabeled sources. Considering all possible domain pairs, we got 870 experiments and observe the average accuracy results over all of them. 
More in details, we started from an Imagenet pretrained model and trained for 6 epochs on source domain using Adam as optimizer with weight decay of $10^{6}$. The batch size used is 16 and the learning rate is $10^{-3}$ for the classifier and $10^{-4}$ for the rest of the network; the learning rate is decayed by a factor of 10 after 4 epochs. We tried both Jigsaw and  Rotation with loss weight parameter set to $\alpha=0.5$.

\noindent\emph{Results:} 
{Table \ref{table:resultsPredictiveDA} collects the obtained results and shows that our multi-task approach significantly improves over the source-only baseline, as well as over LP and MCC.}
AdaGraph, which leverages on both the meta-information and the unlabeled data, shows the top result. Considering the limited gap between AdaGraph and our Jigsaw based result, 
we claim that when the meta-data information is noisy or missing, our approach can be used as reliable and inexpensive fallback.

\begin{table}[t!]
\caption{Predictive DA results.  The top result is highlighted in bold.
}\vspace{-8mm}
\begin{center} 
\resizebox{0.5\textwidth}{!}{
\begin{tabular}{c@{~~~}c@{~~~}c@{~~~}c@{~~~}c@{~~~}c@{~~~}c}
\hline
\multicolumn{7}{@{}c@{}}{\textbf{Resnet-18}}\\
\hline
 & {Baseline} & {LP} & {MCC} &  AdaGraph & \multirow{2}{*}{\textbf{Jigsaw}} & \multirow{2}{*}{\textbf{Rotation}}\\ 
 & Souce-Only & {\cite{Iscen_2019_CVPR}} & {\cite{jin2020minimum}} &  \cite{adagraph} & & \\\hline
CompCars & 56.80 & {57.91} & {59.00} & \textbf{65.10} & 63.00 & 61.77\\
\hline
\end{tabular}}
\label{table:resultsPredictiveDA}
\end{center}\vspace{-5mm}
\end{table}

\begin{table*}[t]
    \centering
    \caption{ Accuracy on \textit{Office-Home} under single-source DA setting. The top result is highlighted in bold.
    }
    \label{table:Office-Home-vanilla}
    \vspace{-2mm}
    \begin{adjustbox}{width=1\textwidth}
    \begin{tabular}{@{}l@{~~}c@{~~}c@{~~}c@{~~}c@{~~}c@{~~}c@{~~}c@{~~}c@{~~}c@{~~}c@{~~}c@{~~}c@{~~}|@{~~}c@{~}}
\hline
 \multicolumn{1}{c}{\textbf{Office-Home-DA}}  & \textbf{Ar}$\rightarrow$\textbf{Cl} & \textbf{Ar}$\rightarrow$\textbf{Pr} & \textbf{Ar}$\rightarrow$\textbf{Rw} & \textbf{Cl} $\rightarrow$\textbf{Ar} & \textbf{Cl}$\rightarrow$\textbf{Pr} & \textbf{Cl}$\rightarrow$\textbf{Rw} & \textbf{Pr}$\rightarrow$\textbf{Ar} & \textbf{Pr}$\rightarrow$\textbf{Cl}  & \textbf{Pr}$\rightarrow$\textbf{Rw}  & \textbf{Rw}$\rightarrow$\textbf{Ar}  & \textbf{Rw}$\rightarrow$\textbf{Cl}  & \textbf{Rw}$\rightarrow$\textbf{Pr} &\textbf{Avg.}\\ 
 \hline
\multicolumn{14}{c}{\textbf{Resnet-50}}\\
\hline
		ResNet-50 & 34.90 & 50.00 & 58.00 & 37.40 & 41.90 & 46.20 & 38.50 & 31.20 & 60.40 & 53.90 & 41.20 & 59.90 & 46.10\\
		DAN~\cite{Long:2015}  & 43.60 & 57.00 & 67.90 & 45.80 & 56.50 & 60.40 & 44.00 & 43.60 & 67.70 & 63.10 & 51.50 & 74.30 & 56.30\\
		DANN~\cite{Ganin:DANN:JMLR16} & 45.60 & 59.30 & 70.10 & 47.00 & 58.50 & 60.90 & 46.10 & 43.70 & 68.50 & 63.20 & 51.80 & 76.80 & 57.60\\
		JAN~\cite{LongZ0J17} & 45.90 & 61.20 & 68.90 & 50.40 & 59.70 & 61.00 & 45.80 & 43.40 & 70.30 & 63.90 & 52.40 & 76.80 & 58.30\\
        \hline
        ResNet-50 & 49.36 & 68.86 & 76.25 & 58.71 & 66.18 & 69.33 & 56.59 & 44.80 & 75.80 & 67.66 & 51.21 & 79.52 & 63.69\\ 
        HAFN~\cite{featurenorm_PDA} & 50.20 & 70.10 & 76.60 & 61.10 & 68.00 & 70.70 & 59.50 & 48.40 & 77.30 & 69.40 & 53.00 & 80.20 & 65.40\\
        SAFN~\cite{featurenorm_PDA} & 52.00 & 71.70 & 76.30 & 64.20 & 69.90 & 71.90& 63.70 & 51.40 & 77.10 & 70.90 & 57.10 & 81.50 & 67.30\\
        SAFN+ENT~\cite{featurenorm_PDA}& 52.26 & 73.04 & 77.06 & 66.12 & 72.30 & 72.27 & 64.96 & 52.67 & 78.81 & 72.96 & 58.05 & 82.12 & \textbf{68.55}\\
        \hline
        ResNet-50 & 48.30 & 59.80 & 68.40 & 54.70 & 62.40 & 65.10 & 53.70 & 46.70 & 73.70 & 66.80 &	54.10 &	77.30 &	60.91$\pm$0.15 \\
        {{Jigsaw}$_{\alpha^s=\alpha^t=0.7,\beta=0.8}$} & 47.70  & 58.80 & 67.90 & 57.20  & 64.30  & 66.10  & 56.20  & 50.80  & 75.10  & 67.90  & 55.60  & 78.40  & 62.17$\pm$0.10 \\
        {{Jigsaw}$_{\alpha^s=0, \alpha^t=0.7,\beta=0.8}$} & 47.33  & 58.07 & 67.70 & 57.77  & 63.47  & 65.70  & 56.43  & 50.13  & 74.70  & 68.40  & 55.77  & 79.23  & 62.06$\pm$0.23 \\
        {{Rotation}$_{\alpha^s=\alpha^t=0.8,\beta=0.6}$} & 49.00 & 59.20 & 67.40 & 56.90 & 64.10 & 65.60 & 56.60 & 52.90 & 74.70 & 68.70 & 57.90 & 78.60 & 62.64$\pm$0.13  \\
        {{Rotation}$_{\alpha^s=0,\alpha^t=0.8,\beta=0.6}$} & 48.83 & 56.67 & 67.50 & 57.47 & 63.90 & 65.47 & 56.33 & 52.23 & 74.33 & 68.97 & 57.53 & 78.20 & 62.29$\pm$0.14  \\

        \hline
    \end{tabular}
    \end{adjustbox}
    \vspace{-4mm}
\end{table*}

\subsection{Self-Supervised Domain Adaptation}
\label{sec:exp_DA}

\subsubsection{Single- and Multi-Source Domain Adaptation}
\label{sec:exp_DA_1}

When unlabeled target samples are available at training time we can use any self-supervised task on them.
Indeed we can run patch reordering and orientation recognition on both source and target data to support adaptation of the source classification model.

\begin{table}[tb]
\begin{center} \small
\caption{Multi-source Domain Adaptation results on PACS. 
}\vspace{-5mm}
\begin{adjustbox}{width=0.5\textwidth}
\begin{tabular}{@{}l@{}c@{}c@{~~}c@{~~}c@{~~}c|@{~~}c}
\hline
\multicolumn{2}{@{}c@{}}{\textbf{PACS-DA}}  & \textbf{art\_paint.} & \textbf{cartoon} &  \textbf{sketches} & \textbf{photo} &   \textbf{Avg.}\\ \hline
\multicolumn{7}{@{}c@{}}{\textbf{Resnet-18}}\\
\hline
\multirow{3}{*}{\cite{mancini2018boosting}}& DeepAll & 74.70 & 72.40 & 60.10 & 92.90 & 75.03\\
& Dial & 87.30 & 85.50 & 66.80 & 97.00 &  84.15 \\
& DDiscovery  &  87.70	& 86.90	& 69.60 & 97.00		& 85.30\\
\hline
\multirow{4}{*}{\cite{featurenorm_PDA}}& DeepAll & 76.17 & 73.58 & 55.65 & 96.07 & 75.37$\pm$0.42\\
& HAFN & 84.95 & 79.64 & 64.24 & 97.70 &  81.63$\pm$0.50 \\
& SAFN & 86.78  & 82.72  & 60.26  &  98.26 &  82.01$\pm$0.32 \\
& SAFN+ENT & 89.22  & 87.39  &  60.02 & 98.14  &  83.69$\pm$0.17 \\
\hline
 & DeepAll & 77.83  & 74.26  & 65.81  & 95.71  & 78.40$\pm$0.28 \\
\multicolumn{2}{c}{{Jigsaw$_{\alpha^s=\alpha^t=0.7,\beta=0.8}$}} &  84.49 &	82.07 &	79.86 &	97.98 & 86.10$\pm$0.26 \\
\multicolumn{2}{c}{{Rotation}$_{\alpha^s=\alpha^t=0.8,\beta=0.4}$} & 89.97  & 82.60  &  82.00  &  98.07  & 88.16$\pm$0.51   \\
\multicolumn{2}{c}{{Jigsaw+Rotation $_{{\alpha^s_J=\alpha^t_J=0.2,}\atop{\alpha^s_R=\alpha^t_R=0.8,\beta=0.8}}$}} & 89.67  & 82.87  &  83.93  &  98.17  & {\textbf{88.66}$\pm$0.36}   \\
\hline
\multicolumn{2}{c}{{Jigsaw}$_{\alpha^s=0.7, \alpha^t=0.1,\beta=0.8}$} &  85.40 &	81.49 &	76.93  &	98.35 &	85.54$\pm$1.63 \\
\multicolumn{2}{c}{{Jigsaw}$_{\alpha^s=0.7, \alpha^t=0.3,\beta=0.8}$} &  85.92 &	81.61 &	79.74 &	98.04 &	86.33$\pm$0.58 \\
\multicolumn{2}{c}{{Jigsaw}$_{\alpha^s=0.7, \alpha^t=0.5,\beta=0.8}$} &  87.01 &	81.25 &	78.87 &	98.00 &	86.28$\pm$0.67 \\
\multicolumn{2}{c}{{Jigsaw}$_{\alpha^s=0.7, \alpha^t=0.9,\beta=0.8}$} &  84.21 &	80.38 &	76.64 &	97.86 &	84.77$\pm$0.76\\
\multicolumn{2}{c}{{Rotation}$_{\alpha^s=0.8, \alpha^t=0.1,\beta=0.4}$} &  89.27 & 81.30  &  82.23  & 89.73   &  87.71$\pm$0.13 \\
\multicolumn{2}{c}{{Rotation}$_{\alpha^s=0.8, \alpha^t=0.3,\beta=0.4}$} &  88.73 & 82.20  &  81.47  &  98.27  &  87.67$\pm$0.07 \\
\multicolumn{2}{c}{{Rotation}$_{\alpha^s=0.8, \alpha^t=0.5,\beta=0.4}$} &  89.83 & 80.10  & 81.13   &  98.00  & 87.27$\pm$0.94  \\
\multicolumn{2}{c}{{Rotation}$_{\alpha^s=0.8, \alpha^t=0.9,\beta=0.4}$} &  89.17 &  81.47 &  82.73  &  97.87  & 87.81$\pm$0.21  \\
\multicolumn{2}{c}{{{Jigsaw+Rotation $_{\alpha^t=0, \eta=0}$}}} &  81.07  & 73.97  & 74.67  & 95.93  & 81.41$\pm$0.50   \\
\multicolumn{2}{c}{{{Jigsaw+Rotation $_{\alpha^t=0}$}}} & 82.80  & 77.23  &  77.70  &  97.17  & {83.73$\pm$0.39}   \\
\multicolumn{2}{c}{{{Jigsaw+Rotation $_{\eta=0}$}}} & 84.67  & 78.63  &  80.37  &  97.27  & {85.23$\pm$0.51}   \\
\hline
\end{tabular}
\label{table:resultsDA_PACS}
\end{adjustbox}
\end{center}\vspace{-4mm}
\end{table}

\noindent\emph{Baselines and Datasets:}
We consider as reference four families of DA approaches.
The first is based on measuring the Maximum Mean Discrepancy (MMD, \cite{journalMMD}) across domains and minimizing it to reduce the domain shift: {DAN} \cite{Long:2015}, {JAN} \cite{LongZ0J17}.
The second adopts adversarial approaches as {DANN} \cite{Ganin:DANN:JMLR16} which is based on reverse gradient backpropagation from the auxiliary domain classification network branch.
A third family is that based on batch normalization: {Dial} \cite{carlucci2017just} introduced  adaptive layers to match source and target distribution to a standard gaussian. In {DDiscovery} \cite{mancini2018boosting} the same idea is revisited to first discover the existence of multiple latent domains in the source and then differently adapt their knowledge to the target. 
Finally the fourth family focuses on increasing the feature norms of the two domains with the Hard Adaptive Feature Norm ({HAFN}, \cite{featurenorm_PDA}) method and its step-wise variant {SAFN}.
Several DA approaches minimize the entropy loss as an extra domain alignment condition (\eg SAFN+ENT).
For a fair comparison we also turned on the entropy loss for our method. 
Moreover we solve the self-supervised task either involving both source and target or considering only the latter.
We weight the source and target self-supervised loss equally on the basis of the source cross-validation. 

As datasets we considered 
Office-Home for the single-source experiments and PACS for the multi-source setting. As in the DG case, all the reported results are average over three runs.

\noindent\emph{Results:} 
Tables \ref{table:Office-Home-vanilla}  
shows the single source results on Office-Home.
Our multi-task approach improves over its baseline and over DAN, JAN, DANN but has worse performance than HAFN, SAFN and SAFN+ENT. 
Although not usually presented, we show the specific baseline (ResNet-50) results of the HAFN/SAFN methods to better evaluate their relative gain. Indeed their basic architecture has an extra fully connected layer with respect to a standard ResNet which appears particularly helpful in this cross-domain setting.
We performed also a stability analysis by turning off the self-supervised task on the source $\alpha_s=0$: the minimal results variation indicates that most of the adaptive effect originates from running the self-supervised task on the target.

The multi-source experiments in Table \ref{table:resultsDA_PACS} shed further light on the adaptive abilities of the auxiliary self-supervised objective included in our multi-task approach. When the source domain is rich and covers large style variability, our method is able to outperform not only the batch-normalization based techniques Dial and DDiscovery, but also the state-of-the-art DA approaches HAFN and SAFN which have more difficulties in aligning the norms between the multiple sources and a single target domain. 
Among Jigsaw and Rotation, the second appears more suitable for domain adaptation, with higher performance and better stability to hyper-parameter tuning. 
When the two self-supervised tasks are combined we get on average a small accuracy improvement. 
The bottom part of the table also shows the effect of changing the $\alpha$ value which appears more relevant for Jigsaw than for Rotation. 
{On the Jigsaw+Rotation model we also report the DG result which corresponds to setting $\alpha^t=0$ and $\eta=0$, while keeping all the other chosen parameters. We further show the separate effect of turning off only the self-supervised tasks on the target ($\alpha^t=0$) or the entropy loss ($\eta=0$). This ablation highlights how the major adaptive effect originates from the self-supervised tasks running on the target rather than from the entropy.}

\subsubsection{Partial Domain Adaptation}
\label{sec:exp_DA_2}

The setting with source and target domains sharing exactly the same classes may be too restrictive.
Here we  discuss experimental results on the more realistic PDA setting where the target domain contains only a subset of the source classes. 

\noindent\emph{Baselines:}
We consider as reference five PDA methods 
all based on down-weighting the importance of source classes which are absent in the target. The methods {SAN} \cite{SAN}, {PADA} \cite{PADA_eccv18}, and {DRCN} \cite{li2020deep}, exploit the source model prediction to evaluate the target class distribution.
A different solution is proposed by {IWAN} \cite{IWAN}, where each domain has its own feature extractor and the source sample weight is obtained from the domain recognition model rather than from the source classifier.
The most recent {ETN} \cite{ETN} uses only the relevant source examples to train both the label classifier and the domain discriminator. The relevance (weight) of each source example is computed through an auxiliary domain discriminator, not directly involved in the adaptation phase, which quantifies the source example transferability.

The methods {HAFN} and {SAFN} leverage only the sample norms rather than the whole domain distributions and are quite robust to negative transfer also in the PDA setting, without the need of any weighting mechanism. Thus, we also considered them as reference. Finally, we report the results of DAN and DANN as basic adaptive baselines, to show the effect of methods not originally designed to deal with PDA.

\begin{table}[t!]
\caption{Classification accuracy in the PDA setting on Office-31 (source: 31 classes, target: 10 classes). 
The $^*$ indicates ten-crop testing.}
\vspace{-2mm}
\centering
\begin{adjustbox}{width=0.5\textwidth}

\begin{tabular}{@{}l@{~~}c@{~~}c@{~~}c@{~~}c@{~~}c@{~~}c@{~~}|@{~~}c@{~}}
\hline
 \multicolumn{1}{c}{\textbf{Office-31-PDA}}  & \textbf{A}$\rightarrow$\textbf{W} & \textbf{D}$\rightarrow$\textbf{W} & \textbf{W}$\rightarrow$\textbf{D} & \textbf{A} $\rightarrow$\textbf{D} & \textbf{D}$\rightarrow$\textbf{A} & \textbf{W}$\rightarrow$\textbf{A} & \textbf{Avg.}\\ 
 \hline
\multicolumn{8}{c}{\textbf{Resnet-50}}\\
\hline

Resnet-50           & 75.37 & 94.13 & 98.84 & 79.19 & 81.28 & 85.49 & 85.73 \\ 
DAN\cite{Long:2015}               & 59.32 & 73.90 & 90.45 & 61.78 & 74.95 & 67.64 & 71.34 \\ 
DANN\cite{Ganin:DANN:JMLR16}      & 75.56 & 96.27 & 98.73 & 81.53 & 82.78 & 86.12 & 86.50 \\
\hline
IWAN \cite{IWAN}                & 89.15 & 99.32 & 99.36 & 90.45 & 95.62 & 94.26 & 94.69 \\
SAN*\cite{SAN}                  & 93.90 & 99.32 & 99.36 & 94.27 & 94.15 & 88.73 & 94.96 \\
PADA*\cite{PADA_eccv18}            & 86.54 & 99.32 & 100 & 82.17 & 92.69 & 95.41 & 92.69 \\ 
{DRCN*\cite{li2020deep}} & 86.00 & 88.05 & 95.60 & 100.0 & 95.80 & 100.0 & 94.30\\
{ETN \cite{ETN}}     & 94.52 & 100.0 & 100.0 & 95.03 & 96.21 & 94.64 & \textbf{96.73} \\
\hline
Resnet-50 & 76.05 &	97.52 &	99.36 &	83.23 & 83.89 & 86.18 & 87.71 \\
HAFN \cite{featurenorm_PDA} & 79.89 & 97.63 & 99.57 & 84.93 & 89.59 & 90.08 &	90.28\\
SAFN \cite{featurenorm_PDA} & 84.52 & 97.40 & 98.94 & 84.50 & 92.07 & 92.90 &	91.72\\
SAFN+ENT \cite{featurenorm_PDA} & 87.57	& 98.08 & 99.36 & 88.11 & 93.95 & 93.77 & 93.47\\
\hline
Resnet-50 & 74.35 & 93.90 & 96.81 & 78.13 & 78.46 & 86.81 & 84.74$\pm$0.71 \\
{Jigsaw} & 91.75 & 94.12  & 98.93  & 90.87  & 89.95  & 93.42  &  93.18$\pm$0.46  \\
{Rotation} & 87.91 & 95.14  & 99.57  & 86.84  & 88.73  & 93.98  & 92.03$\pm$1.29    \\
{Jigsaw*}-$\gamma$ & 99.32 & 94.69  & 99.36  & 96.39  & 86.36  & 94.22   & 95.06$\pm$1.86    \\
{Jigsaw*}-$\gamma,\lambda$  & 99.66 & 94.46  & 99.57  &   97.67 & 87.33  & 94.26   &  95.49$\pm$1.19 \\
\hline
\end{tabular}
\end{adjustbox}\vspace{-3mm}
\label{tab:office31} 
\end{table}
\begin{figure}[t!]
\centering
\begin{tabular}{c@{~}c@{~}c}
     \includegraphics[height=0.11\textwidth]{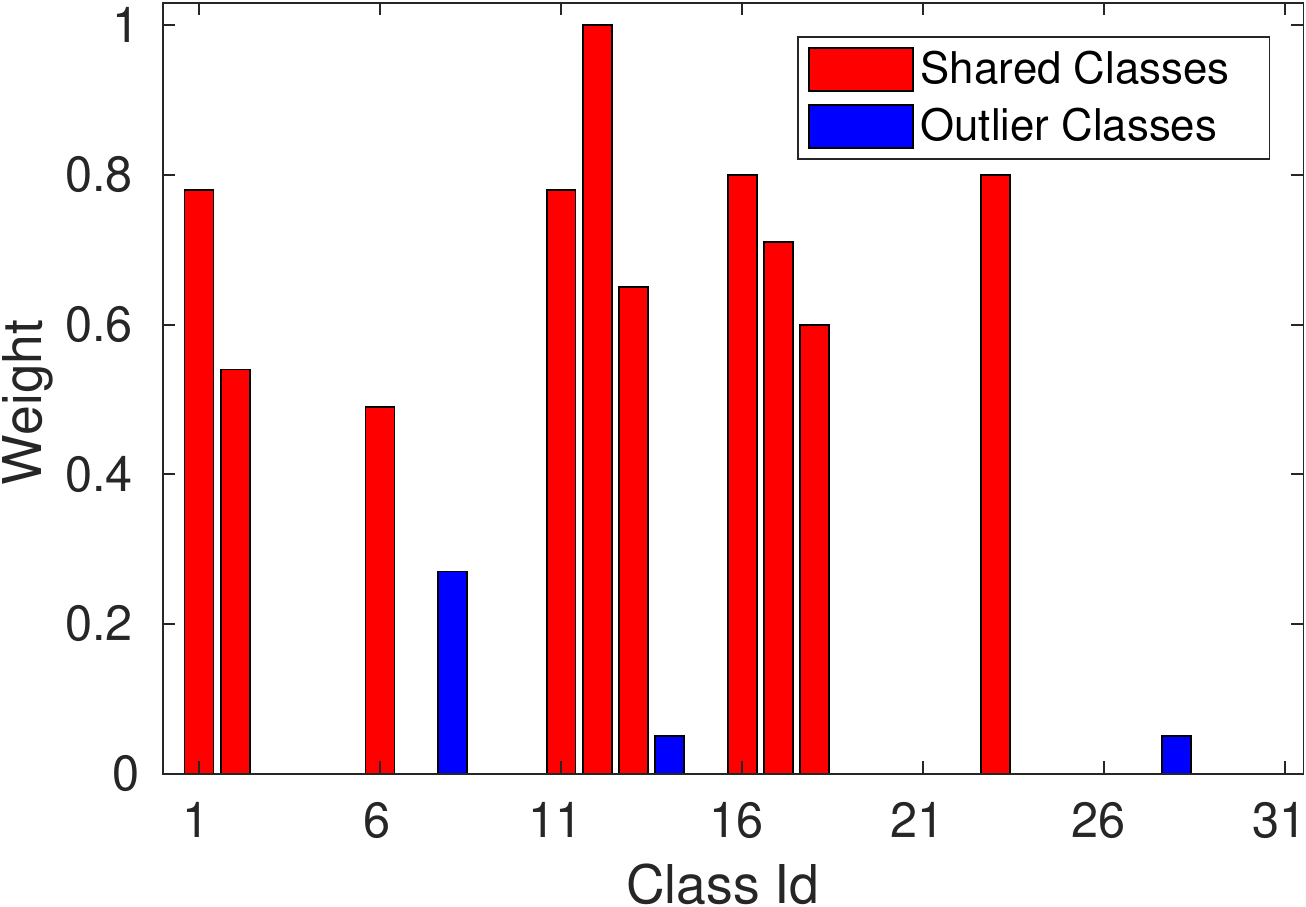} &  \includegraphics[height=0.11\textwidth]{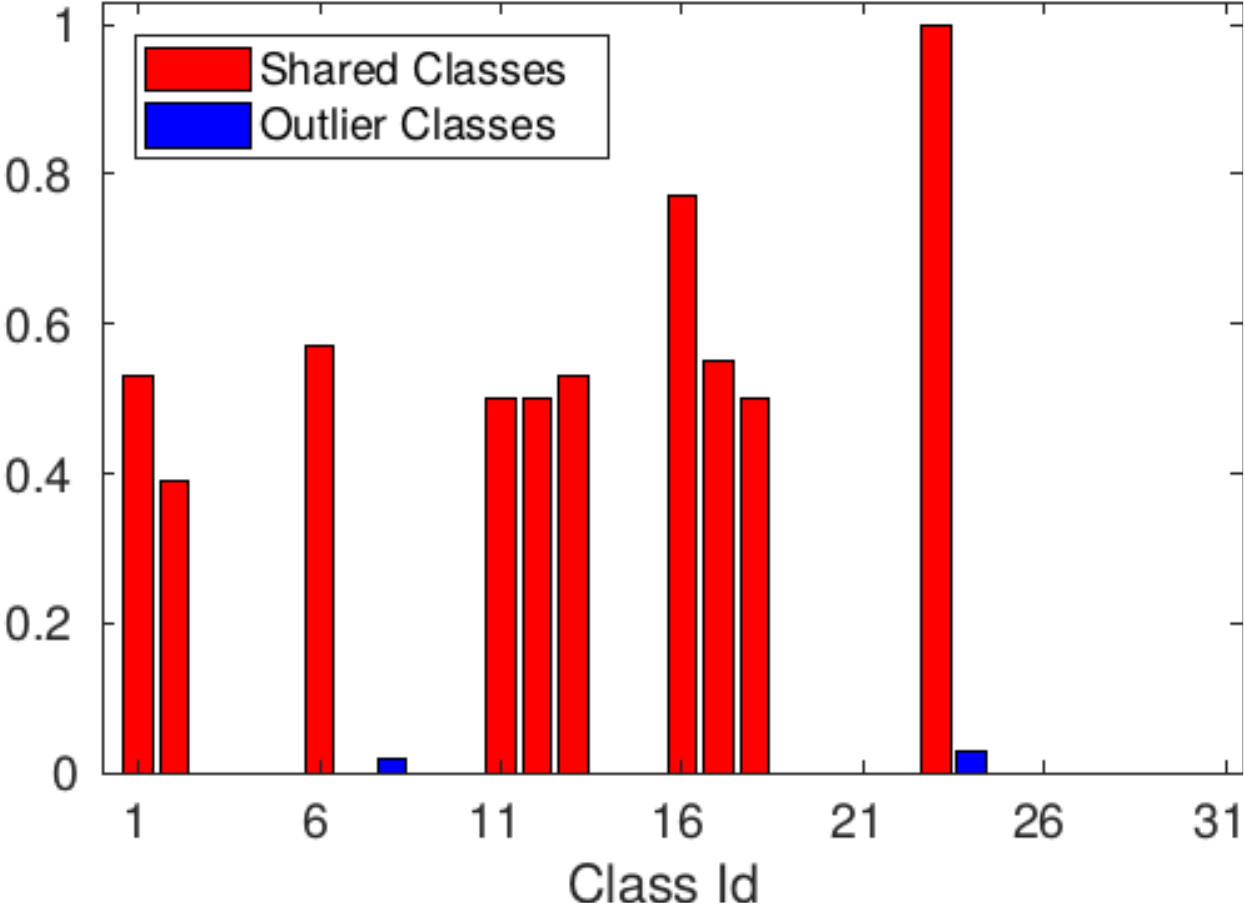} &
     \includegraphics[height=0.11\textwidth]{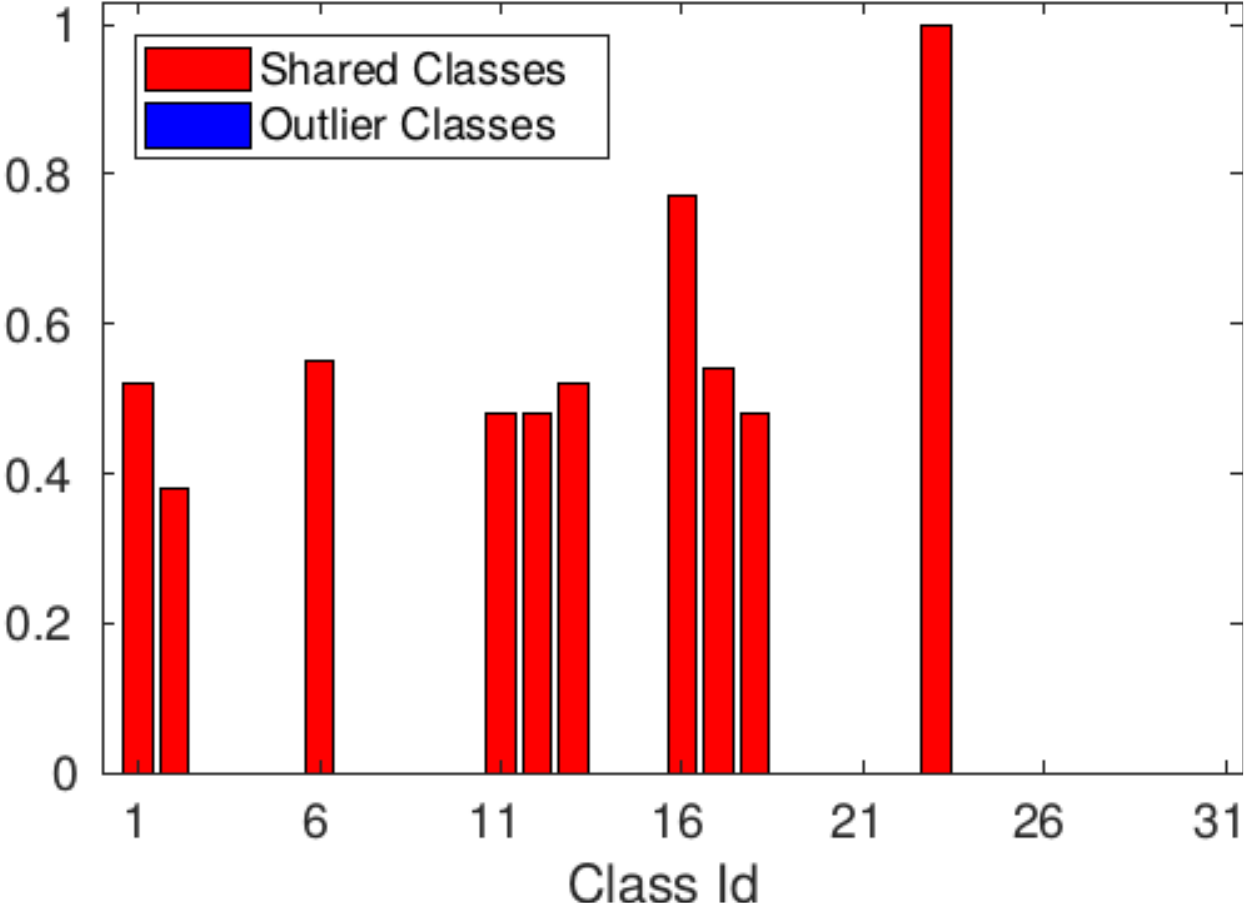}\\
     \footnotesize{PADA}  & \footnotesize{Jigsaw-$\gamma$} & \footnotesize{Jigsaw-$\gamma,\lambda$}\\
\end{tabular}
\vspace{-4mm}
\caption{Histogram showing the elements of the $\gamma$ vector, corresponding to the class weight learned by PADA, SSPDA-$\gamma$ and SSPDA-PADA for the A$\rightarrow$W experiment.}
\label{fig:bars}\vspace{-3mm}
\end{figure}

\noindent\emph{Datasets:}
We follow previous literature in choosing two datasets and their related setting for the PDA experiments. 
We use Office-31 \cite{Saenko:2010} which contains 4652 images of 31 object categories common in office environments. Samples are drawn from three annotated distributions: Amazon (A), Webcam (W) and DSLR (D) which correspond respectively to online vendor website, digital SLR camera and web camera images. 
Similarly to \cite{PADA_eccv18,SAN}, 10 classes are used as target for this dataset (the same classes shared by this dataset with Caltech-256 \cite{caltech256}). 
The second test-bed is VisDA2017, originally used in the 2017 Visual Domain Adaptation challenge (classification track): with respect to the other datasets, it allows us to investigate the proposed multi-task approach on a very large-scale sample size scenario. It has two domains, synthetic 2D object renderings and real images with a total of 208k images organized in 12 categories. We focus on the synthetic-to-real shift, the same considered in the challenge, but keeping only the first 6 categories of the target in alphabetic order. 
For all the experiments we use ResNet-50 as backbone.

\noindent\emph{Results:} Tables \ref{tab:office31} and \ref{tab:visda2017} show the obtained results respectively on Office-31 and VisDA2017 datasets. Each table is organized in four horizontal blocks: the first one shows the results obtained without adaptation or with standard DA methods, the second block illustrates the performance with algorithms designed to deal with PDA, the third one includes
the performance of the norm-based adaptation approaches HAFN/SAFN together with their corresponding ResNet-50 baseline. Finally, the fourth part contains the results of our method. We remind that, as described in Sec. \ref{subsec:PDA}, our approach in the PDA setting does not involve the source data in the auxiliary self-supervised task: {indeed the results obtained in the single source DA setting confirmed that it is possible to set $\alpha^s=0$ without any performance drop (see Table \ref{table:Office-Home-vanilla})}. Moreover, we set $\alpha^t=1.0$ for all the experiments.

All the tables show that both Jigsaw and Rotation outperform the first group of adaptive references. With respect of the PDA techniques in the second group, our method shows better results on VisDA2017 even if many of these competitors take advantage by a ten-crop image evaluation procedure (indicated by the star$^*$).
The top result on Office-31 is obtained by ETN which however, has a dedicated parameter selection procedure for each domain pair, different from our approach for which the parameters are fixed and shared by all the domain pairs of a dataset.
Finally the HAFN/SAFN variants in the third group confirm the effectiveness of the norm-based methods also for PDA. Their results are comparable or worse than ours.

Despite not being tailored for the PDA setting, the obtained performance show that the auxiliary self-supervised task supports adaptation also in this scenario. Given that our solution is orthogonal to the sample selection strategies, we further tried to combine them together to evaluate if they complement each other.
Specifically, we focused on Office-31 and the Jigsaw: we estimated the target class statistics through the weight $\gamma$ and included also a domain discriminator weighted by the parameter $\lambda$, following \cite{PADA_eccv18} as discussed in Sec. \ref{subsec:PDA}. To allow a fair comparison we also adopted the ten-crop evaluation. 
The results in the last two rows of Table \ref{tab:office31}
indicate that estimating the target statistics helps the network to focus only on the shared categories, with an average accuracy improvement of two percentage points over the plain Jigsaw method, getting up to a result comparable with that of ETN considering the standard deviation. We can state that the advantage comes from a better alignment of the domain features:  by comparing the $\gamma$ values on the A$\rightarrow$W domain shift we observe that Jigsaw-$\gamma$ is more precise in identifying the missing classes of the target (see Figure \ref{fig:bars}). 
We indicate with Jigsaw-$\gamma,\lambda$ the case that includes the domain classifier: since the produced features are already well aligned across domains, we fixed $\lambda$-max to $0.1$ and observed a further small average improvement. From the last bar plot on the right of Figure \ref{fig:bars} we also observe a better identification of the target classes.

\begin{table}[t]
    \caption{Classification accuracy in the PDA setting on VisDA2017 (source: 12 classes, target: 6 classes).
    }\vspace{-2mm}
\centering
\begin{tabular}{@{}l@{~~~~~~~~}c@{~~}}
\hline
 \multicolumn{1}{c}{\textbf{VisDA2017-PDA}}  & \textbf{Synthetic}$\rightarrow$\textbf{Real}\\ 
 \hline
\multicolumn{2}{c}{\textbf{Resnet-50}}\\
\hline
Resnet-50          & 45.26 \\
DAN\cite{Long:2015}               & 47.60 \\
DANN\cite{Ganin:DANN:JMLR16}      & 51.01  \\
\hline
PADA*\cite{PADA_eccv18}           & 53.53 \\
{DRCN*\cite{li2020deep}} & 58.20 \\
\hline
Resnet-50 & 49.89\\
HAFN\cite{featurenorm_PDA} & 65.06\\
SAFN\cite{featurenorm_PDA} & 67.65\\
SAFN+ENT*\cite{featurenorm_PDA} & 70.40 \\
\hline 
Resnet-50 & 58.65$\pm$0.66\\
{Jigsaw}  & 68.18$\pm$1.36 \\
{Rotation} & \textbf{71.95}$\pm$0.39 \\
\hline
\end{tabular}
\vspace{-4mm}
\label{tab:visda2017} 
\end{table}

\section{Conclusion}
\label{conclusions}
This work provides an extensive study on the use of self-supervised learning across domains. In particular we focused on solving jigsaw puzzles and recognizing image orientation, showing that they can be easily integrated in a multi-task approach with supervised learning. The results show an improvement in cross-domain robustness and an advantage on generalization performance: the obtained results are competitive with that of more elaborate domain adaptation and domain generalization methods. 
Our work paves the way for many other adaptive methods exploiting the invariances captured by the most recent self-supervised solutions \cite{gidaris2020learning,jenni2020steering}, also beyond object classification towards other challenging tasks like semantic segmentation \cite{Wang_2020_CVPR_SEAM}, detection \cite{dinnocente2020oneshot} or 3D visual learning \cite{alliegro2020joint} where the domain shift effect strongly impacts the deployment of methods in the wild.

\ifCLASSOPTIONcompsoc
  \section*{Acknowledgments}
\else
  \section*{Acknowledgment}
\fi
\label{sec:ack}
This work was partially founded by the ERC grant 637076 RoboExNovo (BC, SB, AD) and took advantage of the GPU donated by NVIDIA (Academic Hardware Grant, TT).

\ifCLASSOPTIONcaptionsoff
  \newpage
\fi

\bibliographystyle{ieee}
\bibliography{ebib}

\end{document}